\documentclass[a4paper, 11pt]{article}
\pdfoutput=1
\usepackage{etoolbox,verbatim}
\usepackage{scalefnt}
\usepackage{graphicx}
\usepackage{subcaption}
\usepackage{booktabs}

\newtoggle{arxiv}
\toggletrue{arxiv}
\newtoggle{toreturn}
\togglefalse{toreturn}
\newtoggle{icml}
\togglefalse{icml}
\newtoggle{customthms}
\toggletrue{customthms}
\newtoggle{colt}
\togglefalse{colt}

\usepackage{xcolor}
\usepackage{tcolorbox}
\tcbuselibrary{skins,breakable}

\tcbset{
  aibox/.style={
      width=\linewidth,
      top=6pt,
      bottom=0pt,
      left=3.5pt,
      right=1.5pt,
colback=blue!60!gray!5,
      colframe=black,
      colbacktitle=black,
      enhanced,
      center,
      attach boxed title to top left={yshift=-0.1in,xshift=0.15in},
      boxed title style={boxrule=0pt,colframe=white,},
    }
}
\newtcolorbox{AIbox}[2][]{aibox,title=#2,#1}

\definecolor{primalcolor}{HTML}{A60000}
\definecolor{contrarycolor}{HTML}{00A6A6}
\definecolor{darkcontrarycolor}{HTML}{004C4C}
\definecolor{lightblue}{HTML}{2970CC}
\definecolor{lightpurple}{HTML}{673147}
\definecolor{ForestGreen}{HTML}{FF5733}
\definecolor{myred}{HTML}{AA4A44}
\definecolor{hyppurple}{HTML}{800080}

\newcommand{\linkcolor}{darkcontrarycolor}
\newcommand{\urlcolor}{darkcontrarycolor}
\newcommand{\citecolor}{darkcontrarycolor}

\newcommand{\thmcolordark}{red!30!black}

\usepackage{wrapfig}
\usepackage[T1]{fontenc}
\usepackage{courier}
\usepackage{capt-of}
\usepackage[font=small,labelfont=bf]{caption}
\usepackage{tocloft}

\usepackage{natbib}
\usepackage{amssymb,upgreek,bm}
\usepackage{mathtools}
\usepackage[english]{babel}  
\usepackage{multirow,tcolorbox,tikz-cd,turnstile,xspace,xparse}
\usepackage{relsize,breakcites,appendix}
\usepackage{longtable,tablefootnote,booktabs,float,makecell}
\usepackage[normalem]{ulem}
\usepackage{tabu}
\usepackage[shortlabels]{enumitem} 
\usepackage{footnote}
\usepackage{boxedminipage}
\usepackage{multirow,nicefrac}
\usepackage{verbatim} 
\usepackage{pifont}   
\usepackage[utf8]{inputenc}
\usepackage{arydshln}

\usepackage[colorlinks=true,linkcolor=\linkcolor,urlcolor=\urlcolor,citecolor=\citecolor,breaklinks]{hyperref}

\usepackage{prettyref}

\usepackage[capitalise,nameinlink]{cleveref}

\Crefname{table}{Table}{Tables}
\AddToHook{cmd/appendix/before}{%
    \crefalias{section}{appendix}%
    \crefalias{subsection}{appendix}
}
\iftoggle{colt}{
    \DeclareRobustCommand{\qed}{
        \usepackage{thmtools}
          \ifmmode \mathqed
          \else
            \leavevmode\unskip\penalty9999 \hbox{}\nobreak\hfill
            \quad\hbox{\qedsymbol}%
          \fi
    }
}
{
    \usepackage{amsthm}
     
}
\usepackage{thm-restate}

\iftoggle{arxiv}
{
    \usepackage{mathrsfs}

    \usepackage[
  letterpaper,
  left=1in,
  right=1in,
  top=0.9in,
  bottom=0.9in,
  headheight=14pt,
  headsep=14pt,
  footskip=20pt
]{geometry}

\usepackage{fancyhdr}
\usepackage{lipsum} 

\pagestyle{fancy}
\fancyhf{} 
\fancyfoot[C]{\small \thepage}

\renewcommand{\headrulewidth}{0.4pt}
\renewcommand{\footrulewidth}{0pt}

    \usepackage[bitstream-charter]{mathdesign}
     
    \usepackage[scaled=0.92]{PTSans}

    \usepackage{caption}
    \usepackage{subcaption}

    \usepackage{algorithm}
    \usepackage{algorithmic}
}
{
    \usepackage{mathrsfs}
    \usepackage{eqparbox}
}

\DeclareMathAlphabet{\mathbfsf}{\encodingdefault}{\sfdefault}{bx}{n}

\numberwithin{equation}{section}

\iftoggle{arxiv}
{
    \newcommand{\colorbold}[1]{
    \textbf{\textcolor{\thmcolor}{#1}}}}
{
    \newcommand{\colorbold}[1]{\textbf{#1}
}
}
\iftoggle{arxiv}
{
    \newcommand{\colorpar}[1]{
    \paragraph{\textcolor{\thmcolor}{#1}}}
}
{
    \newcommand{\colorpar}[1]{\paragraph{#1}}
}

\usepackage{thmtools}

\Crefname{equation}{Eq.}{Eqs.}
\Crefname{assumption}{Assumption}{Assumptions}

\Crefname{component}{Component}{Components}
\Crefname{condition}{Condition}{Conditions}
\Crefname{claim}{Claim}{Claims}
\Crefname{property}{Property}{Properties}
\Crefname{construction}{Construction}{Constructions}

\iftoggle{customthms}
{
\declaretheoremstyle[
    headformat=\normalfont\textcolor{\thmcolordark}{\bfseries\NAME\,\NUMBER}\NOTE,%
    notefont={\normalfont\textcolor{\thmcolordark}{\bfseries}}, 
    notebraces={}{},
    bodyfont=\normalfont\itshape,
    spaceabove = 6pt,
    spacebelow = 6pt,
    ]{coloredthmversion}

\declaretheoremstyle[
    headformat=\normalfont\textcolor{\thmcolordark}{\bfseries\NAME\,\NUMBER}\NOTE,%
    bodyfont=\normalfont\itshape,
    spaceabove = 6pt,
    spacebelow = 6pt,
    ]{coloredthm}

\declaretheoremstyle[
    headformat=\normalfont\textcolor{\thmcolordark}{\bfseries\NAME\,\NUMBER}\NOTE,%
    bodyfont=\normalfont,
    spaceabove = 6pt,
    spacebelow = 6pt,
    ]{coloreddef}
    \theoremstyle{coloredthmversion}

}
{}

\iftoggle{customthms}{
  \theoremstyle{coloredthm}
  \newtheorem{theorem}{Theorem}
  
  \newtheorem{lemma}{Lemma}[section]
  \newtheorem{corollary}{Corollary}[section]
  \newtheorem{proposition}[lemma]{Proposition}

    \newtheorem*{thminformal*}{Informal Theorem}

}
{
  \newtheorem{lemma}{Lemma}[section]

}

\iftoggle{customthms}{
    \theoremstyle{coloreddef}
    \newtheorem{definition}{Definition}[section]

    \newtheorem{property}{Property}[section]
}
{
  
}

\newtheorem{assumption}{Assumption}[section]
\newtheorem{condition}{Condition}[section]

\makeatletter
\newcommand{\neutralize}[1]{\expandafter\let\csname c@#1\endcsname\count@}
\makeatother

\iftoggle{customthms}{
    \newtheoremstyle{named}{}{}{\itshape}{}{\bfseries}{}{.5em}{\Cref{#3} {\normalfont (informal)} }{}
    \theoremstyle{named}
    
    \theoremstyle{plain}
}
{}

\iftoggle{customthms}
{
\newtheorem*{theorem*}{Theorem}
\newtheorem*{lemma*}{Lemma}
\newtheorem*{corollary*}{Corollary}
\newtheorem*{proposition*}{Proposition}
\newtheorem*{claim*}{Claim}
\newtheorem*{fact*}{Fact}
\newtheorem*{observation*}{Observation}
\newtheorem*{definition*}{Definition}
\newtheorem*{remark*}{Remark}
\newtheorem*{example*}{Example}
}{}

\def\x{\mathbf{x}}

\def\ddefloop#1{\ifx\ddefloop#1\else\ddef{#1}\expandafter\ddefloop\fi}
\def\ddef#1{\expandafter\def\csname bb#1\endcsname{\ensuremath{\mathbb{#1}}}}
\ddefloop ABCDEFGHIJKLMNOPQRSTUVWXYZ\ddefloop

\def\ddefloop#1{\ifx\ddefloop#1\else\ddef{#1}\expandafter\ddefloop\fi}
\def\ddef#1{\expandafter\def\csname frak#1\endcsname{\ensuremath{\mathfrak{#1}}}}
\ddefloop ABCDEFGHIJKLMNOPQRSTUVWXYZ\ddefloop

\def\ddefloop#1{\ifx\ddefloop#1\else\ddef{#1}\expandafter\ddefloop\fi}
\def\ddef#1{\expandafter\def\csname fr#1\endcsname{\ensuremath{\mathfrak{#1}}}}
\ddefloop ABCDEFGHIJKLMNOPQRSTUVWXYZ\ddefloop

\def\ddefloop#1{\ifx\ddefloop#1\else\ddef{#1}\expandafter\ddefloop\fi}
\def\ddef#1{\expandafter\def\csname eul#1\endcsname{\ensuremath{\EuScript{#1}}}}
\ddefloop ABCDEFGHIJKLMNOPQRSTUVWXYZ\ddefloop

\def\ddefloop#1{\ifx\ddefloop#1\else\ddef{#1}\expandafter\ddefloop\fi}
\def\ddef#1{\expandafter\def\csname scr#1\endcsname{\ensuremath{\mathscr{#1}}}}
\ddefloop ABCDEFGHIJKLMNOPQRSTUVWXYZ\ddefloop

\def\ddefloop#1{\ifx\ddefloop#1\else\ddef{#1}\expandafter\ddefloop\fi}
\def\ddef#1{\expandafter\def\csname b#1\endcsname{\ensuremath{\mathbf{#1}}}}
\ddefloop ABCDEFGHIJKLMNOPQRSTUVWXYZ\ddefloop

\def\ddefloop#1{\ifx\ddefloop#1\else\ddef{#1}\expandafter\ddefloop\fi}
\def\ddef#1{\expandafter\def\csname bhat#1\endcsname{\ensuremath{\hat{\mathbf{#1}}}}}
\ddefloop ABCDEFGHIJKLMNOPQRSTUVWXYZ\ddefloop

\def\ddefloop#1{\ifx\ddefloop#1\else\ddef{#1}\expandafter\ddefloop\fi}
\def\ddef#1{\expandafter\def\csname btil#1\endcsname{\ensuremath{\tilde{\mathbf{#1}}}}}
\ddefloop ABCDEFGHIJKLMNOPQRSTUVWXYZ\ddefloop

\def\ddefloop#1{\ifx\ddefloop#1\else\ddef{#1}\expandafter\ddefloop\fi}
\def\ddef#1{\expandafter\def\csname bst#1\endcsname{\ensuremath{\mathbf{#1}^\star}}}
\ddefloop ABCDEFGHIJKLMNOPQRSTUVWXYZ\ddefloop

\def\ddefloop#1{\ifx\ddefloop#1\else\ddef{#1}\expandafter\ddefloop\fi}
\def\ddef#1{\expandafter\def\csname bst#1\endcsname{\ensuremath{\mathbf{#1}^\star}}}
\ddefloop abcdeghijklmnopqrstuvwxyz\ddefloop

\def\ddefloop#1{\ifx\ddefloop#1\else\ddef{#1}\expandafter\ddefloop\fi}
\def\ddef#1{\expandafter\def\csname bhat#1\endcsname{\ensuremath{\hat{\mathbf{#1}}}}}
\ddefloop abcdefghijklmnopqrstuvwxyz\ddefloop

\def\ddefloop#1{\ifx\ddefloop#1\else\ddef{#1}\expandafter\ddefloop\fi}
\def\ddef#1{\expandafter\def\csname b#1\endcsname{\ensuremath{\mathbf{#1}}}}
\ddefloop abcdeghijklnopqrstuvwxyz\ddefloop

\def\ddefloop#1{\ifx\ddefloop#1\else\ddef{#1}\expandafter\ddefloop\fi}
\def\ddef#1{\expandafter\def\csname barb#1\endcsname{\ensuremath{\bar{\mathbf{#1}}}}}
\ddefloop abcdefghijklmnopqrstuvwxyz\ddefloop

\def\ddef#1{\expandafter\def\csname c#1\endcsname{\ensuremath{\mathcal{#1}}}}
\ddefloop ABCDEFGHIJKLMNOPQRSTUVWXYZ\ddefloop
\def\ddef#1{\expandafter\def\csname h#1\endcsname{\ensuremath{\widehat{#1}}}}
\ddefloop ABCDEFGHIJKLMNOPQRSTUVWXYZ\ddefloop
\def\ddef#1{\expandafter\def\csname hc#1\endcsname{\ensuremath{\widehat{\mathcal{#1}}}}}
\ddefloop ABCDEFGHIJKLMNOPQRSTUVWXYZ\ddefloop
\def\ddef#1{\expandafter\def\csname t#1\endcsname{\ensuremath{\widetilde{#1}}}}
\ddefloop ABCDEFGHIJKLMNOPQRSTUVWXYZ\ddefloop
\def\ddef#1{\expandafter\def\csname tc#1\endcsname{\ensuremath{\widetilde{\mathcal{#1}}}}}
\ddefloop ABCDEFGHIJKLMNOPQRSTUVWXYZ\ddefloop

\iftoggle{arxiv}
{
\newcommand{\togglepar}[1]{\colorpar{#1}}
}
{
\newcommand{\togglepar}[1]{\textbf{#1}}
}

\newcommand{\algcomment}[1]{\texttt{\color{blue} #1}}

\DeclarePairedDelimiter{\prn}{(}{)}

\newcommand{\I}{\mathbf{I}}

\DeclareMathOperator*{\argmin}{arg\,min}
\DeclareMathOperator*{\argmax}{arg\,max}

\DeclareMathOperator*{\Expop}{\mathbb{E}}

\let\Pr\relax
\DeclareMathOperator{\Pr}{\mathbb{P}}

\newcommand{\Exp}{\mathbb{E}}

\iftoggle{arxiv}{
\newcommand{\Normal}{\mathrm{N}}
}{
\newcommand{\Normal}{\mathcal{N}}
}
\newcommand{\iidsim}{\overset{\mathrm{i.i.d}}{\sim}}

\newcommand{\eye}{\mathbf{I}}

\newcommand{\ballkr}[1][r]{\cB_{k}(r)}

\newcommand{\rmd}{\mathrm{d}}

\DeclareMathSymbol{\shortminus}{\mathbin}{AMSa}{"39}

\newcommand{\R}{\mathbb{R}}

\newcommand{\thetaema}{\theta_{\textsc{ema}}}
\newcommand{\flagfont}[1]{\texttt{#1}}
\newcommand{\flagvalfont}[1]{\textbf{\texttt{#1}}}
\newcommand{\Qtarg}{Q_{\mathrm{targ}}}
    \newcommand{\Buffer}{\cB}
    \newcommand{\criticflag}{\flagfont{critic\_flag}}
    \newcommand{\criticflagmean}{\flagvalfont{mean}}
    \newcommand{\criticflagmin}{\flagvalfont{min}}
    \newcommand{\criticflagsub}{\flagvalfont{subsample}}

\iftoggle{arxiv}
{\newcommand{\Lcritic}{L_{\mathrm{critic}}}}
{\newcommand{\Lcritic}{L_{Q}}}

\newcommand\mdpenv{M_\textsc{Env}}
\newcommand{\Jcost}{J}

\newcommand{\piigp}[1][\theta]{\bar{\pi}_{#1}}
\newcommand{\igpfirst}{K}

\newcommand{\st}{s_t}
\newcommand{\stnext}{s_{t+1}}
\newcommand{\igplast}{0} 
\newcommand{\igpiter}{k}

\newcommand{\igpinc}{k-1}

\newcommand{\xlast}{x_{\igplast}}
\newcommand{\xiter}{x_{\igpiter}}

\newcommand{\gaussinit}{\mathrm{N}(0, \mathrm{I})}

\newcommand{\afirst}{a_{\igpfirst}}

\newcommand{\alast}{a_{\igplast}}
\newcommand{\aiter}{a_{\igpiter}}
\newcommand{\ainc}{a_{\igpinc}}
\newcommand{\aitert}[1][t]{a_{#1,\igpiter}}
\newcommand{\ainct}[1][t]{a_{#1,\igpinc}}

\newcommand{\aplant}{a_{\textsc{BoN},t}}
\newcommand{\abont}{a_{\mathrm{BoN},t}}

\newcommand{\afirstt}[1][t]{a_{#1,\igpfirst}}

\newcommand{\alastt}[1][t]{a_{#1,\igplast}}

\newcommand{\GCP}{GCP}

\newcommand{\alastnextt}[1][t+1]{a_{#1,\igplast}}

\newcommand{\useoffline}{\texttt{use}\_\texttt{offline}}

\newcommand{\piigpbc}{\piigp^{\mathrm{BC}}}
\newcommand{\targ}{\mathrm{targ}}
\newcommand{\sample}{\textsc{Take}\_\textsc{Step}}
\newcommand{\updateq}{{\normalfont\textsc{UpdateQ}}}
\newcommand{\updateigp}{{\normalfont\textsc{UpdateGCP}}}

\newcommand{\done}{\texttt{done}}
\newcommand{\rb}{\mathcal{D}_{\text{roll}}}
\newcommand{\succb}{\mathcal{D}_{\text{succ}}}
\newcommand{\rboff}{\mathcal{D}_{\text{off}}}
\newcommand{\rbbatch}{\textsc{B}_{\text{itr}}}
\newcommand{\sbbatch}{\textsc{B}_{\text{succ}}}
\newcommand{\stsucc}{\st^{\text{succ}}}
\newcommand{\atsucc}{\alastt^{\text{succ}}}

\newcommand{\ratio}{\omega_\theta}
\newcommand{\group}{\textsc{G}}
\newcommand{\advgrpo}{\hat{A}^\group}

\newcommand{\Lbc}{L_{\texttt{BC}}}
\newcommand{\Lppo}{L_{\texttt{PPO}}}
\newcommand{\Ltotal}{L_{\texttt{Total}}}

\newcommand{\nwarmup}{N_\mathrm{warmup}}
\newcommand{\sampletau}{\tau \sim \pi_\theta(\tau)}
\newcommand{\piold}{\pi_{\theta_{\mathrm{old}}}}

\newcommand{\initialize}{\textsc{Initialize}}
\newcommand{\useofflinerl}{\texttt{use\_calql}}
\newcommand{\warmupcritic}{\texttt{warmup\_critic}}
\newcommand{\utd}{\texttt{utd}}
\newcommand{\offlineratio}{\texttt{r}_\texttt{offline}}

\newcommand{\cmark}{\ding{51}}
\newcommand{\xmark}{\ding{55}}
\newcommand{\pmark}{\textcolor{gray}{\ding{51}}}
\newcommand{\pcmark}{%
  \raisebox{0.7ex}{%
    \makebox[0pt][l]{\textcolor{black}{\rule{1.7ex}{1pt}}}%
  }\textcolor{black}{\cmark}%
}

\newcommand{\yesyes}{\cmark\,\cmark}
\newcommand{\nono}{\xmark\,\xmark}
\newcommand{\kindakinda}{\pmark\,\pmark}

\newcommand{\yeskinda}{\cmark\,\pmark}

\newcommand{\kindano}{\pmark\,\xmark}

\newcommand{\mehkinda}{\pcmark\,\pmark}

\newcommand{\mehno}{\pcmark\,\xmark}

\newcommand{\omegaslow}{\omega_{\theta,\mathrm{slow}}}
\newcommand{\Pslow}{P_\mathrm{slow}}

\newcommand{\piref}{\pi_\mathrm{slow}}
\newcommand{\tauref}{\tau_\mathrm{slow}}

\newcommand{\pifive}{\pi0.5}

\newcommand{\aindex}{j}
\newcommand{\advcons}{\hat{A}^\mathrm{cons}}
\newcommand{\advindie}{A_{\aindex,m}}

\newcommand{\fk}{\textsc{Franka-Kitchen}}
\newcommand{\fkcomplete}{\textsc{Kitchen-Complete}}
\newcommand{\fkpartial}{\textsc{Kitchen-Partial}}
\newcommand{\fkmixed}{\textsc{Kitchen-Mixed}}
\newcommand{\robomimic}{\textsc{Robomimic}}
\newcommand{\rmsquare}{\textsc{Square}}
\newcommand{\rmtoolhang}{\textsc{Toolhang}}
\newcommand{\rmtransport}{\textsc{Transport}}

\newcommand{\libero}{\textsc{LIBERO}}

\newcommand{\LCFM}{\hat L_{\mathrm{CFM}}}
\newcommand{\ratiofpo}{\hat r_{\mathrm{FPO}}}

\newcommand{\algofont}[1]{{\scalefont{1.0}{\texttt{\textbf{#1}}}}}

\definecolor{dppocol}{HTML}{777777}
\definecolor{fpocol}{HTML}{6EC1C5}
\definecolor{awrcol}{HTML}{8A8635}
\definecolor{awogpocol}{HTML}{97Af8D}
\definecolor{miqcol}{HTML}{FE81D4}
\definecolor{fqlcol}{HTML}{FFB366}
\definecolor{dsrlcol}{HTML}{925BFF}
\definecolor{qccol}{HTML}{2A52BE}
\definecolor{expocol}{HTML}{50C878}
\definecolor{srcol}{HTML}{5B1166}
\definecolor{bpttcol}{HTML}{FFB366}

\newcommand{\DPPOcolor}{dppocol}
\newcommand{\DPPOnc}{\algofont{DPPO}}

\newcommand{\DPPO}{{\color{\DPPOcolor}\DPPOnc}}

\newcommand{\FPOcolor}{fpocol}
\newcommand{\FPOnc}{\algofont{OFPO++}}

\newcommand{\FPO}{{\color{\FPOcolor}\FPOnc}}

\newcommand{\AWRcolor}{awrcol}
\newcommand{\AWRnc}{\algofont{AWR}\text{-}\algofont{FM}}

\newcommand{\AWR}{{\color{\AWRcolor}\AWRnc}}

\newcommand{\AWOGPOcolor}{awogpocol}
\newcommand{\AWOGPOnc}{\algofont{AW}\text{-}\algofont{OGPO}}

\newcommand{\AWOGPO}{{\color{\AWOGPOcolor}\AWOGPOnc}}
\newcommand{\AWOGPOpos}{{\color{\AWOGPOcolor}\AWOGPOnc\text{-}\algofont{NN}}}

\newcommand{\FQLcolor}{fqlcol}
\newcommand{\FQLnc}{\algofont{FQL}}

\newcommand{\FQL}{{\color{\FQLcolor}\FQLnc}}

\newcommand{\MIQcolor}{miqcol}
\newcommand{\MIQnc}{\algofont{MIQ}}

\newcommand{\MIQ}{{\color{\MIQcolor}\MIQnc}}

\newcommand{\DSRLcolor}{dsrlcol}
\newcommand{\DSRLnc}{\algofont{DSRL}}

\newcommand{\DSRL}{{\color{\DSRLcolor}\DSRLnc}}
\newcommand{\DSRLplus}{{\color{\DSRLcolor}\DSRLnc+}}

\newcommand{\QCcolor}{qccol}
\newcommand{\QCnc}{\algofont{QC}}

\newcommand{\QC}{{\color{\QCcolor}\QCnc}}

\newcommand{\QCplus}{{\color{\QCcolor}\QCnc +}}

\newcommand{\SRcolor}{srcol}
\newcommand{\SRnc}{\algofont{S/R}}

\newcommand{\SR}{{\color{\SRcolor}\SRnc}}

\newcommand{\EXPOcolor}{expocol}
\newcommand{\EXPOnc}{\algofont{EXPO}}

\newcommand{\EXPOplus}{{\color{\EXPOcolor}\EXPOnc +}}
\newcommand{\EXPO}{{\color{\EXPOcolor}\EXPOnc}}
\newcommand{\BPTTcolor}{bpttcol}
\newcommand{\BPTTnc}{\algofont{BPTT}}

\newcommand{\BPTT}{{\color{\BPTTcolor}\BPTTnc}}

\newcommand{\OGPOcolor}{red!70!black}
\newcommand{\OGPOtext}{OGPO}
\newcommand{\OGPOnc}{\algofont{\OGPOtext}}

\newcommand{\OGPO}{{\color{\OGPOcolor}\OGPOnc}}
\newcommand{\OGPOnn}{{\color{\OGPOcolor}\algofont{\OGPOtext-NN}}}
\newcommand{\OGPOplus}{{\color{\OGPOcolor}\algofont{\OGPOtext+}}}
\newcommand{\OGPOplusca}{{\color{\OGPOcolor}\algofont{\OGPOtext+CA}}}
\newcommand{\OGPOchisq}{{\color{\OGPOcolor}\algofont{\OGPOtext+}\chi^2}}

\newcommand{\AdroitEnv}{\texttt{AdroitHand}}

\newcommand{\Robomimic}{\textsc{Robomimic}}

\newcommand{\KitchenEnv}{\textsc{Franka}-\textsc{Kitchen}}

\usepackage{preamble/color-edits}

\addauthor{ms}{magenta}
\addauthor{sbp}{orange}
\addauthor{jz}{cyan}
\addauthor{ag}{blue}
\addauthor{ma}{red}

\newcommand{\REF}{{\color{red} \textbf{REF??}}}

\iftoggle{toreturn}
{
\newcommand{\RETURNN}[1]{{\color{red} [\textbf{RETURN??}#1]}}
}
{
\newcommand{\RETURNN}[1]{}
}
\newcommand{\TODO}{\textbf{\color{red} TODO}}
\newcommand{\ignore}[1]{}

\iftoggle{arxiv}
{
\input{preamble/arxiv_title}
}
{}

\title{\OGPO{}: Sample Efficient Full-Finetuning of Generative Control Policies}
\author{
\small
Sarvesh Patil\footnote{\texttt{\{sarveshp, mananaga, ssaxena2, dchen3, chaoyip, naichieh, giria, msimchow\}@andrew.cmu.edu}\label{foot:cmu}}\textsuperscript{,$\$$},
Mitsuhiko Nakamoto\footnote{\texttt{\{nakamoto, svlevine\}@eecs.berkeley.edu} \label{foot:ucb}}\textsuperscript{,$\$$},~
Manan Agarwal\textsuperscript{\ref{foot:cmu}}\textsuperscript{,$\ddagger$},~
Shashwat Saxena\textsuperscript{\ref{foot:cmu}}\textsuperscript{,$\ddagger$},~
Jesse Zhang\footnote{\texttt{\{jessezha, cleahw, abhgupta\}@cs.washington.edu} \label{foot:uw}}\textsuperscript{,$\ddagger$},~ \\ \small
Giri Anantharaman\textsuperscript{\ref{foot:cmu}},~ 
Cleah Winston\textsuperscript{\ref{foot:uw}},~ 
Chaoyi Pan\textsuperscript{\ref{foot:cmu}},~
Douglas Chen\textsuperscript{\ref{foot:cmu}},~
Nai-Chieh Huang\textsuperscript{\ref{foot:cmu}},~\\\small
Zeynep Temel\textsuperscript{\ref{foot:cmu}}, ~  
Oliver Kroemer\textsuperscript{\ref{foot:cmu}},~ 
Sergey Levine\textsuperscript{\ref{foot:ucb}},~
Abhishek Gupta \textsuperscript{\ref{foot:uw}}\textsuperscript{\ref{foot:tri}},~\\ \small 
Hongkai Dai\footnote{\texttt{\{hongkai.dai,paarth.shah\}\@tri.global}\label{foot:tri}}\textsuperscript{,$\dagger$},~
Paarth Shah\textsuperscript{\ref{foot:tri},$\dagger$},~ 
Max Simchowitz\textsuperscript{\ref{foot:cmu},$\dagger$}~\\
\vspace{-.3em}
 \rule{.38\textwidth}{.7pt}
\\
\footnotesize
$^{\$}$Project lead. $\ddagger$Equal Contribution. $^\dagger$Equal advising. \\
\footnotesize
$^{a}$Carnegie Mellon University ~~
$^{b}$University of California, Berkeley ~~
$^{c}$University of Washington ~~\\
\footnotesize
$^{d}$Toyota Research Institute ~~
}
\vspace{-.5em}
\paperwebsite{https://simchowitzlabpublic.github.io/ogpo-site/}
\papercode{https://github.com/simchowitzlabpublic/OGPO_public}
\paperdate{\today}
\date{\vspace{-.5em}}

\begin{document}

\begin{tcolorbox}[
colback=blue!60!gray!5, colframe=gray!50, 
boxrule=0pt, 
    arc=2mm
]
\maketitle
\vspace{-.7cm}
\tcbline
\begin{minipage}[t]{1.0\linewidth}
    \centering
\includegraphics[width=.99\textwidth]{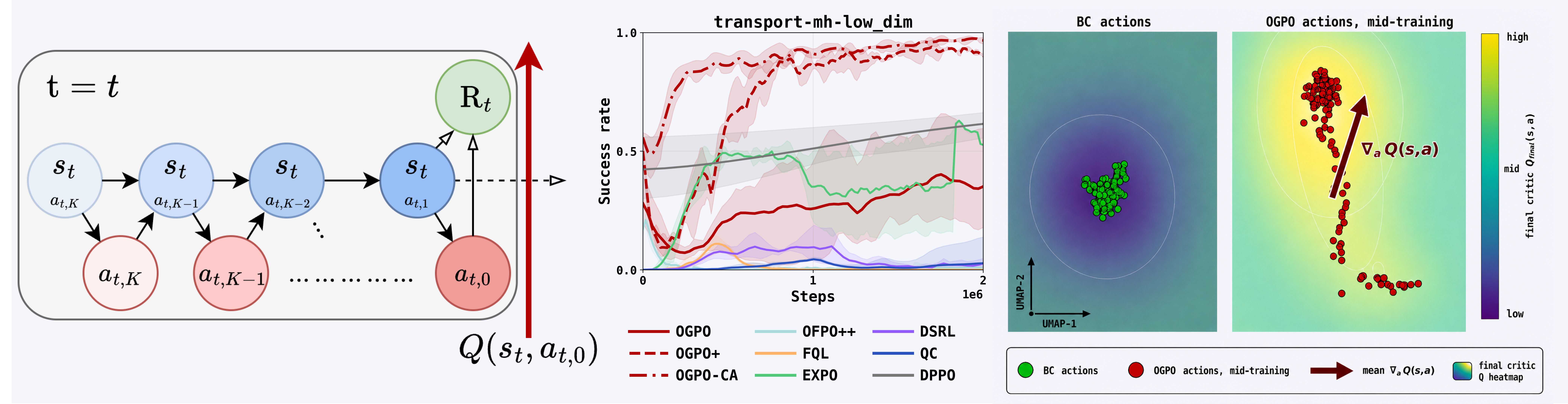}
        \captionof{figure}{
        \footnotesize  \textbf{\OGPO{} enables sample-efficient full-policy finetuning of generative control policies.} \textbf{Left:} A generative control policy (GCP) represents action generation as a sequential computation, constituting a \emph{denoising MDP} at each environment step. Whereas prior work \citep{ren2024diffusion}, embeds the denoising MDP into the environment to form a bi-level MDP, \OGPO{} severs this connection, and instead maximizes an off-policy critic as a terminal reward via PPO-style optimization over the denoising trajectories. \textbf{Middle:} We show that \OGPO substantially improves sample efficiency on challenging manipulation tasks like \robomimic{} \rmtransport{} compared to prior GCP finetuning methods, even with limited hyperparameter tuning. \textbf{Right:} Despite its sample efficiency, \OGPO{} preserves non-trivial variance in the action distribution, preserving the capacity for exploration. As show, action variance is ``squeezed'' to be \emph{perpendicular to critic gradients} during the middle of the training runs. The critic is insensitive to variance in these actions, so action variation does not conflict with high performance.} 
    \label{fig:teaser}
\end{minipage}
  \vskip 0.5cm
  \makeatletter
  \tcbline 
  \vspace{-.7cm} 
\ifdefempty{\metadatalist}{}{\metadatalist\par}
  \makeatother
  
\end{tcolorbox}
\begin{abstract}

Generative control policies (GCPs), such as diffusion- and flow-based control policies, have emerged as effective parameterizations for robot learning. This work  introduces Off-policy Generative Policy Optimization (\OGPO{}), a sample-efficient algorithm for finetuning GCPs that maintains off-policy critic networks to maximize data reuse and propagate policy gradients through the full generative process of the policy via a modified PPO objective, using critics as the terminal reward. \OGPO{} achieves state-of-the-art performance on manipulation tasks spanning multi-task settings, high-precision insertion, and dexterous control. To our knowledge, it is also the only method that can \emph{fine-tune poorly-initialized behavior cloning policies to near full task-success with no expert data in the online replay buffer}, and does so with \emph{few task-specific hyperparameter tuning}. Through extensive  empirical investigations, we demonstrate that \OGPO{} drastically outperforms methods alternatives on policy steering and learning residual corrections, and identify the key mechanisms behind its performance. We further introduce practical stabilization tricks, including success-buffer regularization, two-sided conservative advantages, and Q-variance reduction, to mitigate critic over-exploitation across state- and pixel-based settings. Beyond proposing \OGPO{}, we conduct a systematic empirical study of GCP finetuning, identifying the stabilizing mechanisms and failure modes that govern successful off-policy full-policy improvement.



\end{abstract}


\section{Introduction}
Autonomous acquisition of new skills is an important challenge for modern robot manipulation. While imitation learning via behavior cloning (BC) from human demonstration can enable a robot to learn behaviors across several contexts, performance is typically brittle to subtle changes in tasks and environments. These models rarely exhibit high success rates zero-shot in the diversity of settings encountered in deployment. While this fragility can be remedied through additional data collection, a natural question to ask is - can the robustness of pre-trained imitation learning policies be bolstered autonomously without requiring considerably more manual data collection? 

To this end, there has been a strong interest in finetuning pre-trained robotic policies via reinforcement learning (RL), to autonomously improve behavior via self-collected experience. Of particular relevance is the problem of finetuning \emph{Generative Control Policies} (GCPs) - the parametrization of control policies by expressive generative models, such as diffusion or flow models \citep{chi2023diffusion,black2024pi_0,pan2025much}. These policies have been extremely effective for modern robotic applications \citep{zhang2024affordance, wolf2025diffusion}. 

Current methodology for GCP finetuning succumbs to tradeoffs between data efficiency and the extent of policy improvement during training. Approaches focused on sample efficiency combine \emph{off-policy} critic learning, enabling strong experience reuse,  with either targeted \emph{partial} finetuning of the GCP, such as steering the initial generation noise or learning residual corrections, or instead use behavior cloning to imitate high-return actions. These approaches learn quickly when the base policy has strong coverage of optimal actions, but struggle with exploring new behavior.  
On the other hand, methods focused on eliciting maximum final task performance \citep{lei2025rl,ren2024diffusion,mcallister2025flow}  use \emph{on-policy} policy gradient updates, which drive aggressive policy improvement at the expense of significantly compromised sample efficiency.


In this work, we propose a new algorithm -  \OGPO{} for full-finetuning of expressive GCPs, providing both sample-efficient and expressive policy updates via data-efficient off-policy reinforcement learning. Following \citet{ren2024diffusion,black2023training}, \OGPO{} views GCP optimization as a bi-level MDP, with a nested inner denoising MDP over the action generation steps of a GCP, and an outer environment dynamics MDP over actions actually executed in the environment. Importantly, in real-robotics tasks, there exists a \emph{sample-cost asymmetry}: collecting trajectories from the environment  MDP is expensive, while generating action trajectories through the denoising MDP is purely computational and therefore cheap. 

While direct policy optimization in the unrolled bi-level MDP can be very (environment-)sample inefficient \citep{ren2024diffusion,zhang2025reinflow}, \OGPO{} leverages the aforementioned sample-cost \emph{asymmetry} to perform \emph{decoupled} policy optimization. Specifically, \OGPO{} performs sample-efficient, off-policy Temporal Difference (TD)-learning to learn a Q function in the environment dynamics MDP over \emph{expensive} environment samples, while using data-inefficient but stable on-policy RL updates to extract policies from the inner denoising MDP over \emph{cheap} GCP samples (see \Cref{fig:teaser} left). Doing so allows for an off-policy policy optimization algorithm that is data-efficient (due to TD-learning in the environment dynamics MDP), yet expressive (due to on-policy RL finetuning in the denoising MDP)

Through careful empirical study, we show that \OGPO{} is able to achieve both stable and expressive updates for finetuning GCPs in challenging robotics tasks. Based on empirical analysis of the shortcomings, we further propose \OGPOplus{}, an empirically optimized variant that incorporates improvements in test-time optimization such as Best-of-N planning via Q-functions and policy distillation from successful trajectories obtained via online RL. These improvements allow  \OGPOplus{} to achieve state-of-the-art performance on a set of contact-rich simulation environments with varying horizons, degrees of freedom, and precision requirements, while requiring minimal hyperparameter tuning. Surprisingly, we show that \OGPOplus{} is able to fine-tune policies with \emph{zero expert data} in the policy replay buffer. This is a fundamentally new capability that points towards the future possibility of finetuning models with minimal human data collected in a task-specific manner on deployment. We perform a careful set of analyses to understand the impact of the decoupled optimization central to \OGPO{}, and the impact of the design decision made in \OGPOplus{} - showing the efficacy of full-policy finetuning of GCPs under the right design choices. 

\newcommand{\gammaenv}{\gamma^{\mathrm{env}}}
\newcommand{\statespace}{S}
\newcommand{\actionspace}{A}
\newcommand{\piema}{\pi_{\bar\theta}}

\section{Preliminaries}
\label{sec:prelim}
We formulate our algorithm as a \emph{Markov Decision Process} (MDP) $\mdpenv := (\statespace, \actionspace, P_0, P, R,\gamma)$, with states $s\in\statespace$, actions $a\in\actionspace$, initial state distribution $P_0$, transition probabilities $P$,  reward $R$, and discount factor $\gamma \in (0,1)$. At each timestep $t$, the agent (e.g., robot) observes the state $s_t \in \statespace$, takes an action $a_t \sim \pi(a_t \mid s_t) \in\actionspace$, transitions to the next state $s_{t+1}$ according to $s_{t+1} \sim P(s_{t+1} \mid s_t, a_t)$ while receiving a reward $R(s_t, a_t)$.\footnote{In practice, algorithms may be given incomplete or redundant state observation (e.g., via pixel measurements), in which case we can replace $s$ with an observation $o$. This may violate the Markovian condition in the MDP, but still leads to well-posed algorithms.} For the MDP $\mdpenv$, we let $\Exp^{\pi}$ (resp. $\Pr^{\pi})$ denote the expectation (resp. probability distribution) over trajectories $(s_0, a_0, \dots, s_T, a_T)$ with length $T+1$, with initial state distribution $s_0 \sim P_0$ and transition operator $P$. We train a policy to optimize the cumulative discounted return $\Jcost(\pi_\theta) = \mathbb{E}^{\pi_{\theta}}\!\left[\sum_{t\ge 0} \gamma^t R(s_t,a_t)\right]$. We also recall the Q-function 
\begin{equation}
Q^{\pi}(s,a) := \Exp^{\pi}[\sum_{t \ge 0} \gamma^t R(s_t,a_t) \mid (s_t,a_t) = (s,a)]
\end{equation}
and value function $V^\pi(s) := \Exp_{a \sim \pi(s)}[Q^\pi(s,a)]$. We apply action chunking \citep{zhao2023learning}, where sequences of actions $a_{t:t+h-1}$ are predicted and executed in open-loop. For simplicity, we treat each action chunk as a single action in $\mdpenv$, thereby preserving the standard MDP notation. Thus, for the rest of the paper, \textbf{$a_t$ refers to an entire action-chunk}, and rewards are adjusted appropriately (see \Cref{app:action_chunk} for how).



\colorpar{On-Policy Policy Gradient Methods.} \emph{Policy gradient} (PG) methods (e.g., REINFORCE \citep{williams1992simple}) improve policy performance by approximating the gradient of this objective w.r.t. the policy parameters:
\begin{equation}
   \nabla_\theta J(\pi_\theta) = \mathbb{E}^{\pi_{\theta}} \!\left[\sum_{t \ge 0} \nabla_\theta \log \pi_\theta(a_t \mid s_t)\, r_t(s_t, a_t)\right],
\end{equation}
where $r_t(s_t, a_t) := \sum_{\tau \ge t} \gamma^\tau R(s_\tau, a_\tau)$ is the discounted future return from time $t$,  and $\nabla_\theta \log \pi_\theta(a_t | s_t)$ denotes the gradient of the logarithm of the \emph{likelihood} of $(a_t\mid s_t)$. Myriad improvements exist to reduce variance of gradient estimation and accelerate training stability; following \citep{ren2024diffusion,zhang2025reinflow}, we  build on the  PPO algorithm \citep{schulman2017proximal}. PG methods are called \emph{on-policy} because they optimize over the \emph{current} policy distribution, limiting data re-use and sample efficiency.


\vspace{-.5em}
\colorpar{Off-Policy Reinforcement Learning.}  \emph{Off-policy RL methods} maintain a long horizon replay buffer $\rb=\{(s_t,a_t,s_{t+1},r_t,d_t)\}$ consisting of past states $s_t$, actions $a_t$, subsequent states $s_{t+1}$ from the environment transitions, the observed rewards $r_t$, and the done signal $d_t$.  The buffer is used to train an ensemble of $M$ critic networks $Q_{\phi_i}:S \times A \to \R$, with parameters $\phi_1,\dots,\phi_{M}$, such that $Q_{\phi_i}(s_t,a_t)$ evaluates the expected cumulative return $Q^{\piema}(s_t,a_t)$ of action $a_t$ at state $s_t$ under a current \emph{target policy} $\piema$. 
The critic networks are updated in parallel using the {temporal difference loss}, which enforces the Bellman consistency equation defined by $Q$-functions: 

\iftoggle{arxiv}{}{\vspace{-2em}}
\begin{equation}
    \Lcritic(\phi) = \Exp 
\Big[Q_{\phi}(s_t,a_t) - \Big(r_t + \gamma \cdot \Qtarg(s_{t+1},a_{t+1})\Big)\Big]^2 
\label{eq:Lcrtic},
\end{equation}

\vspace{-0.2em}
where above the expectation $\Exp$ is taken over $(s_t,a_t,r_t,s_{t+1} \sim \Buffer)$ sampled from the replay buffer $\Buffer$, and  each $a_{t+1}$ is sampled independently from the current target policy $\piema(\cdot \mid s_{t+1})$. To avoid overestimation bias, we set $\Qtarg(s,a) = \frac{1}{M} \sum_{i}Q_i$ to be a mean over critic networks, described in the Appendix (\Cref{app:td_loss}). \citep{fujimoto2018addressing, chen2021randomized}.Importantly, \eqref{eq:Lcrtic} enables
 data collected by policies from previous training epochs, thereby increasing sample efficiency. 

\newcommand{\pibarddpm}{\Bar{\pi}^{\textsc{ddpm}}}
\newcommand{\pibarflow}{\Bar{\pi}^{\textsc{flow}}}
\newcommand{\sbar}{\Bar{s}}
\newcommand{\pibar}{\bar{\pi}}

\togglepar{Generative Control Policies.}   Current robotic control policies use generative models as parameterizations of control policies. Following \citep{pan2025much}, we call these generative control policies (GCPs). 
\GCP s represent a stochastic policy $\pi_\theta(\cdot\mid s)$ as a series of iterative computation steps, defined by a mapping $\piigp: S \times A \times \mathbb{N}$. Given a state $\st$, the policy first samples $\afirstt \sim \piigp\prn{\cdot \mid \aitert = \emptyset, \igpiter=\igpfirst,s_t}$ where $k$ is a GCP timestep. Next, we sample $\ainct \sim \piigp\prn{\cdot \mid \aitert,\igpiter,s_t}$ which leads to is an action $\alastt$. We  compactly denote the distribution of this action given the observation as $\alastt \sim \pi_\theta(\cdot \mid s_t)$, turning the GCP into a standard policy. 
Our iteration conventions are \emph{decreasing} in $K$, those in diffusion models. Following the same conventions, we also refer to the index $k$ as the ``denoising step.''

\togglepar{Flow-Based GCPs.} We focus on a popular class of GCPs: {flow-based} control policies \citep{black2024pi_0}. As discussed in \cref{app:igp_instantiations},  our methods and baselines can also be instantiated with Diffusion-based policies \citep{chi2023diffusion} and  other controller parameterizations \citep{pertsch2025fast,frans2024one,pan2025much} .
Flow policies are pretrained using the flow-matching objective: given training pairs $(s,a)$, we sample noise $z \sim \Normal(0,\eye)$. With a continuous noise index $\tau \in [0,1]$, we define an interpolated action $a_{(\tau)} = \tau + (1-\tau)z$, and optimize a velocity field  $v_\theta(a_{(\tau)},\tau;s)$ by minimizing $\Exp_{(s,a,\tau)}\|v_\theta(a_{(\tau)},\tau;s) - (a-z)\|^2$ \citep{albergo2023stochastic,lipman2022flow} . 
Inference is performed by discretizing an ordinary differential equation (ODE) which reverses the noising process $
    \ainct  := \aitert + \frac{1}{K}{v_\theta(\aitert,k/K,s)} $, with $a_{t,0} \sim \Normal(0,\eye)$.   


\section{{O}ff-{P}olicy {G}enerative {P}olicy {O}ptimization}

\label{sec:method}
We propose \textbf{O}ff-\text{P}olicy \textbf{G}enerative \textbf{P}olicy \textbf{O}ptimization, \OGPO, an off-policy full-policy finetuning method for generative control policies. We begin by introducing the basic algorithm, and then describe an improved variant, \OGPOplus. We provide summary pseudocode in \Cref{alg:ogpo_small}, and defer full implementation details to \cref{app:pseudo}. 

\togglepar{Background: Off-Policy Policy Extraction.} Given a replay buffer $\cB = \{(s,a,s',r)\}$, traditional off-policy RL methods consist of two steps: (1) fitting Q- functions via a TD-update \Cref{eq:Lcrtic},  (2) performing policy extraction by choosing actions that maximize the target Q function $\Qtarg$ as a surrogate of future return:
\begin{align}
  \theta \in \argmax_\theta \Exp_{a \sim \pi_\theta(s)}(\Qtarg(s,a))\label{eq:policy_extrac_general}.
\end{align}
The replay buffer facilitates off-policy data-reuse for training $\Qtarg$ (typically via \eqref{eq:Lcrtic}), driving sample efficiency, whereas \eqref{eq:policy_extrac_general} can be computed purely computationally. A historically popular approach to optimize this objective for simple policy parametrization, like Gaussian policies, is the so-called \emph{reparameterization trick} \citep{kingma2013auto,figurnov2018implicit}, where  a stochastic policy is rendered as  $\pi_\theta(s;w)$ for a noise $w$ drawn from a fixed (non-learned) distribution. From here, \Cref{eq:policy_extrac_general} is written as an expectation over $w$, the algorithm directly differentiates $\Qtarg(s,\pi_\theta(s,w))$ with respect to $\theta$ under samples $w$. In principle, the same can be done for GCPs such as flow-policies, sampling an initial noise $a_{t,K}$ and backpropagating through the inference chain (\Cref{fig:pos_neg_grads}, center). However, as we show experimentally (\Cref{app:bptt_vs_ogpo}), doing so leads to an exploding gradient problem as we differentiate through the multiple flow steps. Moreover, it requires differentiating  $\nabla_a \Qtarg$, which can be inaccurate in contact-rich tasks \citep{suh2022differentiable}.

\togglepar{\OGPO: On-Policy PPO  for Off-Policy Policy Extraction.} \OGPO{} is designed for applications, such as robotic manipulation, where environment interactions are more costly than computation, and where action gradients with respect to $\Qtarg$ are noisy or inaccurate \citep{suh2022differentiable}. We maintain off-policy critic learning that facilitates data reuse, and propose a \emph{fully parallelizable zero-order optimizer} that solves \Cref{eq:policy_extrac_general}, avoiding  backpropagation through the denoising chain and differentiation with respect to the target network. 
\begin{figure}[H]
    \centering
    \includegraphics[width=0.9\linewidth]{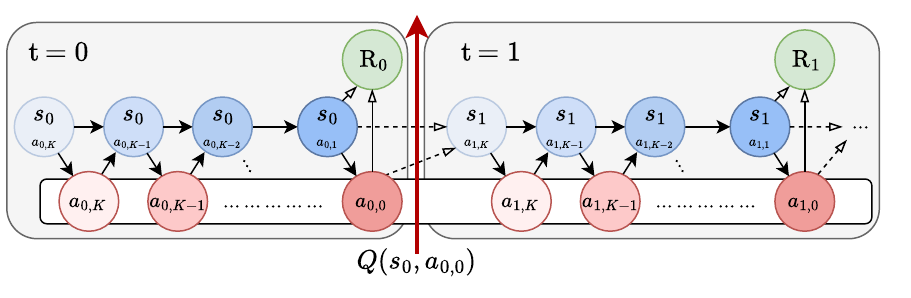}
    \caption{We recall the bi-level MDP from \citep{ren2024diffusion}, which embeds action-level trajectories into the environmental dynamics. \OGPO{} truncates this MDP at the end of each denoising trajectory, using Q-values as a terminal, action-trajectory-level reward, enabling off-policy policy extraction via {on-policy} policy optimization.}
    \label{fig:bi_level_mdp}
\end{figure}

\vspace{-0.5em}
Our starting point is the bi-level MDP formulation adopted from \citep{ren2024diffusion} (\Cref{fig:bi_level_mdp}). Following \citep{black2023training}, we view sequences $a_{t,K:0} = (a_{t,K},\dots,a_{t,0})$ as trajectories in an \emph{denoising} MDP, where time is indexed by denoising step $k$, and state and action at step $k$ are $a_{t,k}$ and $a_{t,k-1}$, respectively. \citet{ren2024diffusion} embeds this action-level MDP into the environment-level MDP $\mdpenv$, resulting in an \emph{bi-level} MDP where states are $\bar{s}_{t,k} = (s_t,a_{t,k})$, the actions are $a_{t, k-1}$ , and the indices $(t,k)$ are lexicographically increasing in $t$ and decreasing in $k$. \Cref{fig:bi_level_mdp}  depicts this bi-level MDP:  transitions within each gray block occur within the denoising-level MDP, and between gray blocks are transitions in $\mdpenv$; see \Cref{app:bilevel} for further details. The \DPPO{} algorithm proposed by \citep{ren2024diffusion} then applies on-policy PPO at the level of this bi-level MDP. Whilst avoiding the aforementioned pathologies associated with backpropagation, this method gives up the sample efficiency afforded by off-policy critic learning.


Our \textbf{key insight} is that  denoising-trajectories can be generated purely \emph{computationally} from policy inference, as they occur in the ``imagination'' of the GCP. We can then use critic learning to sever the bi-level MDP just before environment-MDP state transitions ({\color{\OGPOcolor} red line}, \Cref{fig:bi_level_mdp}), enabling zero-order optimization applied only to the denoising-level MDP. As compared to backpropagation approaches to solving \Cref{eq:policy_extrac_general},  our approach avoids (i) backpropagation through time and (ii) differentiating through the Q-function. Moreover, as compared to pure on-policy zero order optimization through the bi-level MDP \citep{ren2024diffusion}, our zero-order updates are (i) performed purely computationally, in the ``imagination'' of the denoising process (ii) fully parallelized across large batch sizes   (iii) used to optimize a critic network, facilitating full reuse of environment-level trajectories. Moreover, (iv) the   problem horizon of the denoising-level MDP scales only with the denoising steps $K$, and not $K\times \text{(task horizon)}$.
Concretely, we apply the PPO algorithm \citep{schulman2017proximal}, a zero-order policy gradient method, to optimize over  the denoising MDP. 
 Given state $s_t$, denoising trajectory $a_{t,K:0}$, and baseline value estimate $\hat V$,  we apply the standard PPO loss  \emph{only} to the denoising trajectory $a_{t,K:0}$:
\begin{align}
&\ell_{\texttt{PPO}}(\theta; s_t,a_{t,K:0},\hat \mu):=\min(\ratio \hat A,\text{clip}(\ratio, 1-\epsilon, 1+\epsilon) \hat A) \nonumber \\
&~\ratio := \prod_{k=1}^K\frac{ \pi_\theta(a_{t,k-1} \mid s_t,a_{t,k})} {\piema(a_{t,k-1} \mid s_t,a_{t,k})}, ~ \hat A = \Qtarg(s_t,a_{t,0}) - \hat V 
\label{eq:ppo_internal}.
\end{align}

\newcommand{\Nbatch}{N_{\texttt{batch}}}
\newcommand{\Ngroup}{N_{\texttt{group}}}

\iftoggle{arxiv}
{
\begin{figure}[t!]
    \centering
    \includegraphics[width=0.99\linewidth]{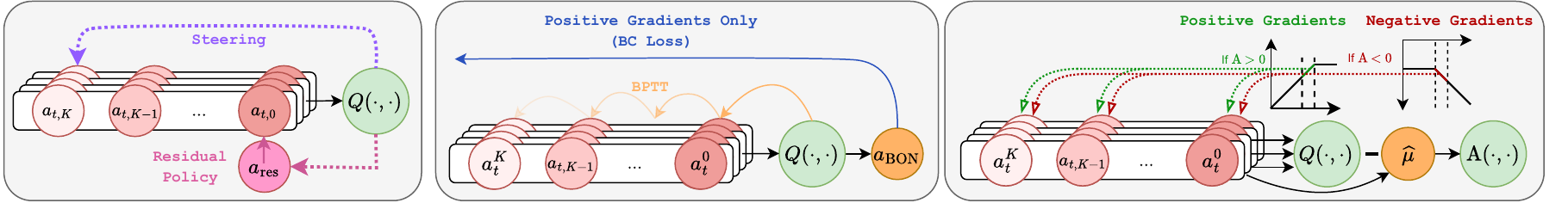}
    \caption{Visual depiction of the different off-policy RL algorithms. \textbf{(left)} \DSRL{} trains an initial noise steering policy, while \EXPO{} trains a residual policy to modify final GCP actions. \textbf{(center)} \QC{} drives policy improvement via supervised finetuning (SFT) of Best-of-N actions ranked via the critic, while \BPTT{} backpropagates the gradients through the entire GCP. \textbf{(right)} \OGPO{} uses an ensemble of critics to compute $\advgrpo$ (\cref{eq:ppo_internal}) that update the GCP via Annealed Importance Sampling, thereby directly conditioning the log-likelihoods over the GCP chain.}
    \label{fig:pos_neg_grads}
\end{figure}
}
{
\begin{figure*}[t]
    \centering
    \includegraphics[width=0.99\textwidth]{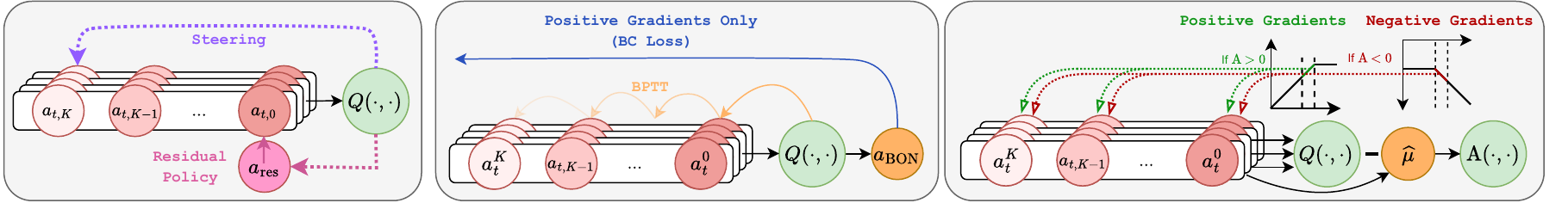}
    \caption{Visual depiction of the different off-policy RL algorithms. \textbf{(left)} \DSRL{} trains an initial noise steering policy, while \EXPO{} trains a residual policy to modify final GCP actions. \textbf{(center)} \QC{} drives policy improvement via supervised finetuning (SFT) of Best-of-N actions ranked via the critic, while \BPTT{} backpropagates the gradients sequentially through the entire GCP. \textbf{(right)} \OGPO{} uses an ensemble of critics to compute $\advgrpo$ (\cref{eq:ppo_internal}) that update the GCP via Annealed Importance Sampling, thereby directly conditioning the log-likelihoods over the GCP chain.}
    \label{fig:pos_neg_grads}
\end{figure*}
}

\togglepar{Multiple Denoising-Trajectory Sampling.}  
\label{sec:grpo}
Because denoising-trajectories are generated computationally, they can be resampled \emph{fully in parallel} from \emph{any} given state $s_t$ in the replay buffer. Moreover, $\Qtarg$ can be evaluated without taking a single transition step in the environment. Taking advantage of this, we evaluate our PPO loss over an average of a batch of parallel-sampled trajectories,  purely in the ``imagination'' of the GCP. 
By analogy to policy optimization in large language models (LLMs), we can think of a state $s_t$ in the buffer as a ``context'' and the denoising trajectory $a_{t,K:0}$ as a ``response''. We draw inspiration from the GRPO algorithm \citep{shao2024deepseekmath} in LLM post-training, where multiple responses are sampled in parallel from a given prompt, and gradients are averaged together to reduce gradient variance.\footnote{GRPO includes an additional variance normalization term, which we omit.} In \OGPO, at each update, we sample $\Nbatch$ states $(s^{(i)})_{1 \le i \le \Nbatch}$ from our replay buffer, and  sample $\Ngroup$  denoising trajectories $(a_{K:0}^{(i,\aindex)})_{1 \le \aindex \le \Ngroup}$ drawn i.i.d. from $\piema(\cdot \mid s^{(i)})$ per state. We then update via the loss
\begin{align}
    \hat L_{\texttt{PPO}}(\theta) = \frac{1}{N_{\texttt{tot}}} \sum_{i}^{}\sum_{\aindex}^{}\ell_{\texttt{PPO}}(\theta; s^{(i)},a_{K:0}^{(i,\aindex)},\hat V^{(i)}). \label{eq:PPO_loss_group}
\end{align}
\Cref{eq:PPO_loss_group} averages  {both} over the states $s_t^{(i)}$  from the buffer (``prompts''), and denoising-trajectories generated in parallel from each given state (``responses'').  This yields a normalization factor of $N_{\texttt{tot}} := \Nbatch\cdot \Ngroup$. Moreover, parallel sampling facilitates estimating the value baseline via a direct Monte-Carlo approximation $\hat V^{(i)} \gets \frac{1}{\Ngroup} \sum_{j} \Qtarg(s^{(i)},a_{0}^{(i,\aindex)})$, obviating the need to learn a separate value-prediction network. 

\togglepar{Debiasing Noise Injection for Flow Policies.} We instantiate \OGPO{} for flow-based policies. To evaluate the likelihood $\ratio$ in \Cref{eq:ppo_internal}, we must ensure the denoising-level action {likelihoods}  $a_{k-1,t} \mid a_{k_t},s_t$ are non-singular.  ReinFlow \citep{zhang2025reinflow} modifies the bi-level PPO algorithm of \citep{ren2024diffusion} for flow-based policies, achieving this by adding additional Gaussian noise to each flow step.  For given choice of noise levels $\sigma_k^2$, this yields the following inference procedure: 
\begin{equation}
    \textstyle\ainc \sim \pibarflow(\cdot\mid\aiter,k,s) := \Normal( v_\theta(\aiter,\frac{k}{K},s), \sigma^2_k \eye)    
\end{equation}
In \OGPO, we anecdotally observe that naively adding noise can degrade policy performance by changing the marginal distributions of actions $a_{t,k}$ generated during denoising. We therefore introduce a correction proposed by \citet{albergo2023stochastic} which (in the infinite step limit) ensures the per-denoising-step marginal distributions of noise-augmented actions match those of standard flow sampling; see also \cite{liu2025flow}. See  \Cref{app:ode_correction} for details.


\newlength{\algindent}
\setlength{\algindent}{1.5em}
\newcommand{\INDSTATE}{\STATE \hspace{1.5em}}
\newcommand{\STATEX}{\item[]}

\begin{algorithm}[h]
\caption{\OGPO{} (Abbreviated)}
\label{alg:ogpo_small}
\begin{algorithmic}[1]
\FOR{each environment step until \done}
    \STATE Execute   $a_t \sim \piema(\cdot \mid s_t)$,  and update buffer  $\cB \gets (s_t,a_t,r_t,s_{t+1},\done)$.
    \STATEX \algcomment{\small \%   Standard Critic Update}
    \STATE Update critic networks  $\phi_1,\dots,\phi_M$ using empirical TD Error \eqref{eq:Lcrtic} over $\cB \sim \rb$.
    \STATEX \algcomment{\small \%   Actor Update via Multiple Denoising Trajectories}
    \FOR{$i = 1,\dots,\Nbatch$}
    \STATE Sample state $s^{(i)}$ from $\cB$, and  action trajectories $a_{K:0}^{(i,j)}\sim \piema(\cdot \mid s^{(i)})$ for $1 \le j \le \Ngroup$.
    \STATE Estimate value baselines via  $\hat V^{(i)} \gets \frac{1}{\Ngroup} \sum_{j} \Qtarg(s^{(i)},a_{0}^{(i,j)})$
        \ENDFOR
        \STATE Update actor using aggregated PPO loss \eqref{eq:PPO_loss_group}
        \STATEX \algcomment{\% EMA parameters}
        \STATE Update target parameters $\bar \theta \gets (1-\tau)\bar \theta + \tau \theta$, $\bar \phi_i \gets (1-\tau) \bar \phi_i + \tau\phi_i$. Set $\Qtarg= \frac{1}{M}\sum_i Q_{\bar\phi_i}$.
\ENDFOR
\end{algorithmic}
\end{algorithm}

\begin{AIbox}{\OGPOnc{}: Core Insights}
\begin{itemize}[leftmargin=2.25mm, itemsep=2pt]
    \item \textbf{Off-Policy Q-learning for On-Policy GCP Extraction.} \OGPO{} severs the bi-level MDP at environment transitions, using $\Qtarg$ as a terminal reward for the GCP's denoising MDP. 
    \item \textbf{$0^\mathrm{th}$-order optimization.} PPO-style zeroth order optimization over denoising trajectories avoids first order backpropagation through $\Qtarg$ and the GCP chain. This simplifies RL-finetuning in high-Lipschitz tasks.
    \item \textbf{Debiased noise injection.} SDE-augmented flow steps yield non-singular likelihoods for the PPO ratio $\ratio$, with the stochastic interpolant correction ensuring marginal distributions match standard ODE inference. All $\Nbatch \times \Ngroup$ trajectories are sampled and scored in parallel.
\end{itemize}
\end{AIbox}


\section{ Improving \OGPO{} by Mitigating Critic Over-exploitation}
\label{sec:exploit}
In this section, we identify a major limitation of \OGPO: over-exploitation of learned critics due to highly expressive policy updates (\Cref{sec:unreg_ogpo}) . We then introduce two modular modifications to \OGPO{} which overcomes this tendency, and which are mutually compatible:
\begin{itemize}
    \item \OGPOplus{} (\Cref{sec:ogpoplus}), which combines \OGPO{} with behavior cloning regularization on \emph{success-only} trajectories
    \item \OGPOplusca{} (\Cref{sec:reg_ogpo_dip}), which uses a  \emph{conservative advantage} for policy extraction, thereby \emph{drastically mitigating the ``dip'' in offline-to-online adaptation}
\end{itemize}
\newcommand{\Nvr}{N_{\mathrm{vr}}}
\newcommand{\Lcriticvr}{L_{\mathrm{critic},\mathrm{vr}}}

For a practitioner, we recommend using $\OGPOplusca$ for highly stable policy extraction. In addition to these modifications, we optionally incorporate $Q$-variance reduction that averages the TD targets over $\Nvr$ actions sampled from the reference actor, thereby improve critic accuracy:
    \begin{align}
    \Lcriticvr(\phi) = \Exp 
\Big[Q_{\phi}(s_t,a_t) - \Big(r_t + \gamma \cdot \frac{1}{\Nvr}\sum_{i=1}^{\Nvr}\Qtarg(s_{t+1},a_{t+1}^{(i)})\Big)\Big]^2 
\label{eq:Lcrtic_vr}, \quad a_{t+1}^{(i)}\iidsim \piema(\cdot \mid s_{t+1}).
\end{align}
We find that \Cref{eq:Lcrtic_vr} always yields some improvement, but this benefit is  most pronounced in pixel-based environments  \RETURNN{for low-data runs}; 
it makes a marginal difference in state-based runs on pre-training from full data.  


\colorpar{Experimental Setup:} 
To identify improved design choices, we consider experiments on the
 state-based and image-based \Robomimic{} tasks, which are described in greater detail in \Cref{sec:exp_setup}. For state-based runs, we use all state-information directly; for image based runs, we pass image observations  using a frozen PaliGemma VLM backbone from the pre-trained $\pifive$ VLA \citep{pifive}.  Further details are given in \Cref{app:envirs}. 


\begin{figure}[t]
    \centering

    \begin{subfigure}[t]{0.32\linewidth}
        \centering
        \includegraphics[width=\linewidth]{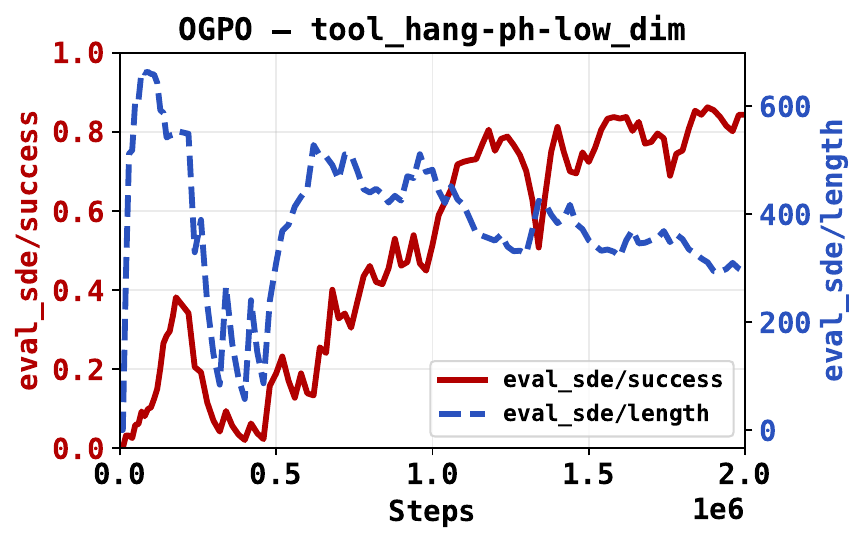}
        \caption{}
        \label{fig:succ_vs_len_ogpo}
    \end{subfigure}
    \hfill
    \begin{subfigure}[t]{0.32\linewidth}
        \centering
        \includegraphics[width=\linewidth]{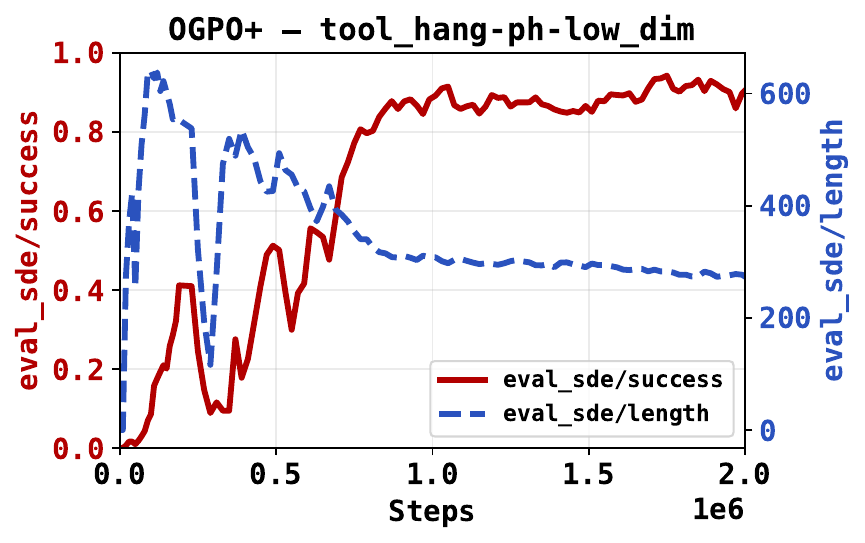}
        \caption{}
        \label{fig:succ_vs_len_ogpoplus}
    \end{subfigure}
    \hfill
    \begin{subfigure}[t]{0.32\linewidth}
        \centering
        \includegraphics[width=\linewidth]{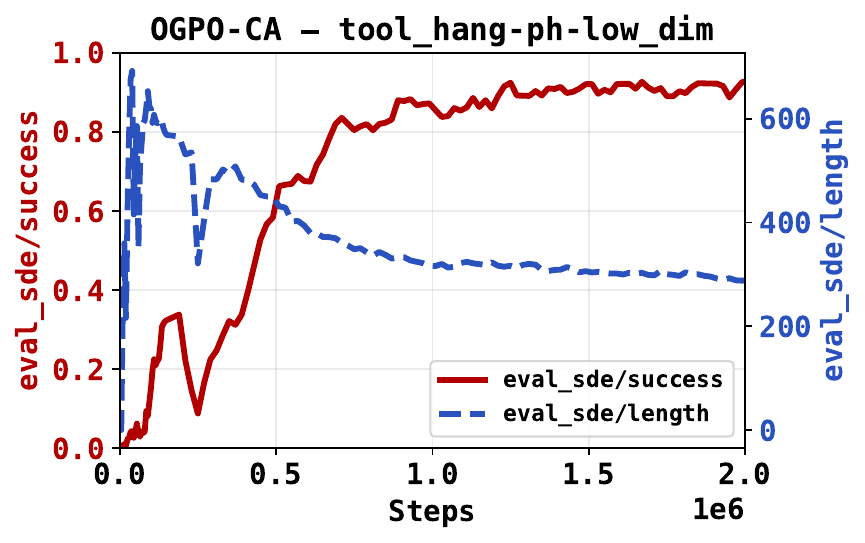}
        \caption{}
        \label{fig:succ_vs_len_ogpoca}
    \end{subfigure}
    \begin{subfigure}[t]{0.32\linewidth}
        \centering
        \includegraphics[width=\linewidth]{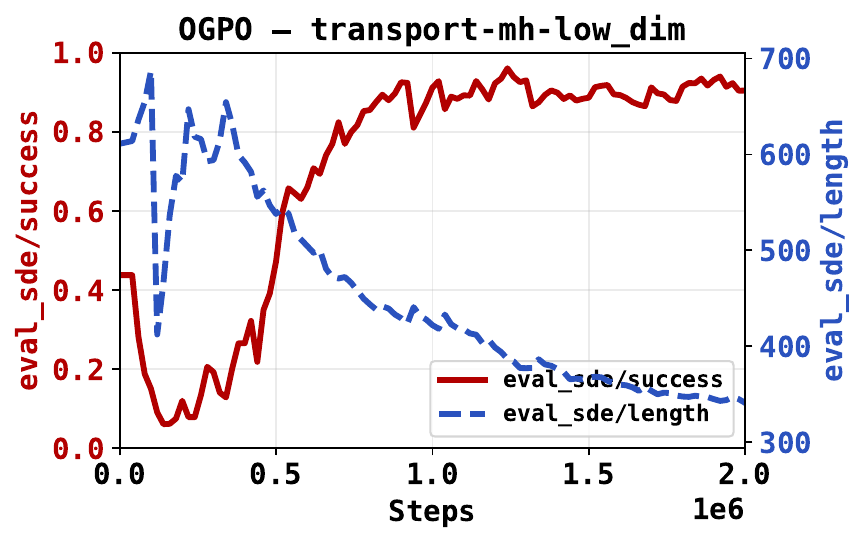}
        \caption{}
        \label{fig:succ_vs_len_transport_ogpo}
    \end{subfigure}
    \hfill
    \begin{subfigure}[t]{0.32\linewidth}
        \centering
        \includegraphics[width=\linewidth]{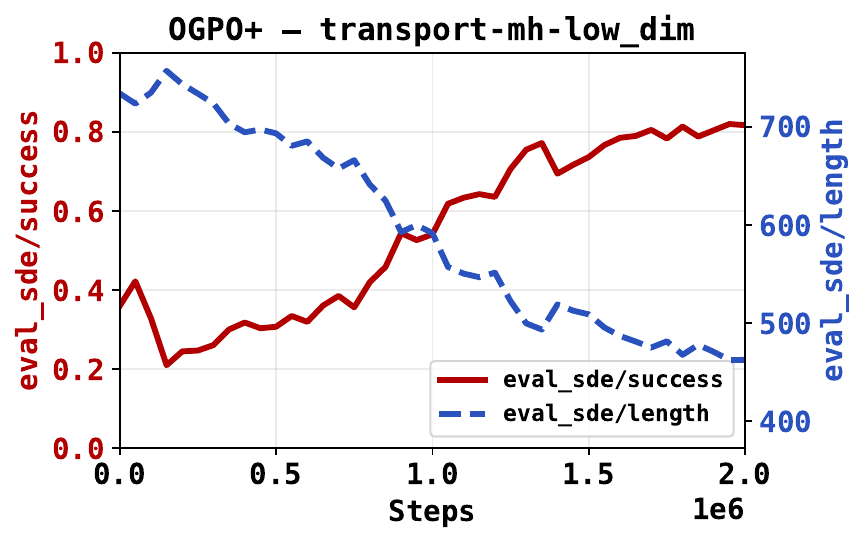}
        \caption{}
        \label{fig:succ_vs_len_transport_ogpoplus}
    \end{subfigure}
    \hfill
    \begin{subfigure}[t]{0.32\linewidth}
        \centering
        \includegraphics[width=\linewidth]{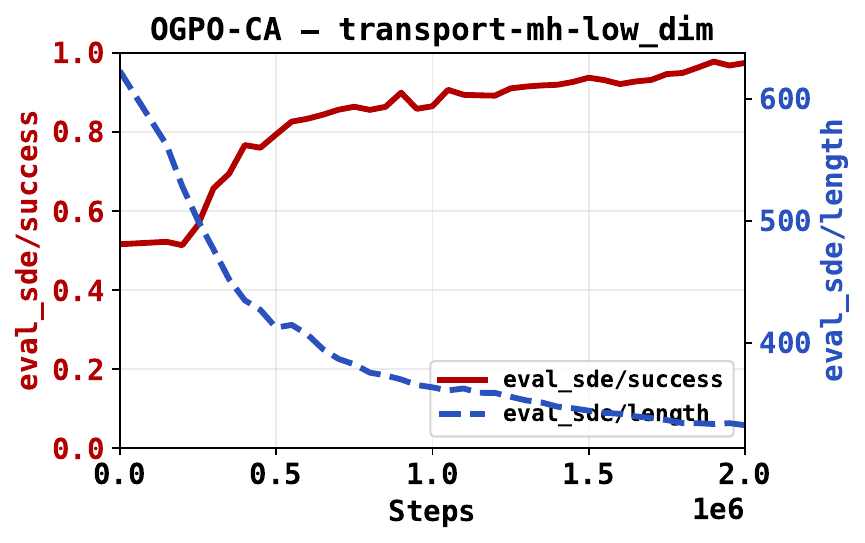}
        \caption{}
        \label{fig:succ_vs_len_transport_ogpoca}
    \end{subfigure}

    \caption{The above plots show the full training comparison between (a) Vanilla \OGPO{}, (b) \OGPOplus{}, and (c) \OGPOplusca{}, on state-based \Robomimic{} tasks. The red axis shows success rate and the blue axis shows the mean length of \emph{successful} trajectories.By aggressively maximizing sparse reward, \OGPO{}  optimizes for both task success rate, and \emph{completion in few steps}. Without further regularization, the two can be in tension, causing a sharp initial decrease in the length of the policy rollouts,   subsequent oscillations in success rates (\rmtoolhang{}, (a) a high-precision task) or plateaus in performance (\rmtransport{}, (d) a long-horizon task) . By adding a success buffer, 
    \OGPOplus{} bias the policy learning objective to favor task success (b, e)). 
    \OGPOplusca{} mitigates the effect of outliers in the critic estimation, thereby fitting the ``dip'' between offline BC and online training (c, f).}
    \label{fig:pos_neg_grads}
\end{figure}

\subsection{Vanilla \OGPO{} Over-exploits  Imperfectly Learned Critics }\label{sec:unreg_ogpo}
Recall that \OGPO{} makes PPO-style updates to the denoising MDP. The combination of the expressive generative policies and PPO updates on the full-denoising trajectory risks causing OGPO to over-optimize the critic, overfitting to advantages which are poorly estimated. 

\colorpar{Success-Speed Tradeoff.} The typical ``sparse-reward'' manipulation setting assigns reward of $-1$  each time step a task remains uncompleted. Thus, minimizing cumulative reward introduces a tension between  completion \emph{rate} and completion \emph{speed}. 
As a result, \OGPO{}  may  attempt to finish tasks too quickly, causing success rates to drop, harming future exploration training stability. This success-speed tradeoff is visible in \Cref{fig:succ_vs_len_ogpo}, where we see average task length rapidly improves, but success rate plateaus. Anecdotally, we found that the variance-reduced critic update \Cref{eq:Lcrtic_vr} did not improve this tradeoff.

\colorpar{Overexploitation is Exacerbated in Pixel-Based RL.} We observe that vanilla \OGPO{} has more severe exploitation in pixel-based settings. We consider a \Robomimic{} \rmsquare{} environment described above, 
\begin{wrapfigure}{r}{0.4\textwidth}
\vspace{-1em}
\includegraphics[width=0.4\textwidth]{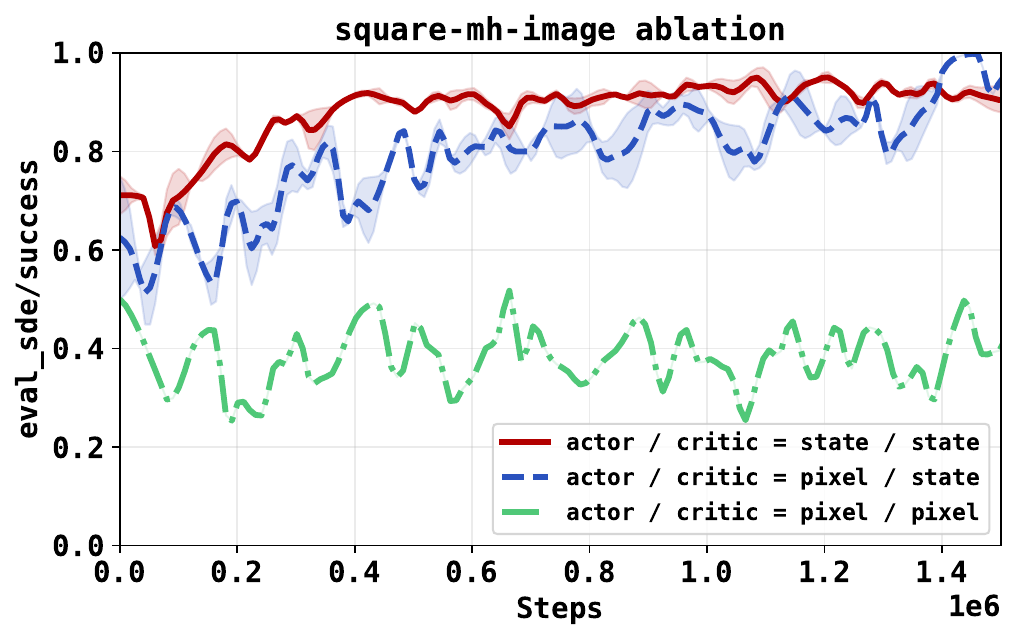}
\vspace{-1.5em}
    \caption{On ablating actor and critic observation modalities, we observe that vanilla \OGPO{} fails to improve policy performance from image-based critics. }
    \label{fig:actor_critic_obs_abl}
\end{wrapfigure}  
where pixels are feautrized using a frozen PaliGemma VLM backbone from $\pifive$. To isolate the effects 
of pixels, we compare four variants: (1) state-based actor/state-based critic; (2) pixel-based actor/state-based critic (3) pixel-based actor/pixel-based (4) state-based actor/pixel-based critic. We plot variants (1-3) in \Cref{fig:actor_critic_obs_abl}, and omit (4) due to collapsing runs. We that the policy trains effectively for both state-based critic runs (1)\&(2), but fails on (3)\&(4), suggesting that \emph{pixel-based} critics prevent learning. We hypothesize that such critics learn less accurately due to the richer observation space, making them more susceptible to exploitation via OGPO. Anecdotally, we found that the variance-reduced critic update made modest but very limited improvements to the pixel-based critics, suggesting the need for further interventions.


\subsection{\OGPOplus{}: Regularizing \OGPO{} With Behavior Cloning of Successful Trajectories}
\label{sec:planning}
\label{sec:ogpoplus}


\newcommand{\Qbon}{Q_{\textsc{BoN}}}


To remedy critic overexploitation, \OGPOplus{} incorporates a regularization term applied only to actions from successful trajectories. This biases policy improvement toward replicating only the actions that led to success \citep{oh2018self}. Specifically, we maintain a \emph{success buffer} $\succb \subseteq \rb$ containing transitions from episodes that achieve task success. During training, we sample mini-batches from $\succb$ and compute 
\begin{equation}
    \Lbc(\theta) = \mathbb{E}_{(\stsucc, \atsucc) \sim \succb} \left[ \textsc{BcLoss}(\bar{\pi}_\theta(\cdot \mid \stsucc), \atsucc) \right]
\end{equation}
\begin{wrapfigure}{r}{0.4\textwidth}
\vspace{-1em}
\leavevmode
    \vspace{-1em}
    \includegraphics[width=0.4\textwidth]{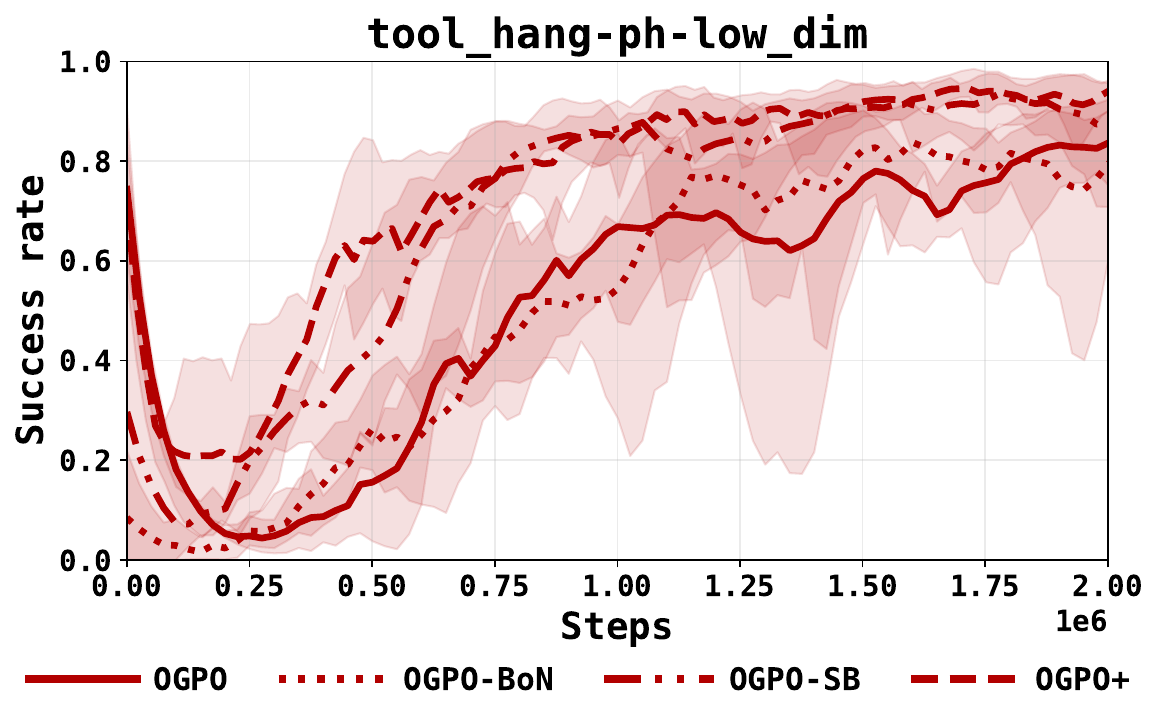}
    \caption{We perform a small sweep of ablations adding Best-of-N (BoN) Inference and Success Buffer on \Robomimic{}\rmtoolhang{}.}
    \label{fig:ogpo_ogpo_bon_ogpo_plus}
\end{wrapfigure}
where $\textsc{BcLoss}$ is the appropriate behavior cloning objective (e.g., denoising score matching for diffusion policies, or flow matching loss for flow policies). Success-imitations ground the policy toward known good actions, while the PPO objective more aggressively explores improvements. For \OGPOplus, the total policy loss combines both terms:
\begin{equation}
    \label{eq:OGPOpl}\Ltotal(\theta) = \Lppo(\theta) + \lambda_{\textsc{bc}} \Lbc(\theta).
\end{equation}

\RETURNN{Though the success buffer is added to prevent over\emph{optimistic} exploitation of critics, it also facilitates learning in limited data-regimes by combating critic under-estimation when successes are infrequent (\REF).}

\colorbold{(Optional) Best-of-N Inference.}  In many domains, such as language modeling, evaluating the quality of an action, or ``verification'' is learned more quickly and accurately than ``generation'' of good actions. This verification-generation gap \citep{setlur2025e3} motivates the popular practice of Best-of-$N$ sampling \citep{brown2024large}, where one generates multiple proposal actions, and selects the best using a learned verifier. 

Best-of-$N$ sampling  has seen widespread adoption in RL training of robotics policies \citep{mark2024policy,dong2025expo,li2025reinforcement}, using the target critic as verifier. In,  \OGPOplus{} we do the same with a slightly modified  critic $\Qbon$ described in \Cref{app:td_loss}. We remark that, due to the aggressive policy extraction, Best-of-$N$ inference yields only \textbf{marginal additional} performance; the success buffer, as described above, is crucial. Thus, we recommend \emph{omitting} Best-of-$N$ when inference cost is constrained. 
\iftoggle{arxiv}
{
\begin{align}
   \abont &:= \argmax\{ \Qtarg(\st, \alastt^{(i)}) : \alastt^{(1)},\dots, \alastt^{(N)} \iidsim \pi_{\thetaema}(\cdot \mid \st)\}. \label{eq:a_plan}
\end{align}
}
{
\begin{align}
\begin{aligned}
   &\aplant \gets \abont := \argmax\{ \Qbon(\st, \alastt^{(i)}) \\ &~~ \mid \alastt^{(1)},\dots, \alastt^{(N)} \iidsim \piema (\cdot \mid \st)\}.
\end{aligned}
\label{eq:a_plan}
\end{align}
}

\subsection{\OGPOplusca: Mitigating the Offline-to-Online Performance Dip via Conservative Advantages}
\label{sec:reg_ogpo_dip}

\begin{figure}[H]
    \centering
    \begin{subfigure}[t]{0.49\linewidth}
        \centering
        \includegraphics[width=\linewidth]{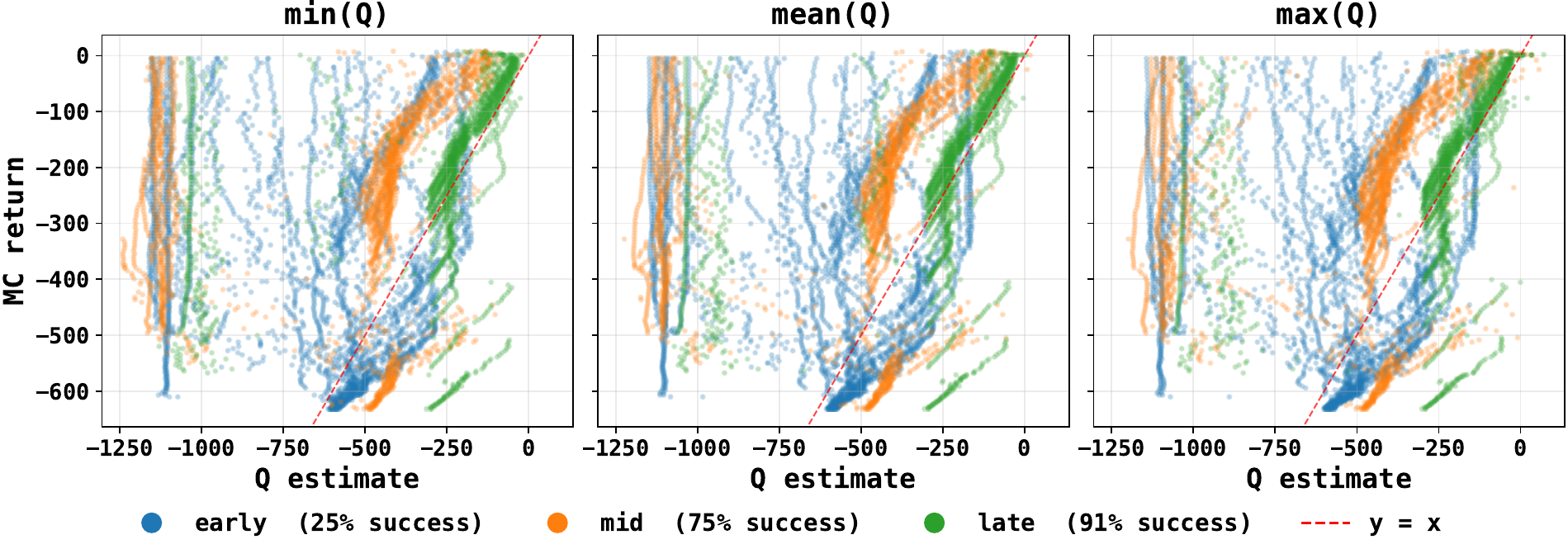}
        \caption{\OGPO{}}
        \label{fig:q_vs_mc_ogpo}
    \end{subfigure}
    \begin{subfigure}[t]{0.49\linewidth}
        \centering
        \includegraphics[width=\linewidth]{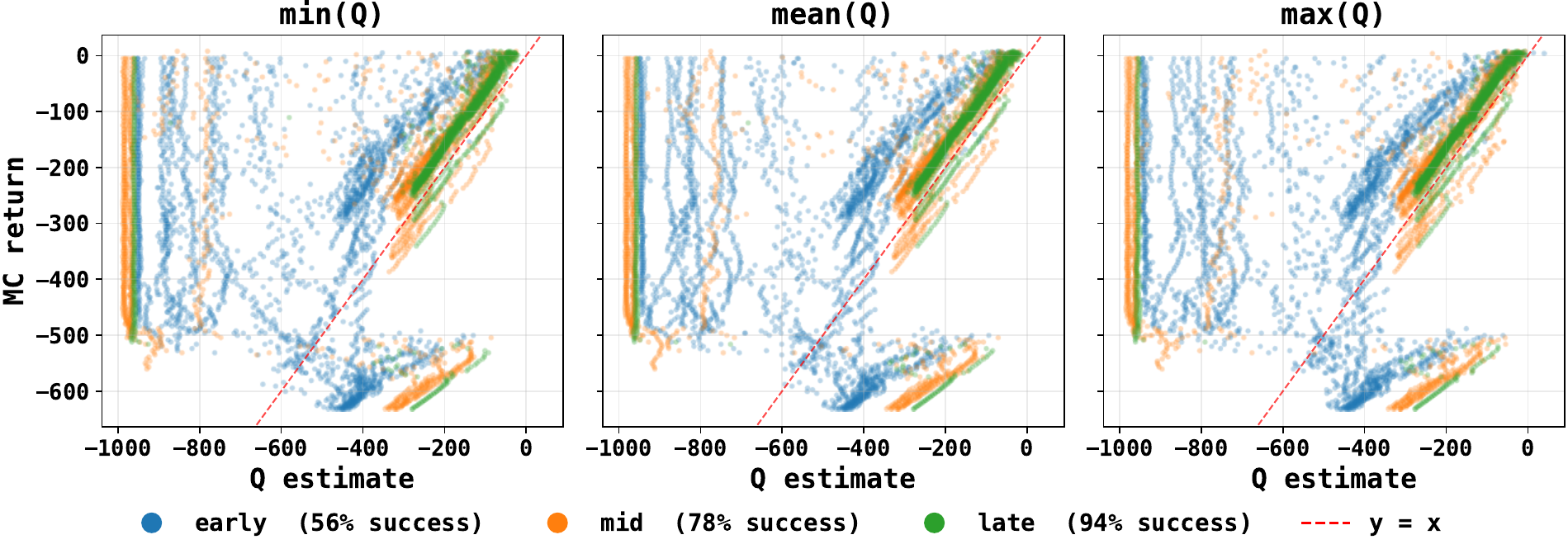}
        \caption{\OGPOplusca{}}
        \label{fig:q_vs_mc_ogpo_ca}
    \end{subfigure}

    \caption{We take early-, mid-, and late- training checkpoints for \OGPO{} and \OGPOplusca{} to rollout 32 trajectories and visualize the min, mean, and max Q vs ground-truth, Monte-Carlo returns. (a) Shows \OGPO{}'s Q values fluctuating widely between over- and under-estimating returns. (b) Shows \OGPOplusca{}'s Q values converging more stably around the $y=x$ axis, demonstrating Q values accurately estimating returns.}
    \label{fig:q_vs_mc_plots}
\end{figure}

A second challenge in offline-to-online RL is the pervasive ``dip'' in performance that arises  transitioning from offline pretraining to online RL. Warm-starting methods like \citep{uchendu2023jump, zhou2024efficient} propose the use of high update-to-data (UTD) ratios and/or offline datasets during online RL, and the use of pessimistic critic updates. Anecdotally, we find that neither of these methods suffice\RETURNN{add experiments}.  
Moreover, from \Cref{fig:q_vs_mc_ogpo}, we see that both over- and under-estimation of the $Q$ values are possible, and both outliers potentially destabilize training. Thus, we instead to have the policy extraction step maximize the  \colorbold{conservative advantages}. This  is made possible because our zero-order extraction takes advantages directly, and also accounts for the fact that global additive errors in critic values are less salient than incorrect \emph{advantage} estimation.

For a given action $a_i$, we set
\begin{equation}
\label{eq:adv-conservative}
\advcons_\aindex \;=\;
\begin{cases}
\min_m \advindie & \text{if } \min_m \advindie > 0, \\
\max_m \advindie & \text{if } \max_m \advindie < 0, \\
0 & \text{otherwise.}
\end{cases},
\end{equation}
where we recall $\advindie = Q_{\phi_m}(s^{(i)},a_{0}^{(i,\aindex)}) \;-\; \frac{1}{\Ngroup}\sum_{i' = 1}^{\Ngroup} Q_{\phi_m}(s^{(i)},a_{0}^{(i,\aindex)})$
is the group-wise advantage using the $m$-th network in the ensemble. \Cref{eq:adv-conservative} provides a non-zero advantage (and thus updates the policy) if and only if \emph{all advantages} have the same sign, thereby robustifying updates to estimation errors in the critic networks. As shown in \Cref{fig:q_vs_mc_ogpo_ca}, we we see that policy extraction with conservative advantages also improves the calibration of critic estimation, in that critic values in earlier states of training more tightly track those in later stages.

\subsection{Conservative Advantages (\OGPOplusca) Enable Stable Training on Images}
\begin{figure}[H]
    \vspace{-1.0em}
    \centering
    \includegraphics[width=0.9\linewidth]{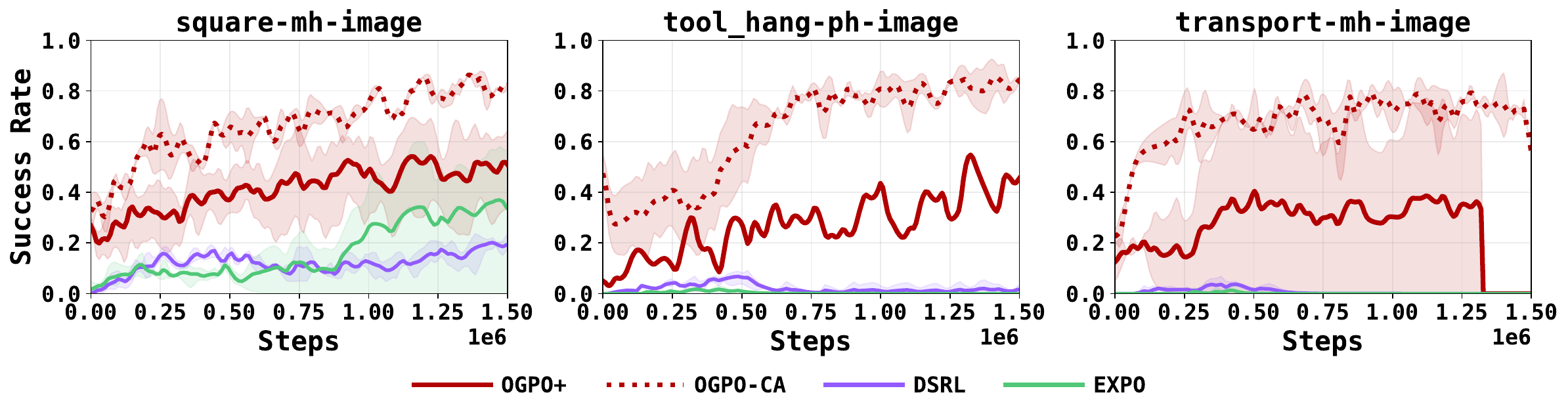}
    \vspace{-0.5em}
    \caption{We compare \OGPOplus{} and \OGPOplusca{} on \Robomimic{} tasks with image-observations}
    \label{fig:ogpo_images}
    \vspace{-0.5em}
\end{figure}
Finally, we consider \Robomimic{} tasks with image observations paired with robot proprioception information as a challenging setting for Q-learning and subsequently, policy extraction. From \cref{fig:ogpo_images}, we see that merely SFT via success buffer is not sufficient to guide policies to convergence. We observe that besides preventing the offline-to-online performance ``dip'', \OGPOplusca{} also plays a crucial role in stabilizing policy improvement in high dimensional settings where learning Q-values over the large embedding spaces, proprioceptions, and actions is challenging. Moreover, baselines such as \DSRL{} and \EXPO{} fail to converge in image-based settings with no offline data in the replay buffer.

\section{When does Full-Finetuning (\OGPO{}) Improve Over Popular Baselines?}
\label{sec:experiments}
In this section, we carefully compare \OGPO{}, \OGPOplus{}, and \OGPOplusca{} to a number of popular baselines to elucidate the merits and limits of its design philosophy--- full policy fine-tuning, off-policy critic learning, and PPO policy extraction. Our experiment environments are representative of many common challenges in robot learning (e.g. high precision, long horizon, mixed data quality), and baselines cover competing design philosophies (e.g. steering, residual learning).

\begin{table}[h]
\centering
\small
\begin{tabular}{@{}lccccc@{}}
\toprule
\textbf{Criterion} 
& \OGPO{}
& \QC{}
& \DSRL{} 
& \EXPO{} 
\\ 
\midrule
Mixed Data Quality
& \yesyes
& \yeskinda
& \kindakinda
& \nono
\\
High Precision Tasks
& \yesyes
& \mehkinda
& \nono
& \kindakinda
\\
Partial Demonstrations
& \yesyes
& \yesyes
& \mehkinda
& \nono
\\
Long Horizon
& \yesyes
& \kindano
& \kindano
& \nono
\\ 
Dense/Dexterous
& \mehkinda
& \mehkinda
& \kindano
& \yesyes 
\\ 
\cdashline{1-6}
High Sample Efficiency
& \yesyes
& \mehno
& \nono
& \kindano
\\ 
\bottomrule
\end{tabular}
\caption{Left (resp. right) symbol indicates achieving high success With (resp. Without) task-specific hyperparameter tuning. \xmark - fails to converge on all tasks; \pmark - converges on some but not all tasks; \pcmark - converges on all tasks, but below SOTA success/efficiency; \cmark - converges on all tasks, competitive with SOTA success/efficiency. We use the optimized variants where possible (e.g. \OGPOplus{} for \OGPO{} and similarly for all the baselines).  \label{tab:method-comparison}}
\end{table}

\colorbold{Summary of Findings.} We summarize comparisons to other off-policy methods in \Cref{tab:method-comparison}. Each method has two columns: left denotes if the method converges with task-optimized hyperparameters, and right denotes fixed hyperparameters across all tasks within the criterion (see \Cref{app:hyperparameters}). The markings are explained in the table caption. 

We find that \textbf{\OGPO{} is able to learn in sparse-reward tasks with mixed/partial data quality and on high-precision/long horizon tasks}, whereas other methods  struggle in one or more of these regimes. It also exhibits (often times drastic) gains in sample efficiency compared to these methods, and order-of-magnitude improvements related to the on-policy \DPPO{} algorithm. However, \OGPO{} is less performant on the dense-reward tasks from the Adroit Hand benchmark (\Cref{fig:adroit_comparisons}). 

Comparisons are detailed further in \Cref{sec:comparison}. Sample efficiency improvements v.s. \DPPO{} are expected (off- vs. on-policy), and we attribute gains against off-policy baselines to exploration behavior and expressive policy updates, studied in \Cref{sec:explore}. 
\Cref{app:ablations} ablates the merits of zero-order policy updates vs. backpropagation through time, the role of \emph{negative-advantage gradients} in encouraging exploration, and the enhancements distinguishing \OGPO{} and \OGPOplus. 

\subsection{Experimental Setup}\label{sec:exp_setup}
\colorbold{Baselines.} We compare against the baselines mentioned in \Cref{sec:rel_work_baselines}, which are described in more detail in \Cref{app:baselines}  In short, we consider: (i) \DPPO{} \citep{ren2024diffusion}, representative of on-policy learning, (ii) \DSRL{} \citep{wagenmaker2025steering}, representative of off-policy noise steering (iii) \EXPO{} \citep{dong2025expo}, representative of learning residual corrections to the GCP, and (iv) to a variant of \QC{} \citep{li2025reinforcement} representative of behavior cloning policy extraction. We do not compare to ReinFlow \citep{zhang2025reinflow} due to reported reduced sample efficiency compared to \DPPO, making the latter a more compelling baseline. We also skip comparison to PA-RL \citep{mark2024policy} for reasons described in \Cref{app:baseline:parl}. Lastly, we introduce a steering+residual learning baseline, (v) \SR, combining \DSRL{} and \EXPO{} to (hypothetically) yield the benefits of both. For a fair comparison with \OGPOplus{}, we implement each baseline with its own best-practices, as described in \Cref{app:baselines}. 

\colorbold{Environments.} Our simulation environments are chosen to elicit key challenges faced in modern robot learning:
\emph{Robomimic:} To test high-precision robotic control, we use three $\robomimic$ tasks \citep{robomimic2021}: $\rmsquare$ (medium-horizon insertion), $\rmtoolhang$ (long-horizon multi-step insertion), $\rmtransport$ (bi-manual long-horizon transfer). $\rmsquare$ and $\rmtransport$ use Multi-Human (MH) datasets; $\rmtoolhang$ uses Proficient-Human (PH) with BC stopped at 50\% success. 
\emph{Franka Kitchen:} We use the $\fk$ benchmark \citep{gupta2019relay} with a Franka robot manipulating 4 kitchen objects, testing sensitivity to multi-step trajectories with complete demonstrations ($\fkcomplete)$,  randomized subtask orders ($\fkmixed$), and sequential partial trajectory data ($\fkpartial$).
\emph{Adroit:} To test performance in dextrous manipulation tasks with dense-reward, we use the 24-DoF {Adroit Hand} benchmark:\textsc{Door-v1}, \textsc{Hammer-v1}, \textsc{Pen-v1}, \textsc{Relocate-v1} for door opening, hammering, pen reorientation, and object relocation. Expert datasets from D4RL/Minari. 
\emph{LIBERO:} Finally, to test image-based language-conditioned manipulation,  we use the $\robomimic{}$ and $\libero$ benchmarks \citep{liu2023libero}. Further details are given in \Cref{app:envirs}.



\colorbold{Experimental Regime: Online RL from a BC Checkpoint.} We emulate real-world robot learning settings where large-scale pretrained policies with varying levels of online success rates are deployed to learn novel tasks \textbf{without access to offline datasets during online RL}. Thus, we pre-train a flow GCP for all baselines, clip it to at most 50\% success rate, and use the same BC checkpoint for all baselines in online RL without additional data. Full details in \Cref{sec:initialization}. \RETURNN{Add the 10\% success experiments here}




\subsection{Comparison to other methods}\label{sec:comparison}

\begin{figure*}[h]
    \centering
    \includegraphics[width=0.9\linewidth]{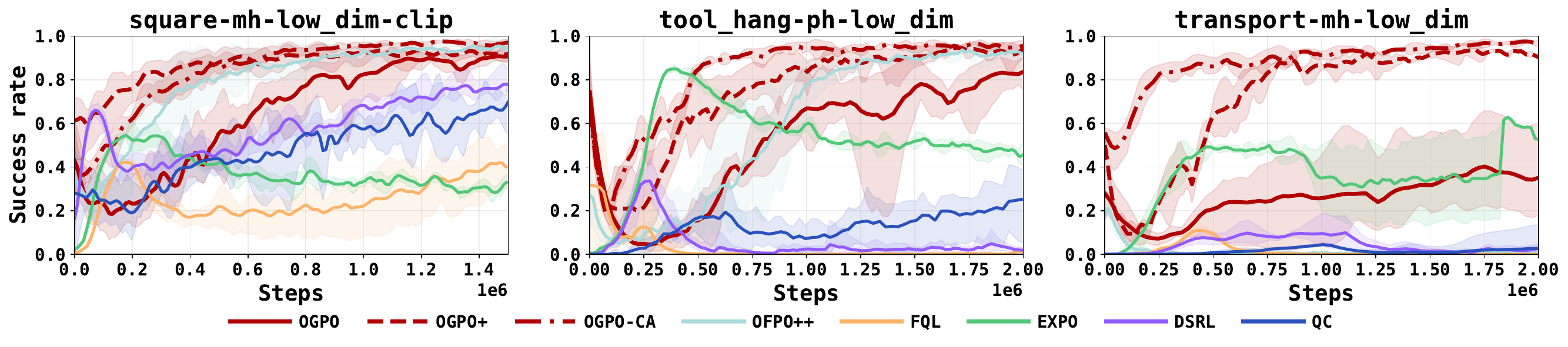}
    \vspace{-0.5em}
    \caption{Comparison with natural off-policy baselines (\EXPO{}, \DSRL{}, \QC{}), and on-policy algorithms modified to use OGPO-style off-policy value functions (\FPO{}, \FQL{}) on \robomimic{} \rmsquare{}, \rmtoolhang{}, and \rmtransport{}.}
    \label{fig:offpolicy_comparisons}
\end{figure*}
\vspace{-0.5em}
\begin{figure*}[h]
    \centering
    \includegraphics[width=0.9\linewidth]{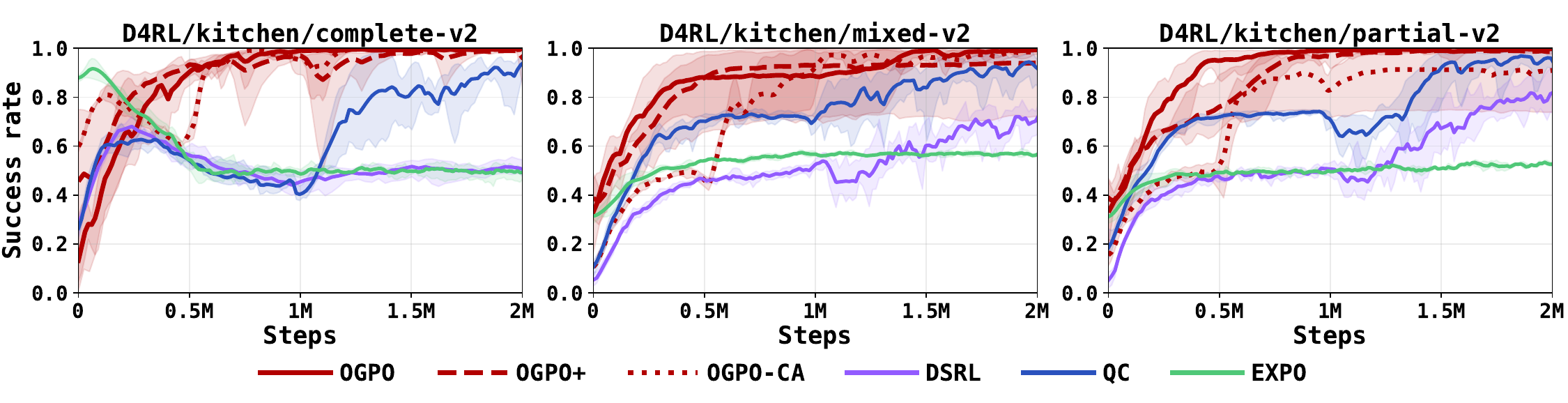}
    \vspace{-0.5em}
    \caption{Comparison against natural off-policy baselines (\EXPO{}, \DSRL{}, \QC{}) on \KitchenEnv}
    \label{fig:kitchen_comparisons}
\end{figure*}
\vspace{-0.5em}
\begin{figure*}[h!]
    \centering
    \includegraphics[width=0.9\linewidth]{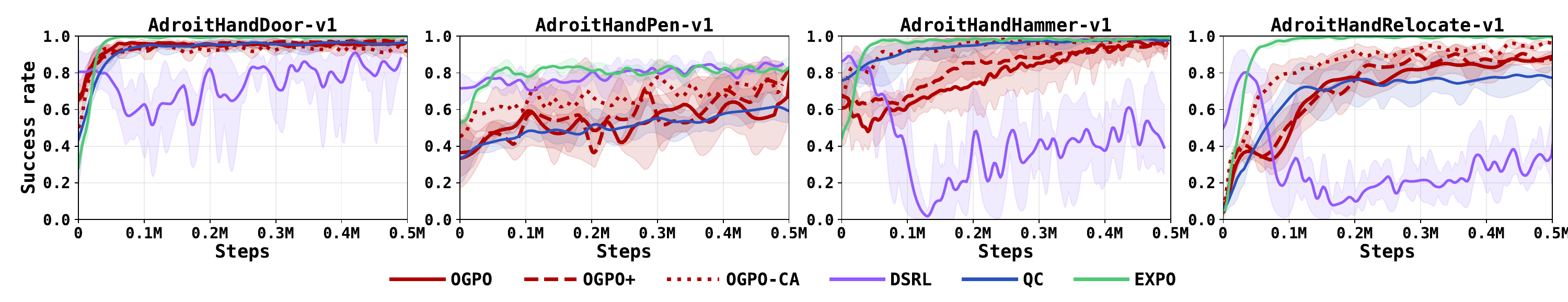}
    \vspace{-0.5em}
    \caption{Comparison against natural off-policy baselines (\EXPO{}, \DSRL{}, \QC{}) on the \AdroitEnv}
    \label{fig:adroit_comparisons}
\end{figure*}



\colorbold{Expressivity: Full-Policy Finetuning (\OGPO) vs. Steering (\DSRL) vs.  Residual (\EXPO).}
Next, we compare \OGPO{} to performant off-policy alternatives that do not fine-tune the full GCP across 11 aforementioned tasks.
\emph{Steering} (\DSRL) can be sample-efficient but relies on sufficient base policy action coverage, leading to suboptimal performance when the base policy's performance is poor, such as in \textsc{Kitchen} tasks. Further, by not updating later steps of the GCP, steering struggles on high-precision tasks such as the \textsc{Adroit} task suite. 
We also empirically found it to be sensitive to hyperparameters; in some tasks, DSRL performance crashes despite heavy tuning. We attribute some of this instability to our use of \DSRL\ on a flow-based GCP instead of a diffusion-based GCP; the original paper uses diffusion GCPs for low-data-coverage experiments. 
However, we are also more sample-efficient than \DSRL's paper-reported numbers on shared tasks.

\emph{Residual learning} (\EXPO) performs well when the base policy is strong and thus only minor residual corrections are needed (it is highly performant in \textsc{Adroit} in \Cref{fig:offpolicy_comparisons}), but, like steering (\DSRL), it generally performs poorly or is unstable when the base policy performance starts lower (\textsc{Kitchen} and most \robomimic\ tasks).
We note that when given offline data, \EXPO\ can perform well (see \Cref{fig:adroit_comparisons}), but our experimental regime is without access to the pre-training data.
Our \emph{Steering + Residual Learning} (\SR) baseline combines \EXPO\ and \DSRL; we plot sample efficiency curves in \robomimic\ tasks in \Cref{fig:dsrl_plus_expo}, where we see that it is better than \EXPO/\DSRL\ alone in \rmsquare, albeit still worse compared to \OGPO, and demonstrates unstable training in the high precision \rmtoolhang{}  task.

\colorbold{Off-Policy Learning vs. Self-Distillation/Behavior Cloning (BC) with \QC.}
Next, we compare \emph{policy extraction} methods. We find the action-chunked \BPTT{} variant proposed in the \QC\ paper to perform poorly (\cref{fig:bptt_vs_ogpo_fig}) on flow policies, and thus use a variant that explores online with Best-of-$N$ action sampling and fine-tunes the BC policy on transitions from the online replay buffer.
\QC\ plateaus at lower performance for most tasks, requires more task-specific hyperparameter tuning, and has worse sample efficiency. We attribute this to SFT's inability to expand the support of the GCP action distribution, required for sufficient exploration. 

\colorbold{Off-Policy \OGPO{} vs. On-Policy \DPPO.} 
Finally, we compare \OGPOplus{} against \DPPO, where the major difference between the two is that \OGPOplus{} truncates the bi-level MDP proposed by \DPPO{} at the end of each denoising trajectory with terminal rewards coming from an off-policy Q-function,
\begin{wrapfigure}{r}{0.6\textwidth}
\vspace{-1em}
    \includegraphics[width=0.6\textwidth]{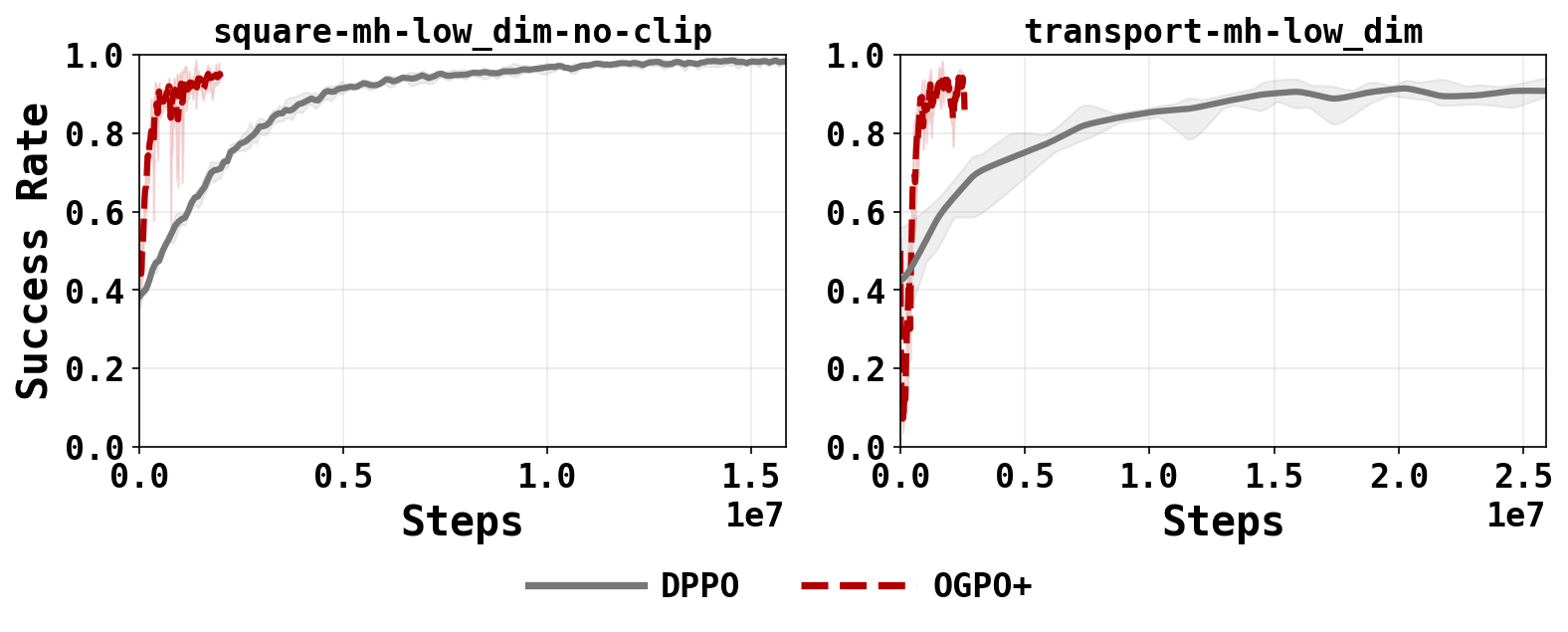}
    \vspace{-2em}
    \caption{{\OGPOplus{} substantially improves sample efficiency compared to the on-policy \DPPO\ algorithm.}}
    \label{fig:q_as_future_substitute}
    \vspace{-1em}
\end{wrapfigure}
while \DPPO\ treats the entire bi-level MDP as a single MDP to train with on-policy RL.
On final success rates across \robomimic{} \rmsquare{} and \rmtransport{}, this off-policy modification results in \DPPO\ taking $\sim10\times$ longer to reach the final success rates achieved by \OGPOplus.
Overall, we find that both \OGPO\ and \OGPOplus\ outperform \DPPO's paper-reported results in both sample efficiency and final performance across all shared tasks, even with matched network architectures and action chunk lengths.

\begin{AIbox}{Summary: \OGPOnc{} outperforms natural baselines}
\OGPO{} outperforms all natural off-policy baselines in sparse reward precise manipulation settings, and is an order of magnitude sample efficient than on-policy methods with minimal hyperparameter tuning. 
\end{AIbox}

\section{Understanding and Ablating The Merits of \OGPO{}}

\subsection{Does  \OGPO{} Encourage Exploration?}
\label{sec:explore}

\begin{figure}[H]
    \centering
    \begin{minipage}[c]{0.52\textwidth}
        \centering
        \includegraphics[width=\linewidth]{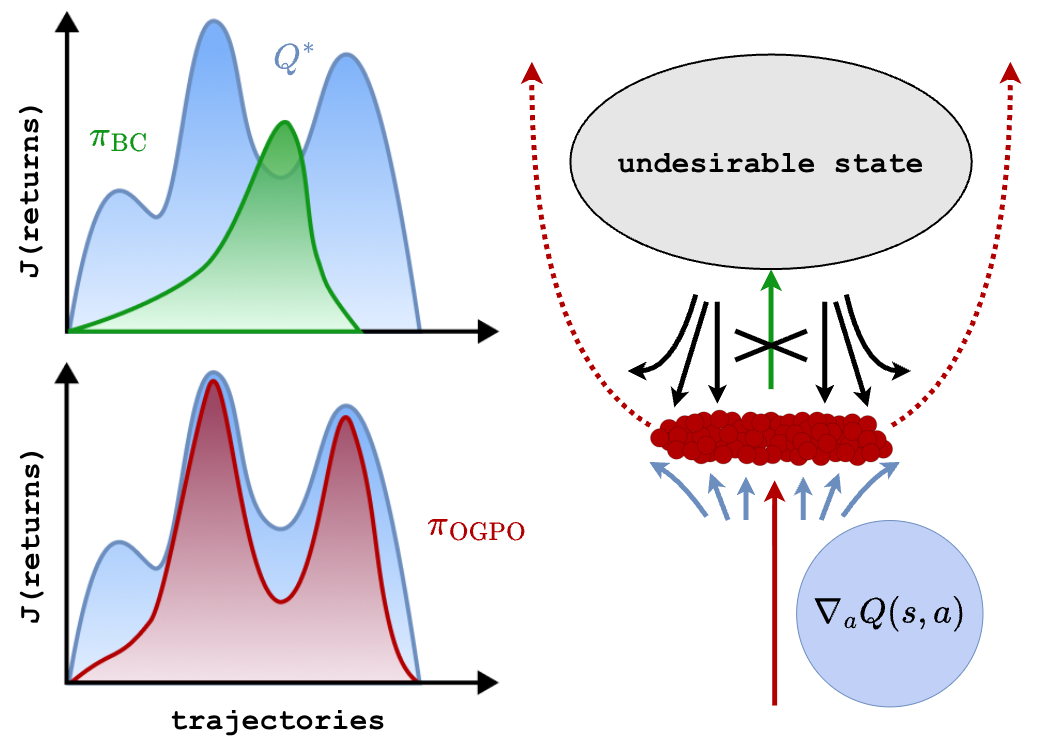}
    \end{minipage}
    \hfill
    \begin{minipage}[c]{0.43\textwidth}
        \caption{
        \textbf{Left:} Consider a policy with two equally near-optimal modes that are only weakly covered by the BC data (green). 
        \OGPO{} maintains coverage of both modes even after convergence. \emph{How?} 
        \textbf{Right:} We illustrate our mental model with an example where bi-modality arises from a bifurcation around an obstacle or undesirable state, shown in gray. 
        In this setting, $\nabla_a Q(s,a)$ points toward the obstacle, while directions orthogonal to $\nabla_a Q(s,a)$ move perpendicular to it. 
        By preserving action variance orthogonal to $\nabla_a Q(s,a)$, \OGPO{} maintains coverage over action chunks that can separate into the ``left'' and ``right'' trajectory modes.}
        \label{fig:grad_q_exploration}
    \end{minipage}
\end{figure}



By aggressively exploiting the critic (\Cref{sec:exploit}), \OGPO{} generates actions beyond the support of the offline data distribution used in the BC phase (\Cref{fig:grad_q_exploration}, \emph{left}), resulting in high task success as well high \emph{task efficiency}, measured in terms of time-steps to completion. Here, we identify a surprising finding:
\begin{AIbox}{}
\begin{quote}\OGPO{} \textbf{generates highly diverse trajectories}, despite aggressively exploiting the critic for high success rates and task efficiency. 
\end{quote}
\end{AIbox}
Whereas diversity, optimality and task efficiency are often regarded as being at odds \citep{huang2025self,setlur2025e3}, we show that \OGPO{} accomplishes all simultaneously. Below, we present extensive evidence for this finding, and propose a mental model, summarized in \Cref{fig:grad_q_exploration}, \emph{right}, as to how \OGPO{} achieves this affect.

\begin{figure*}[t]
    \centering
    \includegraphics[width=0.95\linewidth]{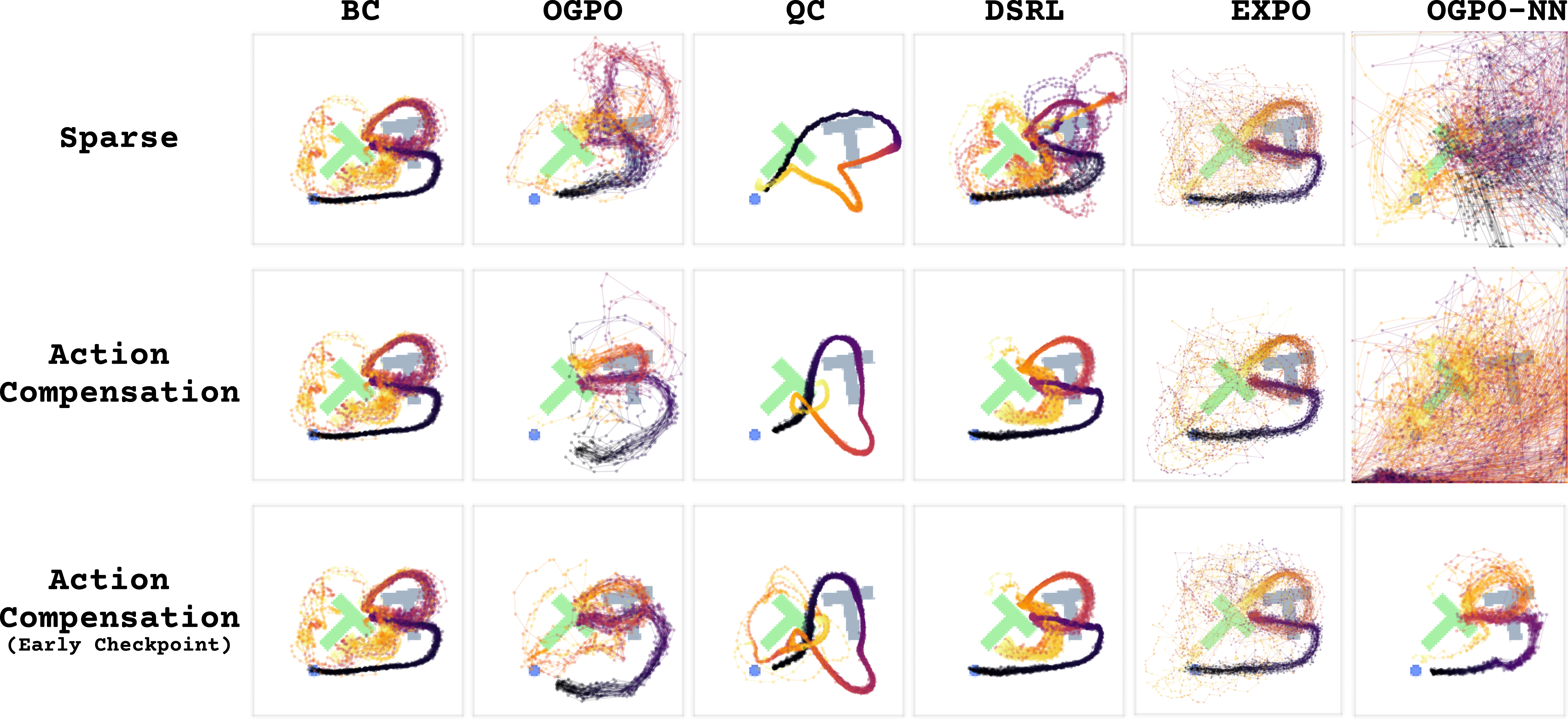}
    \caption{We plot 50 rollout trajectories on the \texttt{pushT} task with (top) sparse reward, (middle) sparse reward with $\Delta a_t$ compensation, and (bottom) early-stage sparse reward with $\Delta a_t$ compensation policy. Compared to the baselines, we observe \OGPO{} learns policies with faster execution, minimal fine adjustments, and action spaces wider than the BC distribution in the sparse reward setting. The observation that action compensation forces OGPO to adhere to the rollouts in the vicinity of BC policies further confirms \OGPO{}'s exploration tendencies. 
    }
    \label{fig:manifold_expansion}
\end{figure*}

\paragraph{\OGPO{} drives greater trajectory diversity.} We study the \texttt{PushT} task \citep{chi2023diffusion}, a classical example of trajectory-level multimodality, where a blue-dot pushes a gray ``T'' to the green goal configuration (\Cref{fig:manifold_expansion}). We consider  two reward settings: the classical sparse reward $r = -\I\{\text{not done}\}$, and an ``action-compensated'' reward $r = -(\I\{\text{not done}\} + \lambda \|\Delta a_t\|)$ which penalize per-step action magnitudes (due to PushT's physics enabling unbounded actions). We compare \OGPO{} against natural baseslines and visualize the learned trajectories in \Cref{fig:manifold_expansion}.  Here,  dark points represent the initial actions in the trajectory, and the color lightens to  yellow ones as the time-step progresses. In the absence of action-compensation, \OGPO{} learns to take larger action that complete the trajectory in fewer time steps (task-efficiency), and with full success.\footnote{ Note that without action-compensation, path-\emph{length} is not constrained, only time to completion.} 

Still, \OGPO{} \emph{preserves a relatively wide manifold of valid actions} \citep{ren2024diffusion}, and seems to \emph{preserve additional trajectory-level modes}. On adding an action compensation term, \OGPO{} takes smaller steps and prunes many of its modes, favoring modes which allow shorter path-length. This makes sense as \OGPO{} directly exploits the critic, yielding actions closer to optimal and further from the base policy.

Comparing to the baselines, \QC{} and \DSRL{} show limited manifold expansion, remaining closer to the BC initialization. \EXPO{}'s residual policy facilitates support expansion but not optimal policy extraction. This can be seen by a range of corrective actions being taken near the T-shape handle.  Lastly, we test \OGPOnn{}, which zeros out all negative advantages and retains only positive advantages. Whereas prior work \citep{setlur2025e3} would suggest that negative advantages \emph{increase exploration},  we find that they also seem necessary for ``sharpening'' \citep{huang2025self} towards optimal modes.

\begin{figure*}[t]
    \centering
    \includegraphics[width=0.95\linewidth]{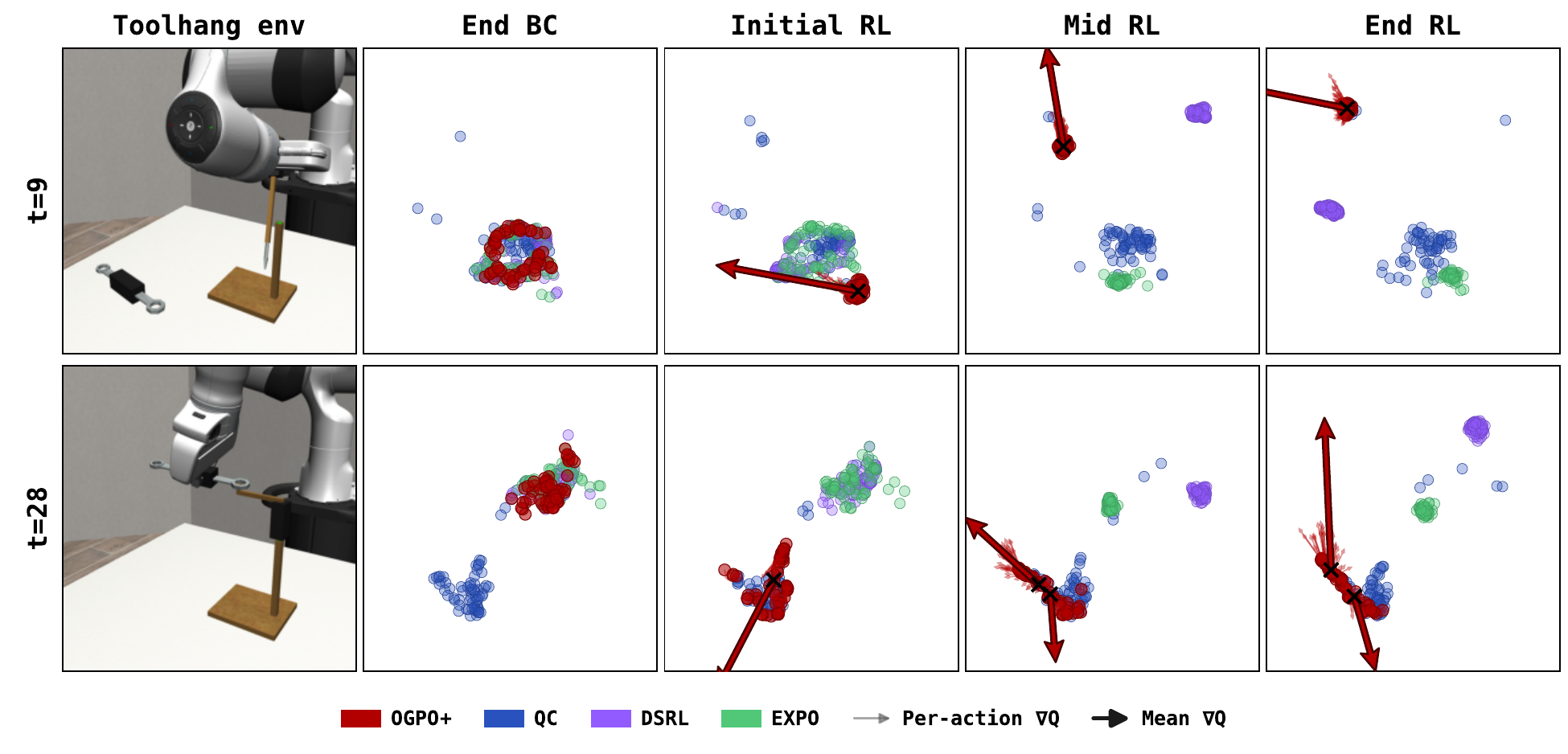}
    \caption{We plot the UMAP embeddings \citep{mcinnes2018umap} of actions generated via \OGPOplus{} and the natural baselines. We show the Q-function gradients with respect to \OGPOplus{} actions ($\nabla_aQ(s,a)$) projected in the same UMAP space as well as a vector sum of the per-action $\nabla_aQ(s,a)$ denoted as the consensus $\nabla Q$. \OGPO{} actions show a sharp variance reduction compared to the baselines, especially in axes orthogonal to the consensus $\nabla Q$ direction.}
    \label{fig:umap_baselines}
\end{figure*}


\paragraph{\OGPO{} preserves action variance ``orthogonal'' to task success.} We now uncover the \emph{concrete mechanism} by which OGPO preserves both \emph{action} and \emph{trajectory}-level diversity. We compare \OGPO{} to relevant baselines on $\rmtoolhang$, isolating two critical states at times $t=9$ (the needle being transported toward the hole) and $t = 28$ (the wrench being inserted) from a single, shared demonstration trajectory with maximal variance in Q-values. Each policy is trained from the same BC checkpoint, removing spurious variation\footnote{note that \QC{} uses additional critic distillation during the BC phase, leading to different actions after offline training}. 

For each time step, we pool together 64 actions from policies trained with each baseline, compute a common UMAP embedding \citep{mcinnes2018umap}, and visualize them in \Cref{fig:umap_baselines}, color coding actions by method. For the \OGPO{} actions, we also plot arrows that compute the gradient $\nabla_a Q(s,a)$ of the mean $Q$ function (from after training), which gives the local direction of steepest ascent for actions to improve the critic value (see caption for details). To visualize the $\nabla_a Q(s,a)$ for each action $a_i$, we compute unit vectors $u_i=\frac{\nabla_{a_i} Q(s,a_i)}{||\nabla_{a_i} Q(s,a_i)||}$ and measure an agreement score $\psi=||\frac{1}{N}\sum_i^N u_i||$. When $\psi>0.6$ we consider majority of actions having the same $\nabla_a Q(s,a)$ unit vectors, and $\psi\leq0.6$ as there not being a consensus, at which, we compute K-means clusters over $\nabla_a Q(s,a)$ with cluster centers shown as black crosses in \Cref{fig:umap_baselines}. We include snapshots across four phases of training, from offline to completion.

Our findings reveal \textbf{that \OGPO{} increases variance in a \emph{selective} manner.} At $t = 9$, there is minimal trajectory level diversity due to the ensuing precision requirements. Thus we see \OGPO{} exhibits the \emph{most aggressive} shrinking of action variance. However, at $t = 28$, greater action action variance is permitted, and preserved even at the \emph{end of training} (\Cref{fig:umap_baselines}, \emph{bottom right}).  However, \OGPO{} does not increase variation isotropically: rather, the remaining action-variance as even \emph{orthogonal} to the critic gradient. Note that, along these directions, differences in actions have \emph{zero effect on critic values}, to first order. Therefore, we find the \OGPO{} \emph{allocates large variance along directions which do not affect task success}. At the same time,  \OGPO{} (a) sharpens the distribution orthogonal to these directions (resulting in the ``thin'' ellipsoid seen in Mid/End training in at $t= 28$), while 
(b) aggressively ``stretching'' the action distribution to align with critic gradients in parts of the action distribution when gradients $\nabla_a Q(s,a)$ exhibit strong consensus, e.g. $\psi > 0.6$. Thus, \OGPO{} can \emph{both} optimize the critic for task performance/completion time while \emph{simultaneously} preserving as much  action diversity as possible.

\paragraph{Mental model: mode-preservation via orthogonal action variance.} Here, we propose a mental model for how \OGPO{}'s selective ``stretching'' tendency preserves trajectory-level multimodality.  
We depict a mental-model of this trajectory-level multimodality in \Cref{fig:grad_q_exploration}. We consider a task with two modes---navigating left or right around an obstacle. The optimal branching point is depicted with (\emph{red dots}). Below these, the optimal actions approach the obstacle, whereas above, they move around it. In the center, $\nabla_a Q(s,a)$ points vertically (either up or down). \OGPO{} preserves variance orthogonal to this direction, preserving actions which ultimately branch into the left and right-modes. Thus, allocating variance at the ``decision point'', while ``stretching'' actions at the extremes, sharpens the trajectory distribution around \emph{both} feasible modes.

\begin{AIbox}{How does \OGPOnc{} preserve exploration?}
    \begin{itemize}[leftmargin=*]
        \item For critical states precursor to future high precision demanding ones, \OGPO{} learns a strong $\nabla_a Q(s,a)$ consensus and subsequently contracts the policy toward a narrow high-value action manifold. 
        \item In states admissible of multiple near-optimal actions, \OGPO{} preserves variance along directions approximately orthogonal to $\nabla_a Q(s,a)$, where action perturbations have negligible first-order effects on the value. 
    \end{itemize}    
\end{AIbox}

\colorpar{Why does \OGPO{} preserve exploration?} An important question to ask is: why does \OGPO{} preserve exploration better than alternatives? A comprehensive account would warrant further study, which we defer to future work. Here, we hypothesize that the key factor which allows \OGPO{} to preserve variance comes from finetuning all the steps of the generative process; in Appendix \cref{fig:umap_policy_extraction}, we observe that other full-finetuning methods (e.g. FPO++ \citep{yi2026flow}), also preserve variance, though to a slightly lesser degree. We hope to pursue the full breadth of this question in a future study. 



\begin{figure}[H]
    \centering
    \includegraphics[width=0.99\linewidth]{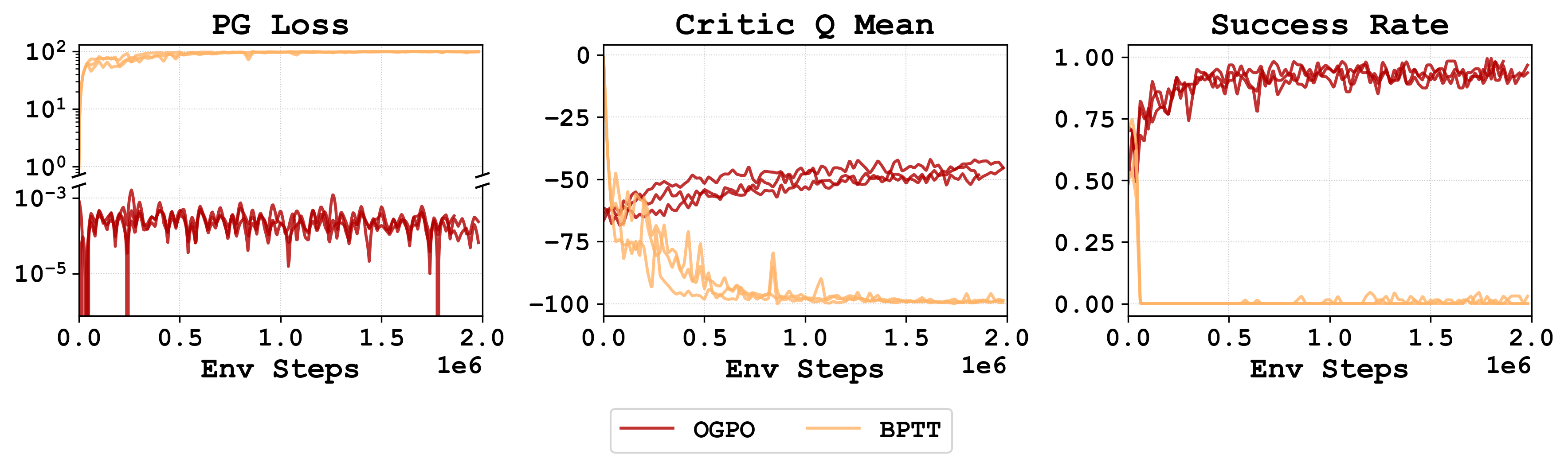}
    \caption{\BPTT{} uses Q-values directly to backpropagate gradients along the entire GCP chain. This results in unstable gradients and poor convergence. In contrast, \OGPO{} uses PPO-style policy gradient loss using Q-functions described \cref{eq:ppo_internal}. This results in stable gradients and sample-efficient convergence.}
    \label{fig:bptt_vs_ogpo_fig}
\end{figure}

\subsection{Does PPO policy extraction outperform natural alternatives (AWR, FPO)?}
\label{sec:pol_extr_expts}
\begin{table}[h]
\centering
\small
\renewcommand{\arraystretch}{1.15}
\begin{tabular}{@{}lccccc@{}}
\toprule
\textbf{Method} 
& \multicolumn{2}{c}{$\advgrpo \geq 0$}
& \multicolumn{2}{c}{$\advgrpo < 0$} \\
\cmidrule(lr){2-3} \cmidrule(lr){4-5}
& $\mathrm{loss}_{(+)}$ & \textbf{$\mathrm{weight}_{(+)}$}
& \textbf{$\mathrm{loss}_{(-)}$} & \textbf{$\mathrm{weight}_{(-)}$} \\
\midrule
\OGPO{}
& PPO likelihood+clip 
& $\advgrpo$
& PPO likelihood+clip
& $\advgrpo$
\\
\OGPOnn{}
& PPO likelihood+clip
& $\advgrpo$
& \xmark
& \xmark
\\
\AWOGPO{}
& PPO likelihood+clip
& $\exp{\left(\advgrpo/\beta\right)}$
& PPO likelihood+clip
& $\exp{\left(\advgrpo/\beta\right)}$
\\ 
\AWOGPOpos{}
& PPO likelihood+clip
& $\exp{\left(\advgrpo/\beta\right)}$
& \xmark
& \xmark
\\
\AWR{}
& CFM loss
& $\exp{(\advgrpo/\beta  )}$
& CFM loss
& $\exp{(\advgrpo/\beta )}$
\\
\FPO{}
& $\exp(\text{CFM}))$+clip 
& $\advgrpo$
& $(\exp(\text{CFM}))$+clip
& SPO weighting
\\ 
\bottomrule
\end{tabular}
\caption{We present a tabular description of the differences in the policy extraction algorithms that are compatible with the \OGPO{}'s policy extraction framework given advantages ($\advgrpo$). \Cref{eq:policy_loss} succinctly describes the combination of $\mathrm{loss}_{(+/-)}$ and $\mathrm{weight}_{(+/-)}$ that contribute to the policy loss for the different methods. Above, PPO-likliehood corresponds to \Cref{eq:ppo_internal}, clipping clips likelihoods as in \Cref{eq:ppo_internal}, $\beta$ is a temperature hyperparameter chosen per-task, and CFM indicates the conditional flow matching loss \citep{lipman2022flow}. Further details are given in \Cref{app:pol_extr}}
\label{tab:policy_extraction_comparison}
\end{table}

We observe that \OGPO{} uses a simple API: apply \emph{any RL} algorithm to the denoising MDP whose terminal rewards are given by the critic. Specifically, we can write a more general loss of the form:
\begin{equation}
\begin{aligned}
    \mathrm{PolicyLoss(\theta \mid s,a^{K:0},\hat A^G, \theta)} &= \I\{\hat A^G \ge 0\} \cdot  \mathrm{loss}_{(+)}(\theta;a^{K:0},s)\cdot \mathrm{weight}_{(+)}(\hat A^G)\\
    &\quad+ \I\{\hat A^G < 0\} \cdot \mathrm{loss}_{(-)}(\theta;a^{K:0},s) \cdot \mathrm{weight}_{(-)}(\hat A^G) 
\end{aligned}
\label{eq:policy_loss}
\end{equation}
where the policy loss under a parameter $\theta$, for state $s$, denoising chain $a^{K:0}$, and advantage estimate $\hat A^G$ consists of a loss depending on $s,a^{K:0}$, and weighting depending on the advantage. To compare with alternatives, we decompose the loss into a separate terms depending on the advantage sign. 

We now describe a number of alternatives based on this formulation \Cref{tab:policy_extraction_comparison}. First, we compare to \AWOGPO{}, which uses exponentiated advantages as in \cite{peng2019advantage}, but instead reweighs the \OGPO{} likelihood ratio given in \Cref{eq:ppo_internal}. For both \OGPO{} and \AWOGPO, we also introduce a positive-only variants of \OGPOnn{} and \AWOGPOpos{}  which zero the loss/weighting when advantages are zero. In addition, we introduce \AWR{}, a natural baseline which up-weights the conditional flow-matching (CFM) loss rather than PPO likelihoods. Generally, this underperforms \AWOGPO{}, so we omit the no-negative advantage variant. For all AWR-style runs, we perform per-task hyperparameter tuning to determine an optimal temperature parameter $\beta$ to ensure a steelman comparison. Finally, we compare to extraction via FPO++ \citep{yi2026flow}, which applies a number of novel design decisions  detailed in \Cref{app:pol_extr}.  All methods use the same replay data, critic training, and group-wise advantages.

\begin{figure*}[h]
    \centering
    \includegraphics[width=0.85\linewidth]{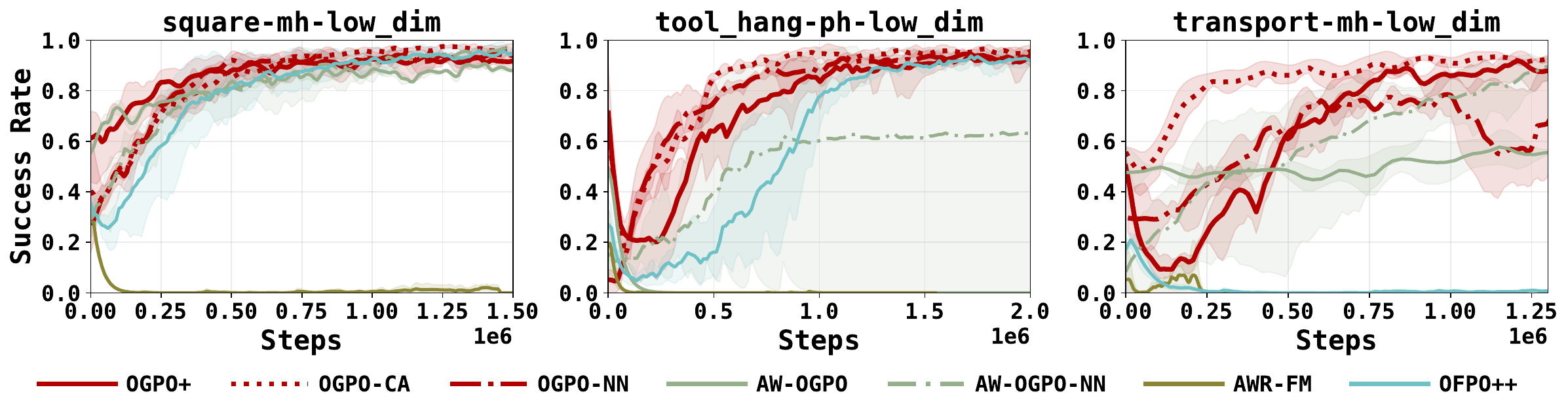}
    \caption{\OGPO{} comparisons with policy extraction ablations with \AWR{}, \AWOGPO, \AWOGPO{} and \FPO{} on \robomimic{} environments}
    \label{fig:pol_extraction_abl}
\end{figure*}

As shown in \Cref{fig:pol_extraction_abl}, \AWR{} fails across the \robomimic{} tasks, indicating that pure advantage-weighting of the flow loss is insufficiently  expressive compared to likelihoods that use the full denoising MDP. 
Using the full likelihoods in
\AWOGPO{}, positive-only \AWOGPO{}, and \FPO{} yields stronger performance, although not on par with \OGPOplusca{}.  In particular, the positive-variant of \AWOGPO{} outperforms that of normal \AWOGPO{}, by virtue of being more aggressive (note that regular \AWOGPO{} still has positive weights on likelihoods when advantages are negative), but still cannot reach full success on \rmtoolhang. On the other hand, \FPO{} collapses on the long-horizon \rmtransport{} task. 

Unlike \AWOGPO/\AWOGPOpos{}, removing negative-advantage gradients makes a minimal impact on \OGPO{} for tasks like \rmsquare{} and \rmtoolhang{}, where merely imitating high-valued action samples is sufficient to sharpen policy distributions (\Cref{fig:pol_extraction_abl}). However, for a task like \rmtransport{}, where avoiding suboptimal policy modes is critical for task success, we observe worse performance for both \OGPOnn{} as well as \AWOGPOpos{}. This suggests that negative advantages are important for mitigating suboptimal action distributions learned during pretraining. 

\begin{AIbox}{Why PPO updates are optimal.} We find that PPO style updates provide the \emph{most aggressive critic exploitation}, whereas AWR style advantages induce more modest updates that are suboptimal in online RL. Similarly, weighting denoising likelihoods (\Cref{eq:ppo_internal}) outperforms weighting the conditional flow-matching loss, again because the former is more aggressive. Stated succinctly, \textbf{just use the best \emph{on-policy} RL algorithm, i.e. PPO, for extraction from the off-policy critic}!
\end{AIbox}
\begin{AIbox}{\OGPOnc{} enables consistent cross-task hyparameters} \OGPO{}'s non-exponentiated advantage weighting removes the $\beta$ hyperparameter, which we find needs to be tuned for different tasks, due to sensitivity to advantage magnitudes. Thus, \OGPO{} functions with the \textbf{same hyperparameters} across domains, making it more suitable to extensions for multi-task learning.
\end{AIbox}





\subsection{Which Further Design Decisions Explain the Performance of \OGPO{} and \OGPOplus{}?}
\label{sec:ogplus_ablation}
\begin{figure*}[h!]
    \centering
    \includegraphics[width=0.85\linewidth]{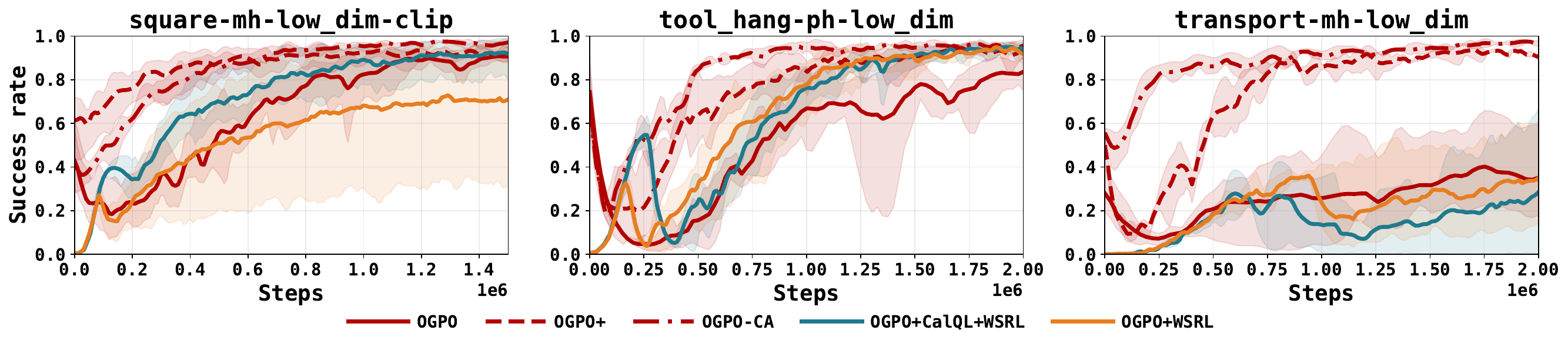}
\caption{\OGPOplus{} and \OGPOplusca{} obviate the need for offline-to-online Q-function RL}
\label{fig:calql_wsrl_ablation}
\end{figure*}

\begin{wrapfigure}{r}{0.35\textwidth}
\includegraphics[width=0.35\textwidth]{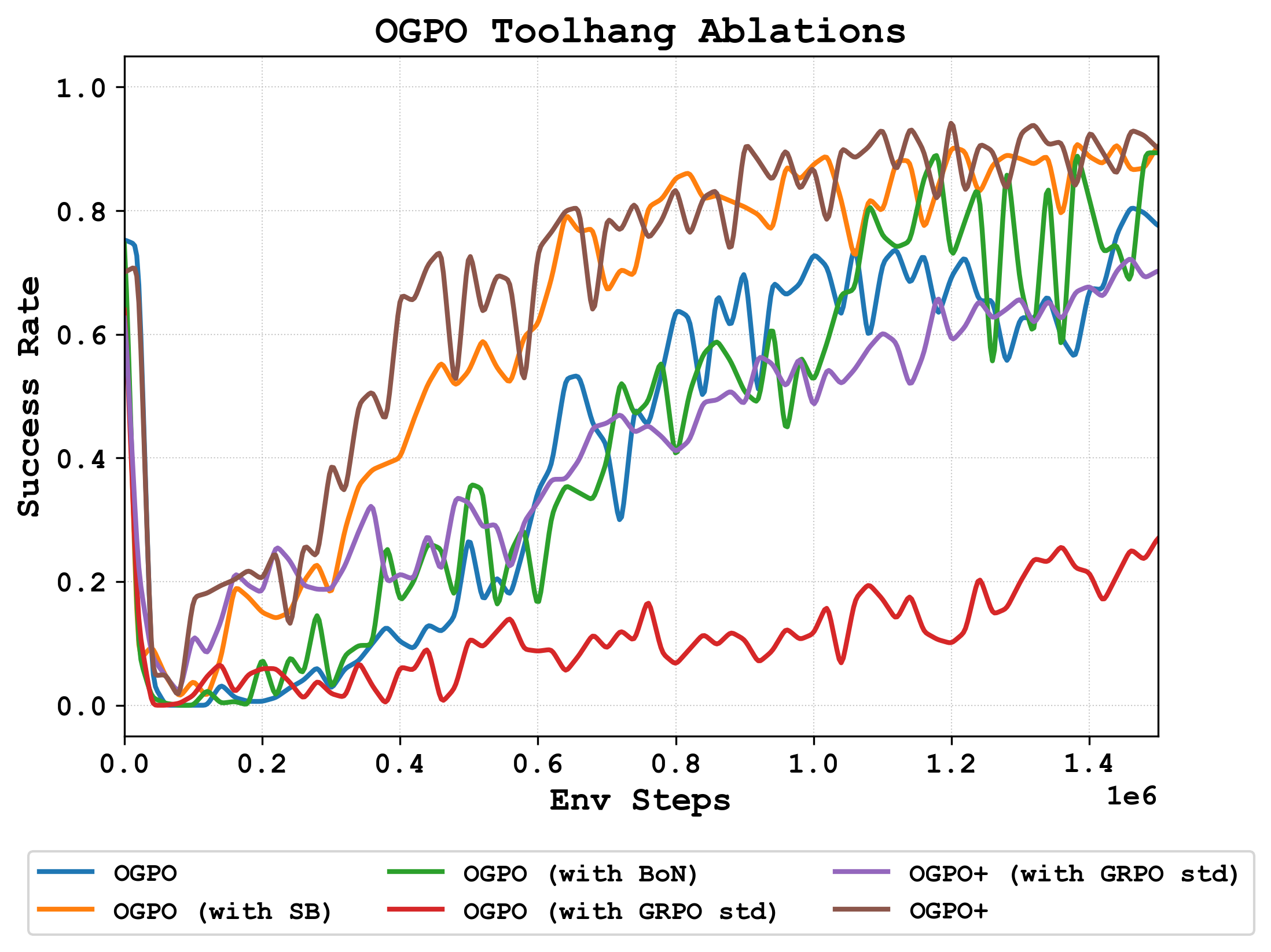}
    \caption{\OGPO{} - \OGPOplus{} design ablations show that success buffer plays a crucial role in \OGPOplus{}'s performance, while Best-of-$N$ plays the role of a verifier for improved critic learning by showing marginal improvements in performance. }
    \label{fig:ogpo_core_ablations}
    \vspace{-2.0em}
\end{wrapfigure}
The following ablations are designed to systematically isolate various subcomponent decisions within \OGPO{} and to explain which design choices align with maximizing sample efficient policy extraction. First, we compare \OGPO{}'s zeroth-order policy extraction to Backpropagation Through Time (\BPTT{}) that backpropagates first order gradients via Q functions and through the entire GCP denoising chain. As shown in \Cref{fig:bptt_vs_ogpo_fig} directly backpropagating through the denoising chain often fails catastrophically, supporting our choice to optimize the GCP via importance sampling rather than through $\nabla_a Q(s,a)$.

Second, using \Cref{fig:ogpo_core_ablations} as reference, Best-of-$N$ inference provides only marginal gains by itself and can increase oscillations when the critic is imperfect. This is consistent with the role of Best-of-$N$ as a verifier of critic learning at inference time, rather as a significant mechanism for policy improvement \citep{chow2025inference, huang2025is}. In contrast, the success buffer used in \OGPOplus{} consistently improves sample efficiency and asymptotic performance by anchoring policy improvement to successful behavior. We provide a mathematical basis for the intuition that conditional flow matching (CFM) loss between $\piigp$, and the success buffer actions increases the GCP lower-bound on successful modes in \Cref{sec:bc_elbo_barrier}. Moreover, we modify the advantage computation from
$\hat A = \frac{\Qtarg(s_t,a_{t,0}) - \hat V}{\hat{\sigma}}$, where $\hat \sigma^{(i)} \gets \sqrt{\frac{1}{\Ngroup} \sum_{j} \left( \Qtarg(s^{(i)}, a_0^{(i,\aindex)}) - \hat V^{(i)} \right)^2}$ and find that GRPO-style variance normalization hurts performance. Finally, we ablate the offline-to-online Q-learning recipe proposed in Warm Start RL (WSRL, \cite{zhou2024efficient}) with and without Calibrated Q-Learning (CalQL, \citep{nakamoto_cal-ql_2024}), and compare against OGPO, OGPO+, and OGPO+CA. We find that CalQL+WSRL slightly improves vanilla \OGPO{}, but fail to mitigate the policy collapse as prevented by \OGPOplus{} and \OGPOplusca{}.






\RETURNN{Add Floq experiments to compare against MIP}
\RETURNN{AWR w/ negative  \& CFM+AWR w/ pos}
\newcommand{\atsign}{\makeatletter @ \makeatother }
\section{Related Work}
\label{app:related_work}
We situate our work within the landscape of generative control policies, reinforcement learning for robotic control, and finetuning strategies for iterative generative models.

\subsection{Generative Control Policies}
The success of diffusion models in image generation~\citep{ho2020denoising, song2020denoising, rombach2022high} has inspired their adoption for robotic control. Diffusion Policy~\citep{chi2023diffusion} demonstrated that denoising diffusion probabilistic models (DDPMs) can effectively parameterize visuomotor policies by iteratively denoising action sequences conditioned on observations. Flow-matching policies~\citep{lipman2022flow, liu2022flow} offer a more efficient alternative by learning velocity fields that transport noise to action distributions through ordinary differential equations (ODEs), achieving comparable performance with fewer integration steps.

Recent work has sought to improve the generative modeling capacity. Notably, shortcut models~\citep{frans2024one} condition on desired step sizes to enable few-step generation, while consistency models~\citep{song2023consistency} distill multi-step diffusion into single-step generation. Recently, \citep{pan2025much} introduced Minimally Iterative Policies (MIP), demonstrating that two-step regression-based policies can match full flow model performance, suggesting that distributional learning may be less critical than previously believed. Orthogonally, tokenized autoregressive policies such as FAST~\citep{pertsch2025fast} encode continuous action chunks via discrete cosine transforms to enable efficient training of vision-language-action (VLA) models on high-frequency control data.

For \OGPO{}, we demonstrate flow and diffusion-based policies as representative of the general IGP formulation and leave generalization to other formulations as future work. 

\subsection{Reinforcement Learning for Robotic Policy Finetuning}
The incorporation of Reinforcement Learning (RL) into robotic policy training mirrors the post-training paradigm in large language models \citep{ouyang2022training, shao2024deepseekmath}. On-policy methods such as REINFORCE \citep{williams1992simple} and PPO \citep{schulman2017proximal} update policies using only data from the current policy iteration, ensuring stable but sample-inefficient learning. DPPO \citep{ren2024diffusion} extends PPO to diffusion policies by computing policy gradients through the denoising chain, while Reinflow \citep{zhang2025reinflow} applies similar principles to flow-matching policies.

Off-policy algorithms promise greater sample efficiency by maintaining replay buffers of past experiences. Classical approaches such as SAC \citep{haarnoja2018soft}, TD3 \citep{fujimoto2018addressing}, and REDQ \citep{chen2021randomized} learn Q-functions from off-policy data to guide policy updates. Temporal difference learning mitigates the requirement of the policy to compute Monte Carlo return to the go. However, naive application to IGPs in the RL-finetuning regime can exhibit training instabilities due to large initial distributional shifts and value overestimation. To mitigate these, \citep{mark2024policy, li2025reinforcement} proposed using Q functions merely to rank stochastic policy actions and fine-tuning the policy using the Best-of-N actions. However, driving policy improvement via Q-function ranking can be inefficient as it requires exploration away from the mean values of the flow policy.

Concurrently, RL-100~\citep{lei2025rl} presents a comprehensive real-world RL framework built on diffusion policies, demonstrating deployment-grade success rates across eight manipulation tasks. RL-100 adopts the same bi-level MDP formulation and clipped PPO surrogate as DPPO, unifying imitation and reinforcement learning under a single objective across both offline and online stages, and additionally incorporates consistency distillation for high-frequency deployment. While RL-100 demonstrates impressive real-world reliability, its policy optimization remains fully on-policy, requiring iterative offline data expansion to achieve sample efficiency. \OGPO{} instead decouples the bi-level MDP via off-policy critic learning, achieving comparable or superior sample efficiency in simulation without requiring multiple rounds of offline RL pre-training.


\subsection{Finetuning Strategies for Generative Control Policies}
\label{sec:rel_work_baselines}
Existing approaches to finetuning GCPs differ along the axis of \emph{what} is optimized. Steering methods, exemplified by \DSRL{} \citep{wagenmaker2025steering}, optimize the distribution over initial noise $\afirst$ while freezing the pretrained denoising network. This constrains policy improvement within the support of the pretrained IGP distribution. Residual policy approaches such as \EXPO{} \citep{dong2025expo} train an additional network $\pi^{\text{res}}$ that modifies the final action $a_{\text{res}} = \pi^{\text{res}}(\alastt, \st)$, allowing mode shifts within the BC policy support but fails to facilitate discovery of new behaviors. 

Policy-agnostic RL (PA-RL)~\citep{mark2024policy} and Q-chunking (\QC{})~\citep{li2025reinforcement} employ Q-functions to rank behavior cloned policies with high-value actions or use $\nabla_a Q(s, a)$. Q-learning with Adjoint Matching (QAM)~\citep{li2026q} uses adjoint matching to convert the critic's action-gradient into a step-wise training objective for expressive flow or diffusion policies, avoiding direct backpropagation through the full denoising process. In the image generation domain, Flow-GRPO~\citep{liu2025flow} concurrently applied GRPO~\citep{shao2024deepseekmath} to flow matching models for text-to-image alignment, sharing with \OGPO{} the ODE-to-SDE conversion for injecting stochasticity into deterministic flow policies and the use of group-relative advantage estimation over parallel denoising trajectories. However, Flow-GRPO operates in the on-policy, bandit-like setting: rewards are terminal (image-level), the ``environment'' is a single-step generation with no dynamics, and advantages are estimated via group normalization of final rewards rather than learned Q-functions. 

In contrast, \OGPO{} addresses the multi-step robotic control setting, where off-policy TD-learning is essential for sample efficiency across long environment horizons, and the two-level MDP structure enables reuse of costly environment transitions while performing on-policy updates purely within the denoising MDP. However, in addition to zero-order optimization via Q functions, \OGPO{} performs SFT via Success Buffer actions for enhanced sample efficiency.


\section{Conclusion and Limitations}
We introduce \OGPO, an approach that combines the best of on-policy and off-policy methods for fine-tuning generative control policies (GCPs) and enjoys high success rates and sample efficiency across numerous tasks. However, \OGPO{} still has limitations, the most important being that the parallel denoising rollouts required to estimate Q-values can be prohibitively expensive for large VLA models due to the high inference costs. Future work focusing on Q-function learning fidelity can help ameliorate this limitation by reducing the number of parallel GCP rollouts. 

\newpage

\begin{AIbox}{TL;DR: Takeaways} 
\begin{itemize}
    \item \textbf{Takeway \#1:} \OGPO{} provides a mechanism for scaling training compute given a limited interaction with the environment. At a coarse level, the GRPO sampling, parallel denoising tracks per-state, full-policy finetuning, and updates to every step of the denoising process can be viewed as axes along which compute is expended (\cref{sec:grpo}). Our findings suggest that, even within the standard Actor-Critic template, simply \textbf{increasing training time computation} can improve sample efficiency drastically (\Cref{fig:offpolicy_comparisons}). 
    \item \textbf{Takeaway \#2:} While critic learning is widely believed to be bottleneck in online RL, our findings suggest that \textbf{better policy extraction alone} can yield substantial improvements in training stability and sample efficiency (\Cref{sec:pol_extr_expts}). 
    \item \textbf{Takeaway \#3:} RL finetuning need not cause ``mode collapse'' or ``distribution narrowing.'' Surprisingly, \textbf{full policy finetuning can increase action diversity} and enhance policy exploration, despite the fact that one is only trying to maximize reward (with no explicit entropy penalties) (\Cref{sec:explore}). Understanding this phenomenon is an exciting direction for future work. 
    \item \textbf{Takeaway \#4:} \textbf{Zero-order policy policy optimization} can be incredibly effective given sufficient computation as it avoids unstable gradients through denoising steps or critics, improving performance on high-precision tasks. Further, the form of the likelihood ratios still provides useful gradient information, and can move policy mass away from the BC distribution (\Cref{fig:grad_q_exploration}).  
    \item \textbf{Takeaway \#5:} Full-finetuning of GCPs can lead to issues of critic overexploitation. However, the best remedy is not to slow down learning through hyperparameter adjustments, enforce pessimistic policy/critic updates or regularize entropy/distance to the base distribution. Instead, \textbf{targeted interventions, like imitating successful trajectories }(\OGPOplus{}) or \textbf{modifying the advantages} (\OGPOplusca{}) are both reliable, preserve training efficiency, and ameliorate the need for task-specific hyperparameter tuning (\Cref{sec:reg_ogpo_dip,sec:planning}).
    \item \textbf{Takeaway \#6:} While there are many options for fine-tuning a multi-step GCP (AWR, FPO, etc), \textbf{simple PPO is the most effective}, most stable, and requires the least amount of hyperparameter finetuning.
\end{itemize}
\end{AIbox}


\subsubsection*{Acknowledgments}
MN would like to thank Qiyang Li for helping with the initial implementation, and Zhiyuan Zhou, Seohong Park and Aviral Kumar for their informative discussions. This research used the Savio computational cluster resources provided by the Berkeley Research Computing program at UC Berkeley. MS would like to thank Aviral Kumar and Andrew Wagenmaker for useful discussions. SBP would like to thank Steven Man, Andrea Bajcsy, and Ken Nakamura for their insightful discussions. We acknowledge support from the Toyota Research Institute (TRI) University 2.0 program.
\bibliographystyle{plainnat}
\bibliography{refs}
\newpage
\tableofcontents

\appendix
\newpage 


\section{A Practitioner's Guide to \OGPO{}}
\label{app:practitioners_guide}
In this section, we enumerate key design decisions, diagnostic tools, and configurations to serve as a reference for practitioners deploying \OGPO{} on new tasks. We defer the pseudocode to \Cref{app:pseudo} and the low level hyperparameters to \Cref{app:hyperparameters}

\RETURNN{Add Jax stuff}
\subsection{Key Design Decisions}
While a large set of hyperparameters remain static across all our experiments, some configurations might have a large impact on \OGPO{}'s performance on tasks beyond the scope of this paper. We list each item by descending priority level denoted by its high level description followed by the variable name in the official code base. 

\colorpar{0. Action Chunking Conventions and Critic Update}
\label{app:action_chunk}
The main paper denotes each action chunk $a_{t:t+h-1}$ simply as $a_t$ for simplicity. Here we describe how this affects our computation of reward when used to 
train the resulting Q-function. Let us consider a standard MDP formulation where $s_t$ is the state at current step, and $a_{t:t+h-1}$ denotes the action chunk.  We follow the value backup formulation proposed in Q-chunking \citep{li2025reinforcement}, where the target uses an $h$-step return over the chunk and bootstraps from the value of the next action chunk at state $s_{t+h}$, with $a_{t+h:t+2h} \sim \pi_\theta(\cdot \mid s_{t+h})$ and $\bar{\theta}$ denoting the parameters of the target network. We use this loss to train the critic for all our off-policy methods, including \OGPO{}, \QC{}, \DSRL{}, and \EXPO{}:

\begin{align}
\Lcritic(\theta)=\mathbb{E}_{s_t, a_{t: t+h}, s_{t+h} \sim \cB}\left[\left(Q_\theta\left(s_t, a_{t: t+h}\right)-\underbrace{\sum_{t^{\prime}=1}^h \gamma^{t^{\prime}} r_{t+t^{\prime}}}_{\text{effective reward}}-\gamma^h \Qtarg\left(s_{t+h}, a_{t+h: t+2 h}\right)\right)^2\right].
\end{align}

\subsubsection*{Algorithmic Choices}
\colorpar{1. Behavior-cloning regularization from the success buffer (\texttt{bc\_coeff}).} The total objective $\Ltotal = \Lppo + \lambda_{\textsc{bc}}\Lbc$ (\Cref{eq:OGPOpl}) anchors the policy to actions from $\succb \subseteq \rb$ — the subset of replay-buffer transitions belonging to successful episodes. The regularizer is asymmetric: it raises the likelihood of empirically successful actions but never lowers the likelihood of failed ones, so $\Lbc$ contributes a strict lower bound on the modes $\Lppo$ is allowed to abandon. Empirically (\Cref{fig:ogpo_core_ablations}) this is the single most consequential modification distinguishing \OGPO{} from \OGPOplus{}.In all experiments, we typically select $\lambda = 1.0$.

\colorpar{2. Conservative advantages (\texttt{adv\_strategy=conservative}, \Cref{eq:adv-conservative}).} The conservative advantage $\advcons_i$ is non-zero \emph{if and only if all $M$ ensemble members agree on the sign of $\advindie$}, in which case it takes the smallest magnitude consistent with that sign. Two consequences follow: (i) actions on which the ensemble disagrees produce no policy gradient, so the policy is updated only along directions of ensemble consensus; (ii) on directions of consensus, the magnitude is bounded by the most pessimistic Q-function, reducing the impact of outliers in the initial stages of online RL. This significantly mitigates the dip in policy evaluation and yields stable policy extraction. 

\colorpar{3. Critic aggregation for $\Qtarg$ and Best-of-$N$(\texttt{q\_agg}).}
\label{app:td_loss}
As referenced in Algorithm \ref{alg:critic_update}, \OGPO{} updates the critic ensemble by minimizing the Temporal Difference (TD) error. To calculate the target values, we employ an ensemble of $M$ target critic networks. The specific method for aggregating these target predictions is determined by the configuration flag $\criticflag$:
\begin{align}
    \Qtarg(s',a') = \begin{cases}
        \min\{Q_{\phi_{i_1}}(s',a'), Q_{\phi_{i_2}}(s',a')\} & \criticflag = \criticflagsub\\
         \min_{i \in [M]} Q_{\phi_{i}}(s',a')  & \criticflag = \criticflagmin\\
        \frac{1}{M}\sum_{i=1}^{M} Q_{\phi_{i}}(s',a') & \criticflag = \criticflagmean\\\end{cases}\label{eq:ytarg}
\end{align}

The setting of $\criticflag$ is optimized per environment (see \ref{sec:experiments}). The $\criticflagmin$ flag uses the minimum all $Q$ networks, which is more aggressively curtails overestimation. The $\criticflagmean$ flag uses the mean, which is less aggressive. Many works have found $\criticflagsub$ to be a happy medium: we take the minimum of two critic networks whose indices $i_1,i_2$ are sampled uniformly from the ensemble $\{1,\dots,M\}$, individually per action. Note that critic training is agnostic to the GCP structure of the policy (\citet{mark2024policy}).

\Cref{eq:ytarg} aggregates the critic ensemble  $\{Q_{\bar\phi_m}\}_{m=1}^{M}$ via $f \in \{\texttt{mean},\,\texttt{min},\,\texttt{subsample}\}$. Across almost all tasks, we find $\texttt{subsample}$ being the best strategy for $\Qtarg$ computation when using synchronous Jax updates, but $\texttt{mean}$ to work best using asynchronous updates. Our experiments are run on using synchronous updates. In both cases, we also find \texttt{subsample} to work optimally for selected the Best-of-$N$ actions \Cref{eq:a_plan}.

\colorpar{4. ODE-to-SDE conversion (\texttt{error\_correct\_sde\_to\_ode}).} 
In \OGPO{}, we add Gaussian noise of standard deviation $\sigma_\tau$ at each flow step to (1) ensure non-singular likelihoods thereby (2) facilitating exploration during online RL. Naively adding isotropic noise to the deterministic update $a_{t,k+1} = a_{t,k} + v_\theta(a_{t,k}, t_k \mid s_t)\Delta t$ causes distribution shift through the denoising chain, so the SDE-inferred policy visits different states than the ODE-inferred policy. Following \citet{albergo2023stochastic}, we instead use a marginal path-preserving SDE formulation that adds a score-based drift correction $\tfrac{\sigma_\tau^2}{2}\nabla\log p_\tau(x_\tau)$. In practice, training a separate score network (as in \citet{liu2025flow}) would require modifying the BC pretraining objective, which is prohibitive for pre-trained VLAs. We instead reparameterize the score through the policy and use a tapering noise schedule $\sigma_\tau = \sigma_\mathrm{init}\sqrt{1-\tau}$, which avoids the $\tau=1$ singularity and yields the simple, numerically stable correction term
\begin{align}
    c = \tfrac{\sigma_\mathrm{init}^2\bigl(\pi_\theta(x_\tau,\tau)\tau - x_\tau\bigr)}{2}.
\end{align}
See \Cref{app:ode_correction} for the full derivation.

\colorpar{5. Warmup Phase}
In the code accompanying \OGPO{}, we facilitate an additional \emph{warmup}-phase to pretrain Q-functions. We provide three warmup options:
\begin{enumerate}
    \item Warm-Start RL \citep{zhou2025efficient} with Calibrated Q-Learning (CalQL) \citep{nakamoto_cal-ql_2024}.
    \item Q-function warmup via TD error using $\pi_\mathrm{BC}$ rollouts.
    \item Q-functions pretrained by regressing MC returns using $\pi_\mathrm{BC}$ rollouts.
\end{enumerate}

For the tasks considered in the paper, we generally observe warmup not being critical to policy improvement. The use of Conservative Advantages and SFT via Success Buffer have a much higher impact on \OGPO{}'s training stability and sample efficiency.


\RETURNN{Add this}

\subsubsection*{Hyperparameters}
\colorpar{1. Group size $\Ngroup$ (\texttt{grpo\_num\_samples}).} We rollout $\Ngroup$ trajectories in parallel from a single $s_t$ to compute a mean value estimate for advantage computation in \Cref{eq:ppo_internal}. Larger $\Ngroup$ values result in higher exploration and diversity of information points at each update at the cost of compute. We find $\Ngroup=32$ to be a sweet spot across all our experiments.

\colorpar{2. PPO clip $\epsilon$ (\texttt{clip\_epsilon}).} The Annealed Importance Sampling ratio $\omega$ computed in \Cref{eq:ppo_internal} is sensitive to small perturbations in the likelihoods of each denoising step of the GCP being used. For 10-step flow policies, we find a clipping value of $\epsilon=0.01$ to work best for stable policy extraction. However, practitioners might need to experiment with this ratio depending on their GCP policy parameterization.

\colorpar{3. Update-to-data ratios} We provide three key update-to-data (UTD) ratios -- \texttt{utd\_warmup} (number of critic updates per base policy rollout step), \texttt{utd\_q}(number of critic updates per online policy rollout step), and \texttt{utd\_pi}(number of actor updates per online policy rollout step). Although a UTD of 1 works across the board, they can be tweaked individually depending on the task setting.

\colorpar{4. Exponential Moving Average}
\label{app:ema}
For all GCP instantiations within \OGPO{}, we maintain an Exponential Moving Average (EMA) of the policy weights, denoted as $\thetaema$. At every training step, after updating $\theta$, we update $\thetaema$ via:
\begin{align}
    \thetaema \leftarrow \alpha \thetaema + (1 - \alpha) \theta,
\end{align}
where $\alpha$ is a decay rate we typically set $\alpha= 0.995$. For \OGPO{}, the EMA serves a dual purpose beyond standard stability. First, it acts as the reference policy $\piold$ in the PPO importance sampling ratio (\cref{eq:ppo_internal}), ensuring that updates are constrained relative to a stable baseline rather than the rapidly changing online policy. Second, for the planning component in \OGPOplus{}, trajectories for Best-of-N ranking are sampled using $\pi_{\thetaema}$ to ensure stability in the candidate actions.

\RETURNN{Add these}
\section{Pseudocode}\label{app:pseudo}
\begin{algorithm}[H]
\caption{\OGPOplus}
\label{alg:ogpo}
\begin{algorithmic}[1]
\STATE $\piigp, Q_{\phi_{1\dots M}}, \rb \gets \emptyset
, \succb \gets \emptyset
$
\STATE $\theta_{\targ} \gets \theta$, $\phi_{\targ_i} \gets \phi_i \quad \forall i\in \{1,2,\dots M\}$
\FOR{iteration = 1, 2, \dots}
    \STATE Initialize state $s_{t=0} = s_0$ in $\mdpenv$
    \STATE $\mathcal{T}_{\text{ep}} \gets \emptyset$ \algcomment{Temporary episode buffer}
    \WHILE{not $\done$}
        \STATE $(s,a,r,s',\done) \gets \sample$ from the environment
        \STATE $\rb \gets \rb \cup \{(s,a,r,s',\done)\}$
        \STATE $\mathcal{T}_{\text{ep}} \gets \mathcal{T}_{\text{ep}} \cup \{(s,a,r,s',\done)\}$
        \STATEX \algcomment{\% Update critic and policy}
        \FOR{epoch $ = 1, 2,\dots,\utd$}
        
            \IF{$\useoffline$}
                \STATE $\rbbatch \sim \{ \offlineratio \rboff \cup (1 - \offlineratio)\rb \}$
            \ELSE
                \STATE $\rbbatch \sim \rb$
            \ENDIF
            \STATE $\sbbatch \sim \succb$ \textbf{if} $\succb \neq \emptyset$
            \STATE \updateq$(\rbbatch)$
            \STATE \updateigp$(\rbbatch, \sbbatch)$
            \STATEX \algcomment{\%Update target networks:}
            \STATE \quad $\phi_{\targ, i} \gets (1-\tau)\phi_{i} + \tau\phi_{\mathrm{\targ},i} \forall i \in {1,\dots,M}$
            \STATE \quad $\theta_{\targ} \gets (1 - \tau)\theta + \tau\theta_{\targ}$ 
        \ENDFOR
    \ENDWHILE
    \IF{episode successful}
        \STATE $\succb \gets \succb \cup \mathcal{T}_{\text{ep}}$ \algcomment{$\succb \subseteq \rb$}
    \ENDIF
\ENDFOR
\STATE \RETURN converged policy $\pi_\theta$
\end{algorithmic}
\end{algorithm}

\begin{algorithm}[H]
\caption{Initialization}
\label{alg:initialize}
\begin{algorithmic}[1]
\STATE \textbf{Function} $\initialize(\mathcal{D}_\mathrm{off})$
    \STATE \COMMENT{\% Policy Initialization}
    \STATE Pre-train GCP $\piigpbc$ on $\mathcal{D}_\mathrm{off}$ using BC loss $\mathcal{L}_\mathrm{BC}(\theta)$
    \STATE $\piigp \gets \piigpbc$
    
    \STATE \COMMENT{\% Critic Initialization}
    \STATE Initialize ensemble of Q functions $Q_{\phi_{1\dots M}}$
    \IF{$\useofflinerl$}
        \STATE Pre-train $Q_{\phi_{1\dots M}}$ on $\mathcal{D}_\mathrm{off}$ using $\mathcal{L}_\mathrm{critic}$ \COMMENT{Optional offline RL}
    \ENDIF
    
    \STATE \COMMENT{\% Buffer Initialization}
    \STATE $\rb \gets \emptyset$
    \STATE $\succb \gets \emptyset$
    
    \STATE \COMMENT{\% Warmup Rollouts}
    \FOR{episode = 1, \dots, $\nwarmup$}
        \STATE Roll out $\piigpbc$ in $\mdpenv$, collect transitions
        \STATE $\rb \gets \rb \cup \{(s,a,r,s',\done)\}_{\text{episode}}$
        \IF{episode successful}
            \STATE $\succb \gets \succb \cup \{(s,a,r,s',\done)\}_{\text{episode}}$
        \ENDIF
    \ENDFOR
    \IF{$\warmupcritic$}
        \FOR{step = 1, \dots, $N_\mathrm{critic\_warmup}$}
            \STATE $\rbbatch \sim \rb$
            \STATE \updateq$(\rbbatch)$ \COMMENT{Critic-only updates}
        \ENDFOR
    \ENDIF
    
    \STATE \RETURN $\piigp, Q_{\phi_{1\dots M}}, \rb, \succb$
\end{algorithmic}
\end{algorithm}

\begin{algorithm}[H]
\caption{Take A Step In The Environment}
\label{alg:sample}
\begin{algorithmic}[1]
\STATE \textbf{Function} $\sample(\st)$
\STATE $\done \gets \texttt{False}$
\STATE $\afirstt \sim \gaussinit$ 
    \FOR{$\igpiter = \igpfirst, \dots, \igplast$}
        \STATE $\ainct \gets \bar{\pi}_{\theta_{\targ}}(\aiter, \igpiter, \st)$
    \ENDFOR
    \STATE $r, \stnext \gets$ Execute $\alastt$ in environment
    \IF{$\stnext$ is terminal}
        \STATE$\done \gets \texttt{True}$
    \ENDIF
    \STATE \RETURN $(\st, \alastt, r, \stnext, \done)$
\end{algorithmic}
\end{algorithm}

\begin{algorithm}[H]
\caption{Critic Update}
\label{alg:critic_update}
\begin{algorithmic}[1]
\STATE \textbf{Function} \updateq($\rbbatch$)
    \STATE $(\st, \alastt, r, \stnext, \done) \gets \rbbatch$
    \STATEX With $\theta$ frozen:
    \INDSTATE $\alastnextt \gets \pi_{\theta_{\targ}}(\cdot \mid \stnext)$
    \INDSTATE $y \gets r + \gamma \cdot \mathbb{I}[\textbf{not } \done] \cdot \Qtarg(\stnext, \alastnextt)$ \COMMENT{Ref. Eq.~\ref{eq:ytarg}}
    \STATEX Update $\phi_{1,\dots,M}$ via gradient descent:
    \INDSTATE $\nabla_{\phi_i} \frac{1}{|\rbbatch|} \sum_{\rbbatch} \left( Q_{\phi_i}(\st, \alastt) - y \right)^2$ for $i = 1,\dots,M$
\end{algorithmic}
\end{algorithm}

\begin{algorithm}[H]
\caption{GCP Update}
\label{alg:policy_update}
\begin{algorithmic}[1]
\STATE \textbf{Function} \updateigp($\rbbatch, \sbbatch$)
    \STATEX On-Policy PPO Update
    \INDSTATE $\st \gets \rbbatch$
    \INDSTATE Sample $G$ actions: $\{\bar{\tau}^{(g)}\}_{g=1}^{G} \sim \pi_{\theta_{\targ}}(\cdot \mid \st)$
    \INDSTATE $\advgrpo = \Qtarg(\st, \alastt^{G}) - \mu(\Qtarg(\st, \alastt^{G}))$ 
    \INDSTATE $\ratio = \frac{\prod_{\igpiter=\igpfirst}^{\igplast} \bar{\pi}^{\group}_\theta(\ainct \mid \aiter, \igpiter, \st)}{\prod_{\igpiter=\igpfirst}^{\igplast}\bar{\pi}^{\group}_{\theta_{\targ}}(\ainct \mid \aiter, \igpiter, \st)}$
    \INDSTATE $\mathcal{L}_{\text{PPO}}(\theta) = \mathbb{E}_{\bar{\tau} \sim \pi^{\group}_{\theta_{\targ}}} \left[\min\left(\ratio \cdot \advgrpo, \text{clip}(\ratio, 1-\epsilon, 1+\epsilon) \cdot \advgrpo\right)\right]$
    
    \STATEX BC Update from Success Buffer
    \INDSTATE $(\stsucc, \atsucc) \gets \sbbatch$
    \INDSTATE $\mathcal{L}_{\text{BC}}(\theta) = \textsc{BCLoss}(\bar{\pi}_\theta(\cdot \mid \stsucc), \atsucc)$ \COMMENT{GCP-specific}
    
    \STATEX Combined Update
    \INDSTATE $\mathcal{L}_{\text{total}}(\theta) = \mathcal{L}_{\text{PPO}}(\theta) + \lambda_{\text{BC}} \mathcal{L}_{\text{BC}}(\theta)$
    \STATE Update $\theta$ via gradient descent on $\mathcal{L}_{\text{total}}(\theta)$
\end{algorithmic}
\end{algorithm}

\section{Generative Control Policies (GCPs): A Unifying Abstraction}
\label{app:igp_instantiations}

We propose a unifying abstraction for a broad family of popular parameterizations of control policies that we call \emph{Generative Control Policies}, or \colorbold{GCP}s. GCPs represent a stochastic policy $\pi_\theta(\cdot\mid s)$ as a series of iterative computation steps, defined by a mapping $\piigp: S \times A \times \mathbb{N}$. Given a state $\st$, the policy samples $\afirstt \sim \piigp\prn{\cdot \mid \aitert = \emptyset, \igpiter=\igpfirst,s_t}$. From then, we sample $\ainct \sim \piigp\prn{\cdot \mid \aitert,\igpiter,s_t}$. The final action proposed is an action $\alastt$. We compactly denote the distribution of this action given the observation as $\alastt \sim \pi_\theta(\cdot \mid s_t)$, turning the GCP into a standard policy. 
Our iteration conventions are \emph{decreasing} in $K$, following typical convention for diffusion models. We also drop $t$ subscripts when clear from context.
\vspace{-.3em}
\colorpar{Examples of GCPs:}In addition to iterative computation, the only other requirement is that the conditional likelihoods, $\log \piigp(\ainct=a\mid\st, \aitert,\igpiter)$ are efficiently represented. A number of popular parameterizations produce actions iteratively and satisfy this mild requirement:

\iftoggle{arxiv}
{
\begin{itemize}
    \item \textbf{Diffusion Policies} \citep{chi2023diffusion} use  Denoising Diffusion Probabilistic Models (DDPMs) \citet{ho2020denoising}. Instantiated as an GCP, these take in pairs $(s,a)$ as training data and iteratively add Gaussian noise to the actions through a forward process $q(a_{k+1}\mid a_k)$ and learn a function $\epsilon_\theta(\aiter,k,s)$ predicting the noise added  to convert $\xlast$ to $\xiter$. To produce an action, we sample $\afirstt\sim \Normal(O,\eye)$, and iteratively generate denoised samples with the following reverse process:
\begin{align}
    \ainc \sim \pibarddpm(\cdot\mid\aiter,k,s) := \Normal( \mu_k(\xiter,\epsilon_\theta(\aiter,k,s)), \sigma^2_k \mathrm{I}) \label{eq:ddpm_reverse}
\end{align}
    \item \textbf{Flow policies} are based on flow matching models. Given training pairs $(s,a)$, we sample noise $z \sim \Normal(0,\eye)$, and define the interpolant $a_{(\tau)}:=\tau a + (1-\tau)z$ with continuous noise index
    $\tau\in [0,1]$. We then learn a velocity field $v_\theta(a_{(\tau)},\tau,s)$ , these predict $\Exp[a-z\mid s, a_{(\tau)}]$. For $K$ discretization steps, we generate samples by initializing $\alast\sim\Normal(0,\eye)$ and discretizing an ordinary differential equation (ODE) which reverses the noising process $
    \ainc  := \aiter + \frac{1}{K}{v_\theta(a_k,k/K,s)} \label{eq:flow}$ 
In its stand form, $\ainc \mid \aiter,s$ is deterministic. Thus, to convert a flow policy into a proper GCP, for which \emph{likelihoods} are well-defined, we must add additional noise at each step (Reinflow \citep{zhang2025reinflow}). For a given choice of noise levels $\sigma_k^2$, this induces the GCP: 
\begin{align}
    \ainc \sim \pibarflow(\cdot\mid\aiter,k,s) := \Normal( v_\theta(\aiter,k/K,s), \sigma^2_k \eye) \label{eq:flow_reverse}
\end{align}

\item \textbf{Minimal Iterative Policies} (MIP) are two-step flow policies which yield a performance comparable to 10-step flow policies with the natural benefit of allowing much faster inference. We defer the formal definition to Appendix \ref{app:mip}

\end{itemize}
The GCP formalism encompasses a number of more recent policy parameterizations as well, such as
\begin{itemize}
    \item \textbf{Shortcut Policies} \citep{frans2024one}: Flow models with learnable step sizes that enable variable-length generation trajectories.
    \item \textbf{Tokenized Autoregressive Policies} (FAST \citep{pertsch2025fast}): Policies that tokenize continuous actions in Fourier space and generate them autoregressively as discrete sequences.
\end{itemize}
In the interest of brevity, we detail the above in the Appendix \ref{app:shortcut}, and Appendix \ref{app:fast} respectively. Conveniently, the GCP formalism abstracts away the details of these varying instantiations, allowing us to state all algorithms cleanly.  
}
{
\textbf{Diffusion Policies} \citep{chi2023diffusion} use  Denoising Diffusion Probabilistic Models (DDPMs) \citet{ho2020denoising}. Instantiated as an GCP, these take in pairs $(s,a)$ as training data and iteratively add Gaussian noise to the actions through a forward process $q(a_{k+1}\mid a_k)$ and learn a function $\epsilon_\theta(\aiter,k,s)$ predicting the noise added  to convert $\xlast$ to $\xiter$. To produce an action, we sample $\afirstt\sim \Normal(O,\eye)$, and iteratively generate denoised samples with the following reverse process: $\ainc \sim \pibarddpm(\cdot\mid\aiter,k,s) := \Normal( \mu_k(\xiter,\epsilon_\theta(\aiter,k,s)), \sigma^2_k \mathrm{I})$. 

\textbf{Flow policies} are based on flow matching models. Given training pairs $(s,a)$, we sample noise $z \sim \Normal(0,\eye)$, and define the interpolant $a_{(\tau)}:=\tau a + (1-\tau)z$ with continuous noise index
$\tau\in [0,1]$. We then learn a velocity field $v_\theta(a_{(\tau)},\tau,s)$ , these predict $\Exp[a-z\mid s, a_{(\tau)}]$. For $K$ discretization steps, we generate samples by initializing $\alast\sim\Normal(0,\eye)$ and discretizing an ordinary differential equation (ODE) which reverses the noising process $
    \ainc  := \aiter + \frac{1}{K}{v_\theta(a_k,k/K,s)} $ 
In its stand form, $\ainc \mid \aiter,s$ is deterministic. Thus, to convert a flow policy into a proper GCP, for which \emph{likelihoods} are well-defined, we must add additional noise at each step (Reinflow \citep{zhang2025reinflow}). For a given choice of noise levels $\sigma_k^2$, this induces the GCP: 
\begin{align}
    \ainc \sim \pibarflow(\cdot\mid\aiter,k,s) := \Normal( v_\theta(\aiter,k/K,s), \sigma^2_k \eye) \label{eq:flow_reverse}
\end{align}
}
While we have presented \OGPO{} in the context of \emph{flow-matching} policies, the algorithm is agnostic to the specific generative parameterization of the GCP, and applies directly to diffusion policies as well. Both flow-matching and score-based diffusion policies define an iterative denoising chain $a_t^K \to a_t^{K-1} \to \dots \to a_t^0$ from a base noise distribution to the action distribution; the only difference is the parameterization of the per-step transition (a learned velocity field $v_\theta$ for flow policies versus a learned score / $\epsilon$-prediction for diffusion). \OGPO{}'s key ingredients --- per-step likelihood evaluation along the denoising chain (\Cref{eq:ppo_internal}) and the SDE-based exploration noise correction (\Cref{app:ode_correction}) --- are derived from generic properties of the underlying SDE and therefore carry over unchanged to a diffusion-policy GCP, provided one substitutes the appropriate noise schedule and score parameterization.

We empirically verify this in \Cref{fig:ogpo_diffusion}, where we instantiate \OGPO{} on top of a diffusion-policy backbone and observe consistent improvement over BC pretraining, mirroring the trends we report for flow-policy backbones in the main paper. In practice, however, we predominantly default to flow-matching policies for our main experiments: flow policies admit substantially fewer denoising steps at inference time (typically $4$--$10$ versus $50$--$100$ for diffusion) while achieving comparable BC performance, which directly translates into faster environment rollouts and meaningfully reduced wall-clock cost for online RL. We therefore view diffusion-policy \OGPO{} as a drop-in alternative whenever the underlying VLA backbone is itself a diffusion model, and flow-policy \OGPO{} as the preferred default when inference compute is a bottleneck.

\subsection{OGPO with Diffusion Policies}
\label{app:ogpo_diffusion}
\begin{figure}[H]
    \centering
    \includegraphics[width=0.55\linewidth]{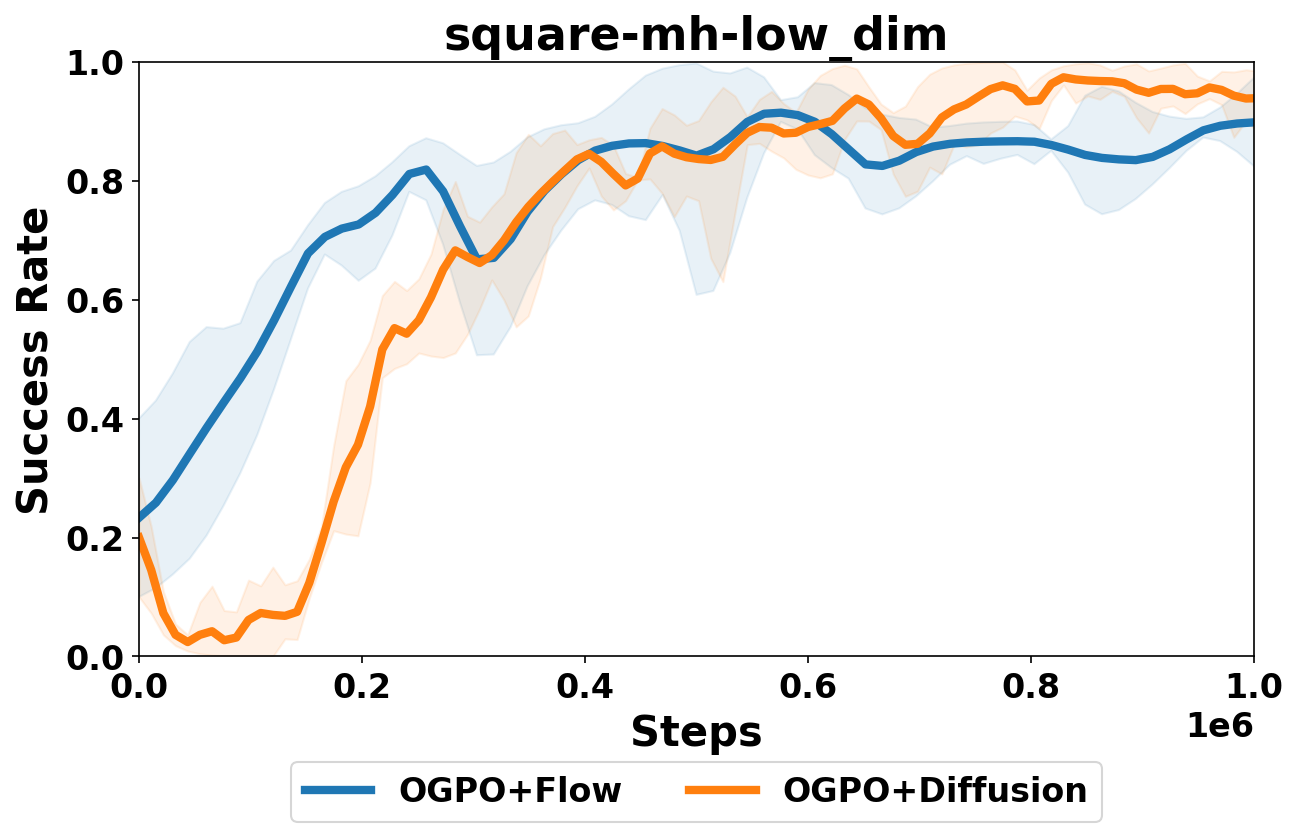}
    \caption{\textbf{\OGPO{} with diffusion policies.} \OGPO{} can successfully improve both flow policy and diffusion policy.}
    \label{fig:ogpo_diffusion}
\end{figure}

\OGPO{} can, in principle, be combined with any GCPs. Here, as an example, we illustrate its use in diffusion policies. We study this on the $\rmsquare$ task, where we pre-train a diffusion policy on the MH dataset and then apply online improvement with \OGPO{}. As shown in Figure~\ref{fig:ogpo_diffusion}, \OGPO{} successfully improves the diffusion policy to achieve mastery.

\subsection{Shortcut Policies}
\label{app:shortcut}
Shortcut policies \citep{frans2024one} are derived from flow-matching models conditioned on a step-size parameter $d$. The model $\piigp(a_t, t, d, o)$ learns to predict the next state of the flow $a_{t+d}$ by taking a shortcut of size $d$. This allows the policy to function as an GCP with a variable number of refinement steps $\igpfirst$. During pretraining, shortcut models utilize a self-consistency loss that enforces the property that one shortcut step of size $2d$ should be equivalent to two consecutive steps of size $d$:
\begin{align}
    \pi_\theta(a_t, t, 2d, o) \approx \frac{1}{2}\pi_\theta(a_t, t, d, o) + \frac{1}{2}\pi_\theta(a'_{t+d}, t+d, d, o)
\end{align}

\subsection{Minimal  Iterative Policy}
\label{app:mip}
Minimal  Iterative Policies (MIP) \citep{pan2025much} represent the simplest GCP instantiation that retains the performance benefits of flow-based policies. The key insight is that the success of generative control policies stems from combining \emph{Stochasticity Injection} during training with \emph{Supervised Iterative Computation}, rather than learning the distributions themselves. MIP uses only $\igpfirst=2$ denoising steps, with the first step computing $a_{t,1} \leftarrow \pi_\theta(\st, a_{t,2}=\bar{0}, t=0)$, then refining via $a_{t,0} \leftarrow \pi_\theta(\st, t^\star a_{t,1}, t^\star)$. The core insight being that merely learning the conditional mean is sufficient to match the performance of complex flow-matching policies, provided the refinement steps allow the policy to adhere to the expert action manifold.

Formally, MIP optimizes the following objective during pretraining, where $t^\star=0.9$ and $z \sim \mathcal{N}(0, I)$ is injected noise:
\begin{align}
    \mathcal{L}_{\text{MIP}}(\theta) = \mathbb{E}\left[ \| \pi_\theta(o, I_0=0, t=0) - a \|^2 + \| \pi_\theta(o, I_{t^\star}, t^\star) - a \|^2 \right],
\end{align}
where $I_{t^\star}$ is the interpolant between action $a$ and noise $z$.

\subsection{Tokenized Autoregressive Policies}
\label{app:fast}
Tokenized policies, such as those using the FAST tokenizer \citep{pertsch2025fast}, represent the action distribution via categorical distributions over a vocabulary of discrete tokens. FAST efficiently handles high-frequency continuous control data by applying a Discrete Cosine Transform (DCT) to action chunks, followed by quantization and Byte-Pair Encoding (BPE).

In this formulation, the GCP is an autoregressive Transformer $\piigp(z_k \mid z_{<k}, \st)$, where $z$ represents the sequence of discrete tokens corresponding to a compressed action chunk. The generative process iteratively samples tokens:
\begin{align}
    z_k \sim \text{Categorical}(\pi_\theta(\cdot \mid z_{<k}, o))
\end{align}
Unlike diffusion or flow policies where iteration occurs in continuous action space (refining the values), here iteration occurs in the token sequence space. In particular, this slightly deviates from the GCP formulation described in the main test by requiring conditioning on the whole token sequence $z_{<k}$. However, the light likelihoods in our PPO update in \Cref{eq:PPO_loss_group} can be easily modified to handle this setting, because $p(z_{1:k}) = \prod_k p(z_k \mid z_{<k})$. 


    
    
    




%

\newcommand\mdpbilevel{M_\textsc{Bilevel}}

\section{Bi-Level MDP}
\begin{figure}[H]
    \centering
    \includegraphics[width=0.99\linewidth]{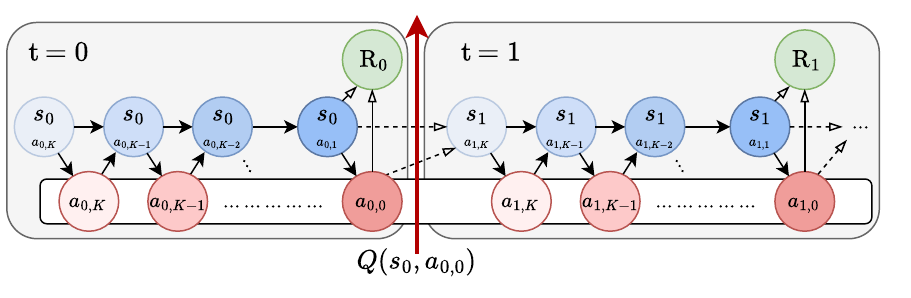}
    \caption{\textbf{Bi-level (two-layer) MDP construction.} Each environment step $t$ is expanded into $K$ inner action-generation steps indexed by $k\in\{K-1,\dots,0\}$. The environment transitions and rewards occur only at $k=0$, while for $k>0$ the state is unchanged and the inner action variable is updated.}
    \label{fig:bi_level_mdp_app}
\end{figure}
\label{app:bilevel}
We formulate the bi-level MDP (\Cref{fig:bi_level_mdp_app}), also called the two-layer MDP in \citep{ren2024diffusion}, by embedding the action-generation dynamics into the environment dynamics.
This yields an augmented MDP $\mdpbilevel$ whose trajectory explicitly interleaves environment time with the $K$ action-generation steps.

Recall the environment MDP $\mdpenv := (\statespace, \actionspace, P_0, P, R,\gamma)$ defined in \Cref{sec:prelim}. In $\mdpbilevel$, we index time by pairs $(t, k)$, where $t$ denotes the environment step and $k \in \{0, \dots, K-1\}$ denotes the action-generation step, with $k=0$ corresponding to executing the final action in the environment. We map $(t, k)$ to a single time index via $\bar{t}(t, k) = tK + (K - k - 1)$, so that the sequence $\bar{t}(t,K-1), \bar{t}(t,K-2), \ldots, \bar{t}(t,0)$ corresponds to the $K$ generation/execution steps within environment step $t$.
The state, action, and reward in $\mdpbilevel$ are defined as
\begin{align*}
    \bar{s}_{\bar{t}(t, k)} = (s_t, a_{t, k+1}), \quad \bar{a}_{\bar{t}(t, k)} = a_{t, k}, \quad \bar{R}_{\bar{t}(t, k)}(\bar{s}_{\bar{t}(t, k)}, \bar{a}_{\bar{t}(t, k)}) = \begin{cases}
        0, &k > 0, \\
        R(s_t, a_{t, 0}), &k=0.
    \end{cases}
\end{align*}

Importantly, rewards are emitted only at indices corresponding to executing the environment action, i.e., when $a_{t, 0}$ is taken. The initial distribution factorizes as $\bar{P}_0 = P_0 \otimes P_{\textsc{Action}, 0}$, where $s_0 \sim P_0$ is the initial environment state and $a_{0, K}$ is sampled independently from $P_{\textsc{Action}, 0}$, the initialization distribution for the action-generation process at $t = 0$.

Finally, the transition kernel is given by
\begin{align*}
    \bar{P}(\bar{s}_{\bar{t}+1} \mid \bar{s}_{\bar{t}}, \bar{a}_{\bar{t}}) = \begin{cases}
        \updelta_{(s_t, a_{t, k})}, & \bar{t} = \bar{t}(t, k),\ k > 0 \\
        P(\cdot \mid s_t, a_{t, 0}) \otimes P_{\textsc{Action}, t+1} & \bar{t} = \bar{t}(t, k), \ k = 0
    \end{cases},
\end{align*}
where $P_{\textsc{Action}, t}$ (for $t \ge 0$) denotes the initialization distribution for $a_{t, K}$.
Intuitively, when $k > 0$, the transition advances the iterative action-generation process by moving from $(s_t, a_{t,k+1})$ to $(s_t, a_{t,k})$ while keeping the environment state fixed; when $k=0$, it executes $a_{t ,0}$ in the environment, samples $s_{t+1} \sim P(\cdot \mid s_t, a_{t,0})$, and re-initializes the next inner process by sampling $a_{t+1,K} \sim P_{\textsc{Action},t+1}$.
\section{Derivations}
\subsection{Policy Gradient Loss} \label{app:pg_loss}
An optimal policy parameterized by $\theta$ can be obtained by maximizing an objective function that computes the expected reward over a trajectory $\sampletau$. Mathematically, $\theta^\star$ = $\argmax_\theta J(\theta)$, where $J(\theta)=\Exp_{\sampletau}\left[ \omega(\tau) \right]$. Hence, the policy gradient objective is given as:
\begin{align}
    \nabla_\theta J(\theta)=\Expop_{\sampletau} \left[ \nabla_\theta \log \pi_\theta(\tau)\omega(\tau) \right]
\end{align}

However, there are two main challenges which make the classical PG loss formulation challenging to converge in practice. (1) Policies parameterized as neural networks can only change a little with each gradient step. (2) High variance environments require a very large number of rollouts to obtain $\pi^\star$, which is prohibitively expensive and potentially unsafe to do on real robots. As proposed by \citep{schulman2015trust}, high variance can be mitigated by estimating an expectation under a distribution from an older policy $\piold$ using importance sampling (IS). This implies use of short horizon replay buffers where actions sampled under $\piold$ are reused to compute IS against $\pi_\theta$. 
This modifies the PG objective as follows:
\begin{align}
\begin{split}
    \nabla_\theta J(\theta)
                &=\Expop_{\sampletau} \left[ \frac{\pi_\theta(\tau)}{\pi_{\theta_{\mathrm{old}}}(\tau)} \nabla_\theta \log \pi_\theta(\tau)\omega(\tau) \right]\\
                &=\Expop_{\sampletau} \left[\left(\sum_{t=t}^{T} \nabla_\theta \log \pi_\theta(a_t \mid s_t) \right) \left(\prod_{t=1}^{T} \frac{\pi_\theta(a_t \mid s_t)} {\pi_{\text{old}}(a_t \mid s_t)} \right) \left( \sum_{t=t}^{T} r(s_t, a_t) \right) \right]
\end{split}
\end{align}

However, the product of importance weights in the trajectory-level estimator leads to vanishing probability products for long horizons $T$. The objective is reformulated using state-action marginals. This separates the expectation over states (dependent on transition dynamics) from the expectation over actions (dependent on the policy):
\begin{align}
    J(\theta) = \sum_{t=1}^T \mathbb{E}_{s_t \sim \rho_{\theta_{\text{old}}}(s_t)} \left[ \frac{\rho_\theta(s_t)}{\rho_{\theta_{\text{old}}}(s_t)} \mathbb{E}_{a_t \sim \pi_{\theta_{\text{old}}}(\cdot|s_t)} \left[ \frac{\pi_\theta(a_t|s_t)}{\pi_{\theta_{\text{old}}}(a_t|s_t)} r(s_t, a_t) \right] \right]
\end{align}

Calculating the state density ratio $\frac{\rho_\theta(s_t)}{\rho_{\theta_{\text{old}}}(s_t)}$ is difficult as it requires knowledge of the system dynamics. Therefore, TRPO and PPO introduce a simplification by ignoring this term. This results in a biased estimator, but the bias is negligible provided $\pi_\theta$ remains close to $\pi_{\theta_{\text{old}}}$. The resulting surrogate objective maximizes the probability of actions with high rewards (or advantages) relative to the old policy:
\begin{align}
    J(\theta) \approx \sum_{t=1}^T \mathbb{E}_{\substack{s_t \sim \rho_{\theta_{\text{old}}} \\ a_t \sim \pi_{\theta_{\text{old}}}}} \left[ \frac{\pi_\theta(a_t|s_t)}{\pi_{\theta_{\text{old}}}(a_t|s_t)} r(s_t, a_t) \right]
\end{align}

Classically, algorithms like PPO parameterize the policy $\pi_\theta(a|s)$ as a unimodal Gaussian distribution $\mathcal{N}(\mu_\theta(s), \Sigma)$. This yields a unimodal importance sampling ratio at every timestep $t$, which naturally struggles to model the multimodal action distributions necessary during RL exploration for complex manipulation tasks. Conversely, the total probability $\piigp(\alastt \mid \st)$ in a GCP is the product of the transition probabilities along the generation steps $\igpiter$. This likelihood is given as: $\pi_\theta(a_{t,0} \mid s_t) = \prod_{k=1}^{K} \pi_\theta(a_{t,k} \mid s_t)$

Substituting this into the standard PPO objective requires computing the ratio of these products. While trajectory-level importance sampling is unstable for long environment MDP chains (where $T \approx 400 - 1000$), the denoising MDP horizon of the generative process can be sufficiently short (typically $\igpfirst \leq 10$) 

Assuming the current policy $\pi_\theta$ and the reference policy (typically an Exponential Moving Average, $\pi_{\text{EMA}}$) are close, we extend the TRPO formulation to the GCP chains to compute the Annealed Importance Sampling (AIS) ratio:
\begin{align}
    \ratio := \prod_{k=1}^K\frac{ \pi_\theta(a{^{k-1}} \mid s,a^k) }{\pi_{\thetaema}(a^{k-1} \mid s,a^k) }    
\end{align}

The probability of the final executed action is the joint probability of the entire chain: $\pi_\theta(\alastt \mid \st) = \prod_{\igpiter=\igpfirst}^1 \pi(\ainct \mid \aitert,\st)$.
We substitute the Monte Carlo return $\omega(\tau)$ with the advantage $\hat A$, which yields the final \OGPO() objective described in \Cref{eq:ppo_internal}. When multiplied with the advantage $\hat A$, the resulting gradients propagate to every step $\igpiter$, updating each in proportion to its contribution to the final action's probability. This end-to-end formulation ensures that generating a high-value action $\alastt$ requires coherent refinement at every step $\aitert$ if the GCP.

\newcommand{\Xsde}{X^{\textsc{sde}}}
\newcommand{\vsde}{v^{\textsc{sde}}}
\newcommand{\noiset}{\sigma_\tau}

\subsection{ODE-to-SDE Exploration Noise Correction}
\label{app:ode_correction}
In order to have nondegenerate likelihoods, we ned to convert deterministic flow inference into a stochastic process. Naively, we could add Gaussian noise (as in \citet{zhang2025reinflow}), but the addition of isotropic noise introduces distribution shift between the original action distribution and the noise-augmented action distribution. We note that the same approach is also adopted by \cite{liu2025flow}.

Specifically, we follow \citet{albergo2023stochastic}, which provides a principled conversion from ODE inference (as in standard flow models) to an SDE). Consider a continuous-time ODE
\begin{equation}
\rmd X_\tau = v_\theta( X_\tau,\tau)\rmd \tau,
\label{eq:probability_flow}
\end{equation}
where $v_\theta(x_\tau,\tau)$ is the flow velocity field.  Next for a time varying diffusion standard deviation $\noiset$, define an stochastic differential equation (SDE)
\begin{equation}
\rmd \Xsde_\tau = \underbrace{\left[ v( \Xsde_\tau,\tau) + \frac{\noiset^2}{2} s( \Xsde_\tau,\tau) \right]}_{\vsde(\Xsde_\tau,\tau)} \rmd \tau + \noiset \, \rmd W_\tau,
\label{eq:forward_sde}
\end{equation}
where $s_\tau( x) = \nabla_x \log \rho_\tau(x)$ is the score function, and where $\rho_\tau$ is the marginal distribution of $X_\tau$. 
\begin{proposition}[\citet{albergo2023stochastic}]\label{prop:same_marginal} For every time $\tau$, the marginal distribution of $X_\tau$ and $\Xsde_\tau$ are the same.
\end{proposition}
The key insight is that the correction in the SDE drift $\vsde_\tau = v_\tau + \epsilon_\tau s_\tau$ directly offsets the effect of the Brownian drift. 
Furthermore, by Tweedie's formulation, the score function can be computed as
\begin{align}
    s(\tilde x_\tau,\tau)  = \frac{1}{\sigma}(\Exp[Z \mid X_\tau + \sigma Z = \tilde x_\tau]), \quad Z \sim \Normal(0,\eye)
\end{align}
In particular, $s_\tau = \frac{1}{\sigma}z_\tau$, where 
\begin{align}
    z_\tau \in \argmin_{z(\cdot)} \Exp\|z_\tau(X_\tau + \sigma Z) - Z\|^2.
\end{align}

\colorpar{Specialization to OGPO via discretization} 
Given the SDE with the score correction during online RL: 
\begin{align}
    dX_\tau = \left[\piigp(x_\tau,\tau) + \underbrace{\frac{\sigma_\tau^2}{2}\nabla\log p_\tau(x_\tau)}_{c}\right]d\tau + \sigma_\tau dW_\tau,
\end{align}


\newcommand{\alphat}{\alpha_\tau}
\newcommand{\betat}{\beta_\tau}
\newcommand{\at}{a_\tau}
\newcommand{\beet}{b_\tau}
and noise schedules $\alphat, \betat$, the score of the Gaussian probability path $p_\tau(x|z)\sim\Normal(x_\tau, \alphat z,\betat^2 \eye_d)$ at timestep $\tau$ is given as
\begin{align}
    \nabla\log p_\tau(x|z) = -\frac{1}{\betat^2}x + \frac{\alphat}{\betat^2}z.
\end{align}
Reparameterizing the policy wrt the score function gives:
\begin{align}
    \piigp(x_\tau,\tau) = \left(\betat^2\frac{\dot{\alphat}}{\alphat} - \dot{\betat}\betat\right)\nabla\log p_\tau(x_\tau) + \frac{\dot{\alphat}}{\alphat}x_\tau
    \label{eq:reparam_trick}
\end{align}

For simplicity, we set $\alphat=\tau,\betat=1-\tau$. This simplifies \Cref{eq:reparam_trick} to
\begin{align}
    \nabla\log p_\tau(x|z) = \frac{\piigp(\x_\tau,\tau)\tau - x}{1-\tau}.
\end{align}
Hence, the score correction term $c$ begets
\begin{align}
    c &= \frac{\sigma_\tau^2}{2}\nabla\log p_\tau(x|z) \\
    &= \frac{\sigma_\tau^2(\piigp(\x_\tau,\tau)\tau - x)}{2(1-\tau)}.
\end{align}
This reparameterization trick obviates the need for computing score function of the SDE policy, however presents an unstable divide by zero operation at $\tau=1$, i.e. the last denoising step of the policy in practice. One way to mitigate this is to consider $\alphat=\tau,\betat=\sqrt{1-\tau^2}$ as is done by \citet{liu2025flow}. However, this requires modification of the BC pretraining objective which is prohibitively expensive for pre-trained VLA models. 

Therefore, we instead propose a tapering noise schedule $\sigma_\tau = \sigma_\mathrm{init}\sqrt{1-\tau}$. This results in the score correction term
\begin{align}
    c = \frac{\sigma_\mathrm{init}^2(\piigp(\x_\tau,\tau)\tau - x)}{2},
\end{align}
that prevents numerical instability at the final step of the SDE rollout. We find this tapered noise schedule-based SDE flow policy to be the most stable implementation for \OGPO{}. We however note that the runs presented in the paper were generated with a constant noise schedule, but our open sourced codebase provides the most optimal implementation of the SDE-flow policy. 


\RETURNN{We provide a few results to compare \OGPOplus{} with constant noise and tapered noise schedule. \TODO Add more results :D}

\subsection{BC on $\succb$ as an ELBO Barrier in Forward-KL Space}
\label{sec:bc_elbo_barrier}

In addition to the policy gradient and pessimism terms described above, $\OGPOplus$ also incorporates a behavior cloning (BC) loss against the success buffer $\succb$. We show here that this BC term serves as a tractable lower bound on the forward KL divergence $D_\text{KL}(\succb \| \pi_\theta)$, thereby aligning $\pi_\theta$ to the modes of successful actions and preventing the policy from dropping their probability mass.

Consider a flow policy $\pi_\theta$ with velocity field $v_\theta(a_\tau, \tau, s)$ trained via the linear interpolant $a_\tau = (1-\tau)\epsilon + \tau a_1$ with target $a_1 - \epsilon$. For any target action distribution $q$, the flow-matching loss admits the bias-variance decomposition
\begin{align}
\mathcal{L}_\text{FM}(\theta; q) &= \Exp_{\tau,\epsilon,a_1 \sim q}\!\left[\|v_\theta(a_\tau, \tau, s) - (a_1 - \epsilon)\|^2\right] \nonumber \\
&= \underbrace{\Exp\!\left[\|v_\theta - v_q^\star\|^2\right]}_{\theta\text{-optimizable}} + \underbrace{\Exp\!\left[\|v_q^\star - (a_1 - \epsilon)\|^2\right]}_{\theta\text{-independent constant } C(q)},
\end{align}
where $v_q^\star(a_\tau, \tau, s) := \Exp[a_1 - \epsilon \mid a_\tau, \tau, s]$ is the optimal velocity field. By \citet{albergo2023stochastic}, integrating $v_q^\star$ via the probability flow ODE in \Cref{eq:probability_flow} recovers $q$ as the terminal marginal at $\tau = 1$. The first term is therefore a tractable lower bound on the marginal forward KL:
\begin{align}
\mathcal{L}_\text{FM}(\theta; q) - C(q) = D_\text{KL}\!\left(q \,\|\, \pi_\theta\right) \geq 0.
\end{align}
This is an ELBO in the sense that an otherwise-intractable marginal KL --- the marginal densities of flow policies have no closed form --- is variationally bounded by a tractable squared-error regression loss.

Instantiating this with $q = \succb$ recovers the BC loss on the success buffer:
\begin{align}
\mathcal{L}_\text{BC}^\text{succ}(\theta) - C(\succb) = D_\text{KL}\!\left(\succb \,\|\, \pi_\theta\right).
\end{align}
Crucially, the outer expectation is taken under $\succb$: every successful action mode is visited at training time. If $\pi_\theta(\atsucc \mid s) \to 0$ for some $\atsucc \sim \succb$, the integrand $\log(\succb / \pi_\theta) \to \infty$ and the velocity-MSE penalty pulls $v_\theta$ back toward $v_{\succb}^\star$ at that point. This mode-preserving \emph{barrier} property characteristic of forward KL provides regularization via the BC term. Any action mode in $\succb$ that $\pi_\theta$ tries to abandon incurs an unbounded penalty. Given the policy gradient conditioning does not strongly pull the GCP distribution against the successful modes, especially in the early training stages, $\pi_\theta$ retains coverage over the full support of successful behaviors throughout online RL.

\section{Baselines}
\label{app:baselines}
In this section, we describe all baselines we compare to in detail. Throughout, we adopt of the action-chunking conventions of \Cref{app:action_chunk}. 

\subsection{Diffusion Policy Policy Optimization (\DPPO{}, \citet{ren2024diffusion}}
\label{app:baseline:dppo}

\DPPO{} fine-tunes diffusion policies by applying PPO directly to the bi-level MDP introduced in \Cref{app:bilevel}. In this construction, each inner denoising step induces an explicit (Gaussian) likelihood, enabling standard policy-gradient updates on the full trajectory in $\mdpbilevel$. \DPPO{} then instantiates the PPO clipping objective on $\mdpbilevel$.

Concretely, let $\bar{\pi}_\theta(\bar{a}_{\bar{t}} \mid \bar{s}_{\bar{t}})$ denote the policy on $\mdpbilevel$ (i.e., the diffusion reverse transition at each denoising step). Given trajectories collected from $\bar{\pi}_{\theta_{\mathrm{old}}}$ and advantage estimates $\hat{A}^{\bar\pi_{\theta_{\mathrm{old}}}}$, \DPPO{} maximizes the PPO clipped surrogate
\begin{align*}
    \Exp_{(s_{\bar{t}}, a_{\bar{t}}) \sim \bar\pi_{\theta_{\mathrm{old}}}} \Big[\min\Big(\frac{\bar\pi_{\theta}(a_{\bar{t}} \mid s_{\bar{t}})}{\bar\pi_{\theta_{\mathrm{old}}}(a_{\bar{t}} \mid s_{\bar{t}})}\ \hat{A}^{\bar\pi_{\theta_{\mathrm{old}}}}(s_{\bar{t}}, a_{\bar{t}}),\ \mathrm{clip}(\frac{\bar\pi_{\theta}(a_{\bar{t}} \mid s_{\bar{t}})}{\bar\pi_{\theta_{\mathrm{old}}}(a_{\bar{t}} \mid s_{\bar{t}})}, 1-\epsilon, 1+\epsilon) \hat{A}^{\bar\pi_{\theta_{\mathrm{old}}}}(s_{\bar{t}}, a_{\bar{t}})\Big)\Big].
\end{align*}
\DPPO{} further uses an advantage estimator tailored to the bi-level structure: since rewards occur only at $\bar{t}(t, 0)$, it computes environment-discounted returns across $t$ and applies an additional denoising discount across  $k$ to downweight earlier (noisier) denoising steps.

\subsection{Diffusion Steering Reinforcement Learning (\DSRL{}, \citet{wagenmaker2025steering}}
\label{app:baseline:dsrl}

\DSRL{} improves a pretrained diffusion (or flow) policy without updating its weights by learning a policy over the \emph{input noise space} while keeping the denoising dynamics fixed. Whereas a base diffusion policy $\pi_{\mathrm{dp}}$ samples an initial latent $w_t$ from a fixed prior (typically $\mathcal{N}(0, \eye)$) to maps it to an executed action $a_{t,0}$ via a deterministic denoising chain (e.g., DDIM), \DSRL{} instead formulates a \emph{latent-action MDP} in which the fixed prior is replaced by a learnable latent policy $\pi^{\mathcal W}_\psi(w_t\mid s_t)$. This policy selects specific noise vectors to steer the frozen denoising process toward actions with higher expected return.

Formally, let $\pi_{\mathrm{dp}}(s_t,w_t)$ denote the action produced by running the (frozen) denoising procedure of $\pi_{\mathrm{dp}}$ initialized at $w_t$, i.e., $a_{t,0}=\pi_{\mathrm{dp}}(s_t,w_t)$. Note that if the denoising sampler is stochastic, interpret $\pi_{\mathrm{dp}}$ as inducing a conditional distribution over $a_{t,0}$ given $(s_t,w_t)$.
This induces a latent-action transition kernel
\begin{align*}
P^{\mathcal W}(s_{t+1}\mid s_t,w_t)
:= P\bigl(s_{t+1}\mid s_t, \pi_{\mathrm{dp}}(s_t,w_t)\bigr),
\end{align*}
and \DSRL{} optimizes the latent policy by maximizing the discounted return in this latent-action MDP:
\begin{align*}
\max_{\psi} J(\psi)
:= \mathbb{E}\left[\sum_{t\ge 0}\gamma^t R\bigl(s_t, \pi_{\mathrm{dp}}(s_t,w_t)\bigr)\right],
\qquad w_t \sim \pi^{\mathcal W}_\psi(\cdot\mid s_t).
\end{align*}
In practice, $\pi^{\mathcal W}_\psi$ is learned with a standard off-policy actor--critic algorithm (e.g., SAC) using transitions $(s_t,w_t,r_t,s_{t+1})$ collected by executing $a_{t,0}=\pi_{\mathrm{dp}}(s_t,w_t)$ in the environment.



\textbf{Optimized Variant.} Our $\DSRLplus$ variant applies best-of-N filtering over steering policy actions using the Q-functions and adds a BC-loss using the success buffer to the steering policy on top of the policy graident loss. 

\subsection{Expressive Policy Optimization  (\EXPO{}, \citet{dong2025expo}}
\label{app:baseline:expo}

\EXPO{} is designed to stably fine-tune \emph{expressive} pocilices (e.g., diffusion/flow policies) with online RL by avoiding direct value maximization through the expressive policy parameters. Instead, \EXPO{} maintains (i) a \emph{base} expressive policy $\pi_{\mathrm{base}}$ trained with a stable imitation (suprevised) objective, and (ii) a lightweight Gaussian \emph{edit} policy $\pi_{\mathrm{edit}}$ that performs local action refinement toward higher $Q$-values. At interaction time, \EXPO{} constructs an \emph{on-the-fly} (OTF) policy that samples candidate actions from $\pi_{\mathrm{base}}$, refines them with $\pi_{\mathrm{edit}}$, and executes the candidate with the highest critic value; the same OTF selection is also used inside the TD backup.

Given $a \sim \pi_{\mathrm{base}}(\cdot \mid s)$, \EXPO{} samples an additive edit $\delta \sim \pi_{\mathrm{edit}}(\cdot \mid s, a)$ and forms the refined action $\tilde a = a + \delta$. The OTF policy selects the better of the candidates according to the critic, $a^*(s) \in \argmax_{a'\in\{a,\tilde a\}} Q_{\phi}(s, a').$
The edit policy is updated to increase the value of refined actions (with entropy regularization).
\begin{align*}
    \max_{\pi_{\mathrm{edit}}} \Exp_{(s, a) \sim \mathcal{D}, \delta \sim \pi_{\mathrm{edit}}} \left[Q_{\phi}(s, a + \delta) - \alpha \log \pi_{\mathrm{edit}}(\delta \mid s, a)\right].
\end{align*}
The critic is trained by TD regression using the same OTF selection for the next-state action computed as : $\min_{\phi} \Exp \left[\big(r+\gamma Q_{\phi'}(s',a^{*}(s'))-Q_{\phi}(s,a_t)\big)^2\right]$.
Finally, $\pi_{\text{base}}$ is updated only through imitation-style regression (not direct $Q$-maximization), with value improvement coming from $\pi_{\text{edit}}$ and the OTF selection.

\textbf{Improve Variant.} \EXPOplus{} modifies the behavior cloning term in the standard \EXPO{} for the ``success buffer'' variant described in \Cref{sec:ogpoplus}.

\subsection{Q-Chunking (\QC{}, \citet{li2025reinforcement})}
\label{app:baseline:qc}
Recall that, in our notation, we use a single action $a_t$ to decode an entire action-chunk in a the true environment, $a_{t:t+h-1}$. The \QC{} algorithm proposes multiple variants. One of which, when specialized to GCPs, would require backpropagation through denoising steps, which we show leads to poor performance in \Cref{fig:bptt_vs_ogpo_fig}. Therefore, we opt for the other variant, which amounts to simply best-of-$N$ inference plus behavior cloning.  This variant of \QC{} consists of three simple components:
\begin{itemize}
    \item Learn a critic $Q(s,a)$, following the action-chunking conventions in \Cref{app:action_chunk}. Use this to train the critic via \Cref{eq:Lcrtic}.
    \item Compute the Best-of-$N$ action,  by $\Qtarg$, as following \Cref{eq:a_plan}.
    \item Finally, we use a behavior cloning loss applied to past $(s,a)$ pairs collected by the above planning mechanism,. 
\end{itemize}

\textbf{Optimized Variant.} Our $\QCplus$ variant only applies BC loss to successful actions.

\subsubsection{Q-Chunking v/s OGPO}
\label{app:qc_vs_ogpo}
Q-Chunking learns Q-functions that evaluate entire action chunks as atomic units, treating $Q(s, a_{1:H})$, where $a_{1:H}$ denotes the full action sequence over a horizon $H$. This formulation is agnostic to how the action chunk is generated—whether via a flow policy, a diffusion model, or direct regression. Policy improvement is guided using the Q-functions to rank a batch of actions and perform supervised fine-tuning (SFT) using BC loss on the Best-of-N actions. In contrast, \OGPO{} explicitly leverages the iterative structure of the Generative Control Policy (GCP) by computing annealed importance sampling ratios over the denoising chain \Cref{eq:ppo_internal}. Moreover, the advantage computation evaluates the group relative Q values over the entire action chunk and the policy gradient loss propagates through \emph{every} denoising step $k$. This end-to-end formulation ensures that producing a high-value action requires coherent refinement at every GCP step, rather than treating the generation process as a black box.

\subsection{ReinFlow (\citet{zhang2025reinflow}, not compared)}
\label{app:baseline:reinf}

The ReinFlow algorithm \citep{zhang2025reinflow} is nearly identical to \DPPO{}, except that it uses a flow policy as a base policy instead of Diffusion. To get non-singular likelihood rations, it augments the flow model with additional noise. However, their reported numbers are less sample efficient than \DPPO{} (the flow sampling, however, improves \emph{computational} efficiency), so we only use \DPPO{} as a stronger baseline.

\subsection{PA-RL (\citet{mark2024policy}, not compared)}
\label{app:baseline:parl}

The PA-RL \cite{mark2024policy} algorithm is similar to QC, but includes an additional gradient ascent step $a' \gets \nabla_a Q(s,a)$ to further improve actions. These gradient computations present a significant computational overhead, and perform best on TPU hardware. We found this method infeasible to run given our compute budget. Furthermore, given the instability of Q-gradients in non-smooth tasks \citep{suh2022differentiable}, we conjecture this method would struggle in the contact-rich \textsc{RoboMimic} tasks. 

\section{Understanding Exploration Behavior of \OGPO{}}
\label{app:exploration}

This section elaborates on the exploration dynamics of \OGPO{} discussed in Section \cref{sec:explore}. We provide visualizations that clarify how \OGPO{} expands the action manifold of pretrained policy distributions while maintaining stable policy improvement.

\paragraph{Sample Efficiency vs. Execution Efficiency}
In the training dynamics of \OGPO{}, we observe two colliding optimization objectives: (1) \textbf{Sample Efficiency}: Minimizing the number of environment interactions required for policy convergence, and (2) \textbf{Execution Efficiency}: Minimizing the number of timesteps the policy takes to complete a task during inference. \OGPO{} excels at the former via off-policy stitching, but the latter introduces unique instabilities. The discount factor $\gamma < 1$ in the Bellman equation $Q(s,a)=r+\gamma\Qtarg(s', a')$ creates a contraction map that conditions the policy to solve tasks as quickly as possible to maximize the expected return-to-go. This causes the GCP to generate actions that could potentially maximize the speed of achieving the goal, but do not necessarily abide by physical constraints like gravity, acceleration, and robot joint position and velocity limits. This explains the oscillations in the success rate during RL-finetuning induced by rapid policy convergence via Q functions. 

\begin{figure*}[h]
    \centering
    \includegraphics[width=0.75\linewidth]{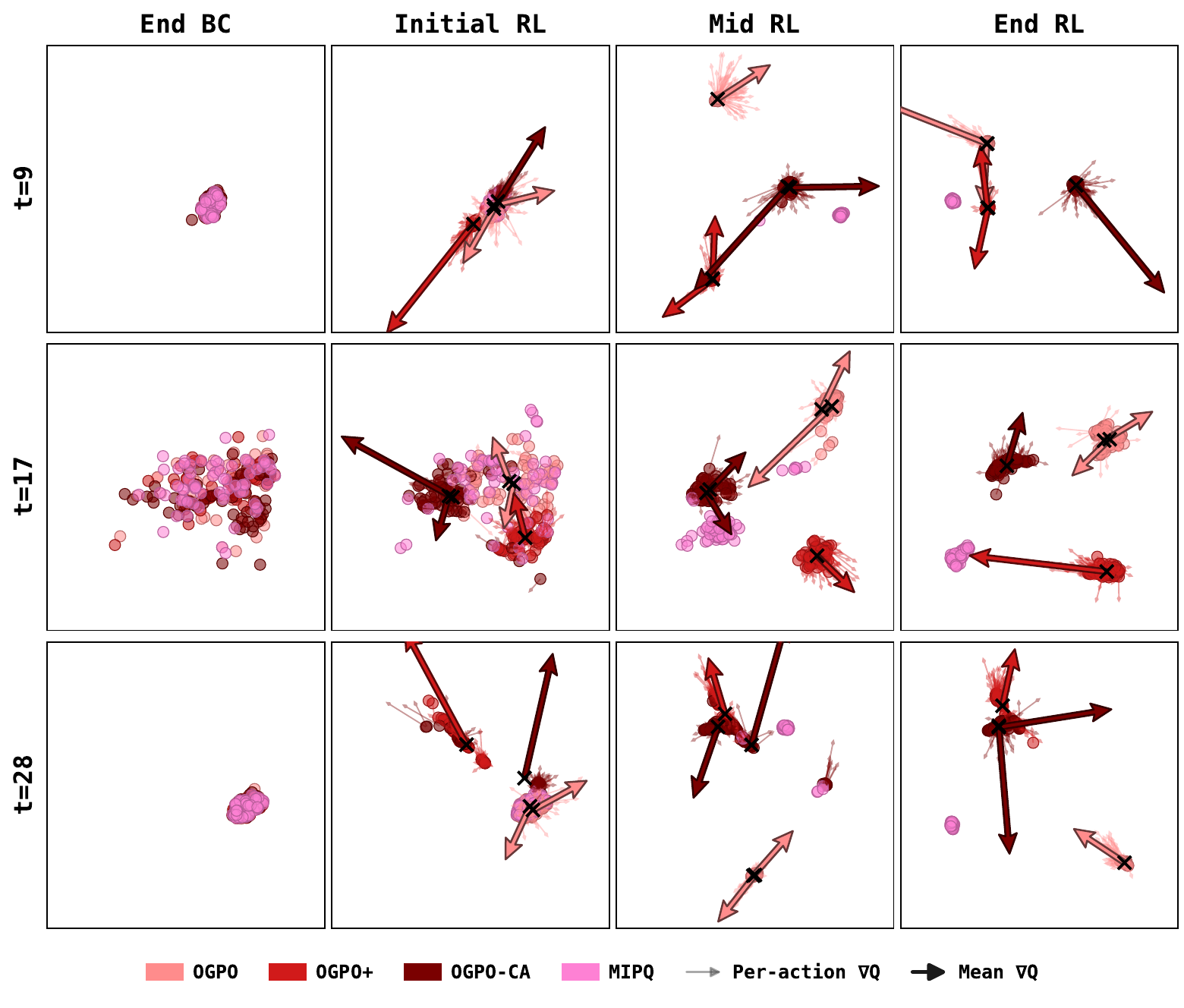}
    \caption{UMAP plot of \OGPO{}, \OGPOplus{}, and \OGPOplusca{} on \robomimic{} \rmtoolhang{}}
    \label{fig:umap_ogpo}
\end{figure*}

\begin{figure*}[h]
    \centering
    \includegraphics[width=0.75\linewidth]{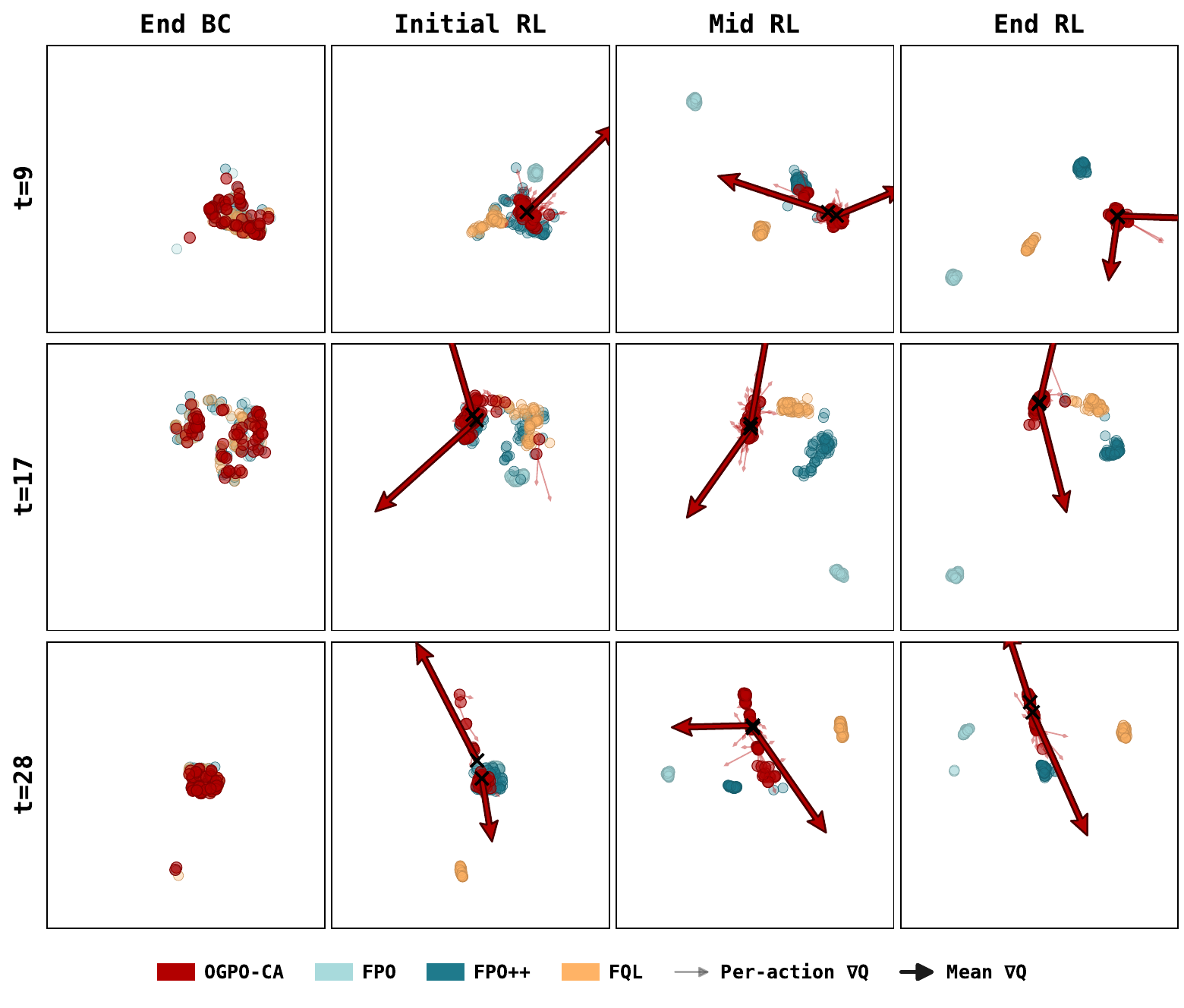}
    \caption{UMAP plot of \OGPO{} comparison with various policy extraction methods on \robomimic{} \rmtoolhang{}}
    \label{fig:umap_policy_extraction}
\end{figure*}


\iftoggle{arxiv}{}
{
\begin{figure}[h]
    \centering
    \includegraphics[width=0.8\linewidth]{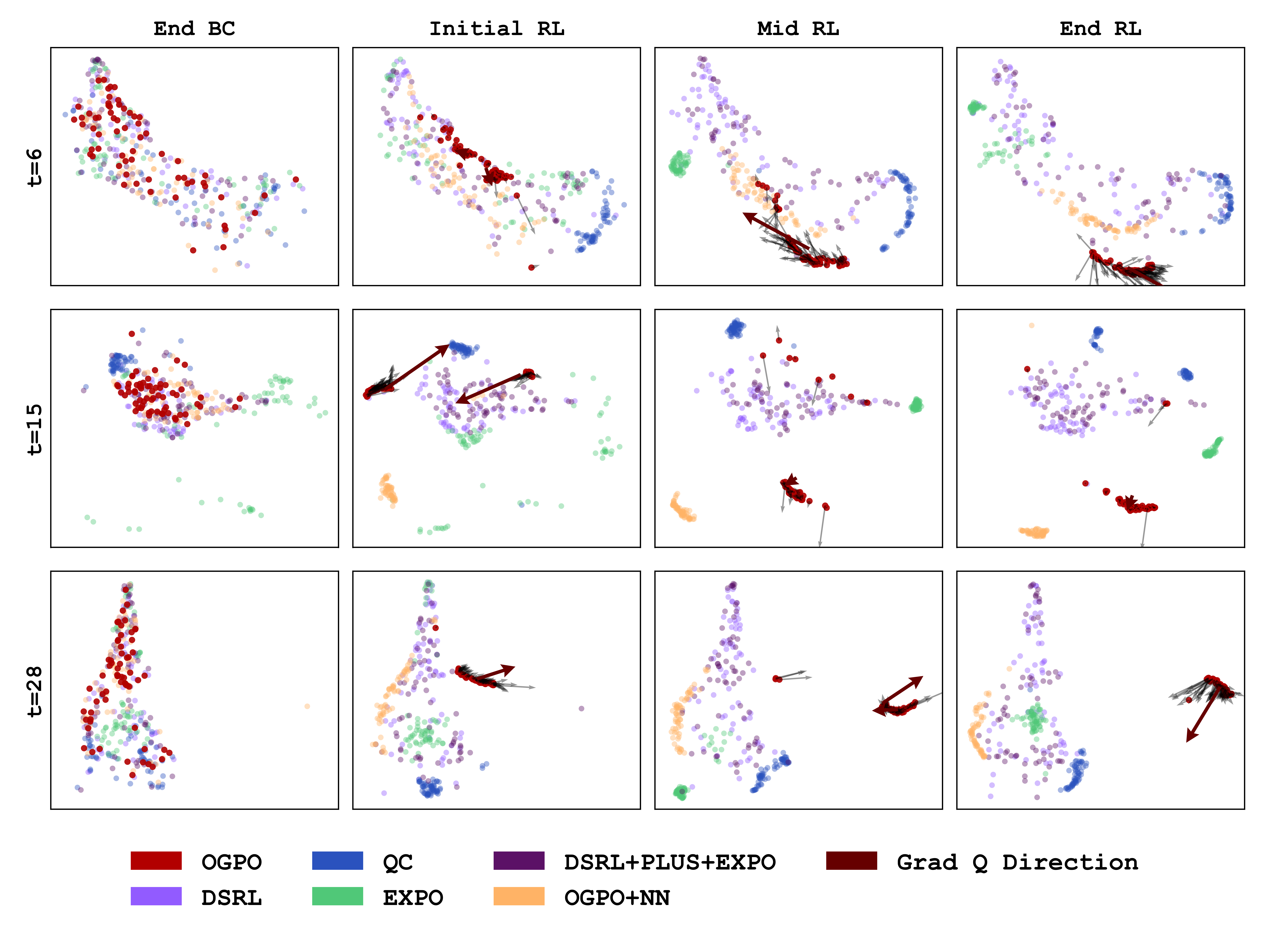}
    \caption{We consider 3 critical states (rows) in $\rmtoolhang$ shown in \cref{fig:critical_states}, and generate a UMAP plot of 64 action samples from \OGPO{} and all the baselines combined and plotted separately at the end of BC pretraining, Initial-stage RL checkpoint Mid-stage RL checkpoint, and End-stage RL checkpoint for each critical state. We further show $\frac{d\mathrm{UMAP}}{da}\nabla_aQ(s,a)$ vectors at each \OGPO{} action to demonstrate the orthogonal conditioning of $\nabla_aQ(s,a)$ to the action spread at critical states.}
    \label{fig:critical_states_umap}
\end{figure}

\begin{figure}[h]
    \centering
    \includegraphics[width=0.8\linewidth]{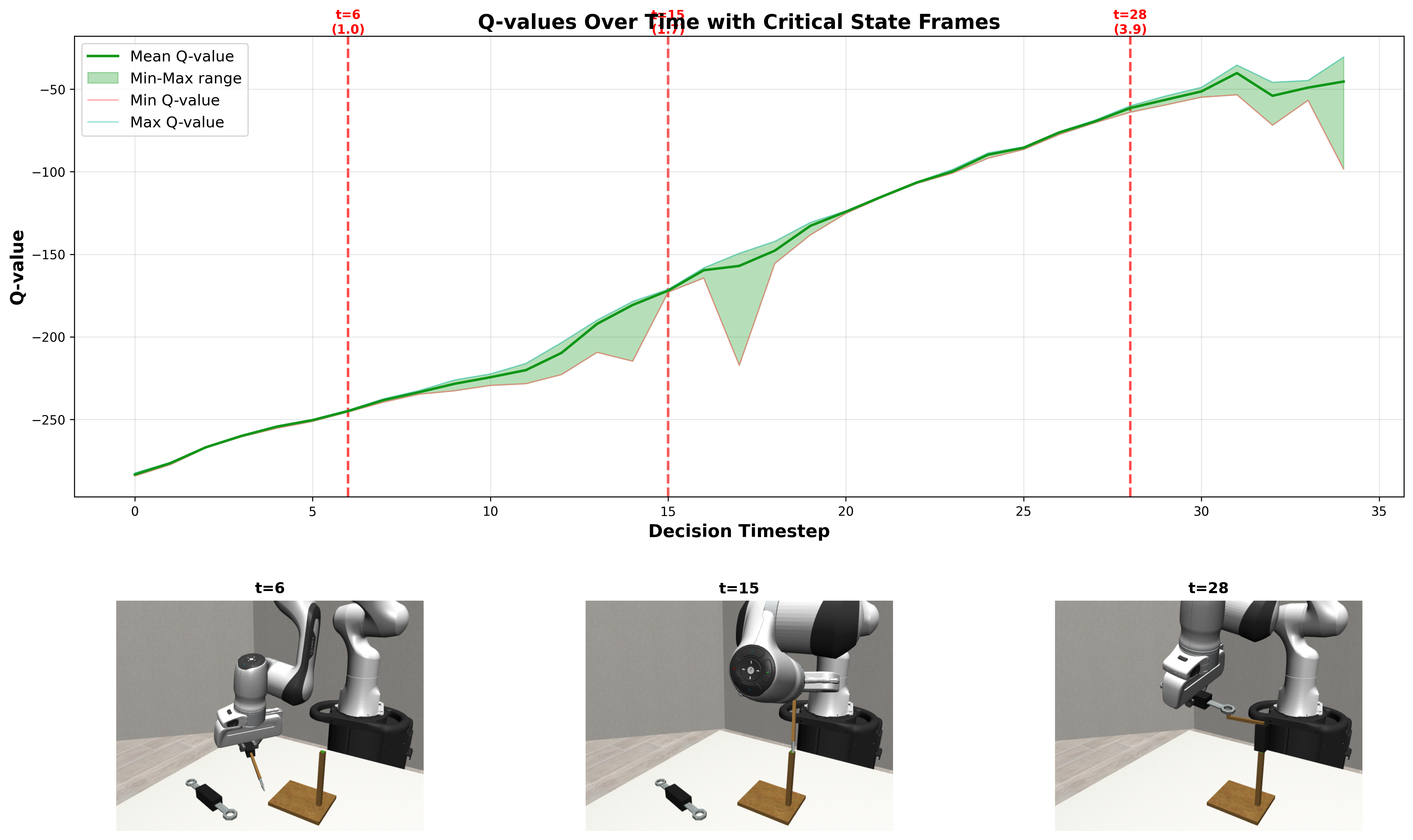}
    \caption{We rollout a successful \OGPOplus{} policy and obtain critical state frames in $\rmtoolhang$ and visualize the corresponding variance in $Q*$-values during the rollout}
    \label{fig:critical_states}
\end{figure}
}

\section{Ablations and Limitations of \OGPO/\OGPOplus}
\label{app:ablations}

\subsection{\BPTT{} vs \OGPO{}}
\label{app:bptt_vs_ogpo}
The most direct way to train off-policy RL policies is to perform gradient ascent on the Q-values. Although this works for simpler policy parameterizations like Gaussian \citep{fujimoto2018addressing}, or Squashed Gaussian \citep{haarnoja2018soft} policies, directly using Q values to sequentially backpropagate through the GCP (also referred to as \emph{Back Propagation Through Time (\BPTT{})}) can be unstable \citep{bengio1994bptt}. \OGPO{} modifies the off-policy learning paradigm for a general class of GCPs by (1) retaining the TD error loss for Q function updates, and (2) using Q functions as substitutes for Monte Carlo rollouts and computing relative advantages $\advgrpo$ for PPO-style updates over the entire GCP chain for the policy updates.



\subsection{\OGPO{} v/s \OGPOplus{}, with and without GRPO std ($\sigma$)}
\label{sec:ogpo_ogpoplus_ablations}
GRPO formulation uses group relative advantage computation similar to \OGPO{}. However, the GRPO advantage uses the standard deviation of the critic ensembles to normalize the advantage values. We found this to be empirically detrimental to \OGPO{}'s success. We attribute this pattern to the sensitivity of the Annealed Importance Sampling ratio $\omega$ to very large and very small advantage values. We leave an extensive empirical validation of this sensitivity as future work. 

\subsection{\OGPO{} vs Steering + Residual Ablation}
\begin{figure}[H]
    \vspace{-0.5em}
    \centering
    \includegraphics[width=0.7\linewidth]{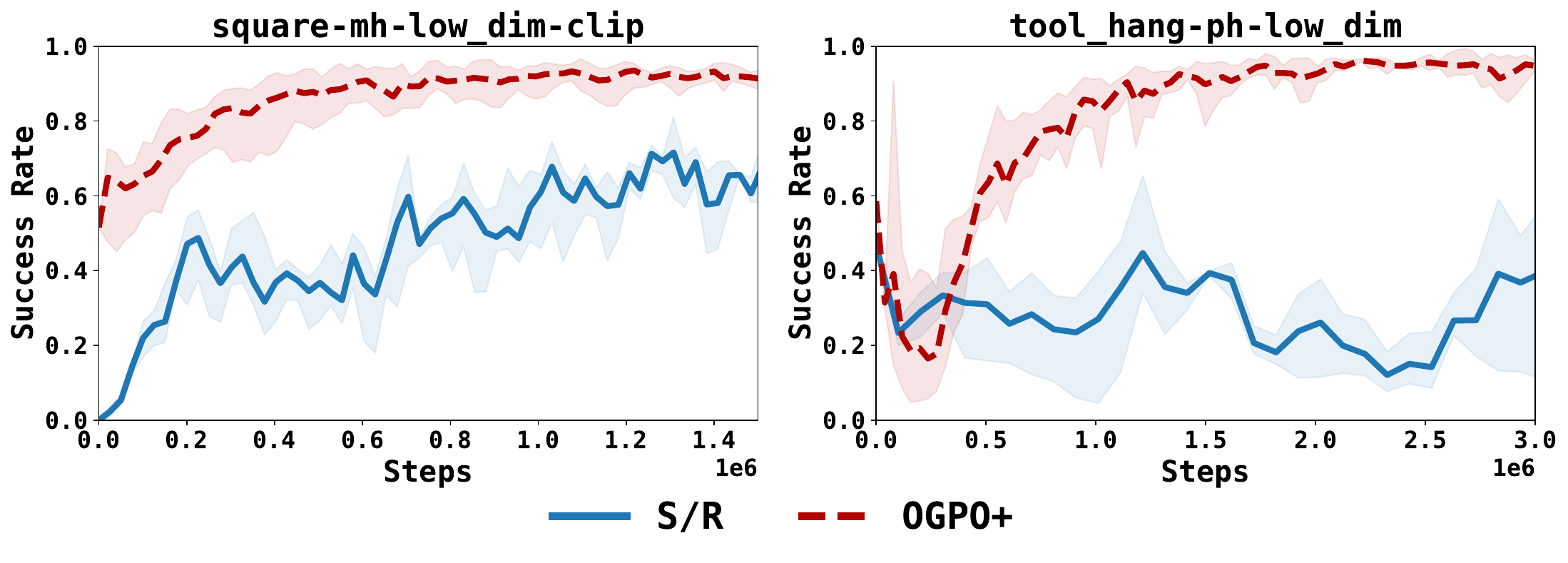}
    \caption{\OGPOplus{} comparison with an ablation of simultaneous steering and residual learning baseline: \SR{}}
    \label{fig:dsrl_plus_expo}
    \vspace{-0.8em}
\end{figure}


\subsection{Policy Extraction Alternatives (AWR, ASPO from FPO)}
\label{app:pol_extr}

\OGPO{} separates critic learning from policy extraction: after learning $Q_\phi$ with the TD objective, the actor update only needs a mechanism for increasing the probability of high-advantage actions and decreasing the probability of low-advantage actions. This makes it natural to ask whether the PPO-style extraction in \OGPO{} is necessary, or whether simpler weighted-regression or flow-matching objectives suffice. To isolate the effect of the extraction objective, all variants below use the same replay buffer, critic update, EMA policy $\piema$, and group-relative advantages $\advgrpo$ as in \Cref{eq:ppo_internal}; only the actor loss is changed.

\subsubsection{Advantage-Weighted Regression and Advantage-Weighted \OGPO{}}
\label{sec:AWR}

AWR-style extraction replaces the clipped PPO ratio with weighted flow-matching regression toward samples from the reference policy. For a sampled final action $a_0 \sim \piema(\cdot \mid s)$, we define
\begin{equation}
    w_{\mathrm{AWR}}(s,a_0)
    =
    \exp\!\left(\frac{\advgrpo(s,a_0)}{\beta}\right),
    \label{eq:awr_weight}
\end{equation}
where $\beta$ controls how sharply the update concentrates on high-advantage samples. In our flow-policy implementation, the actor loss is not a discrete denoising-chain log-likelihood; it is a weighted conditional flow-matching objective:
\begin{equation}
    \L_{\mathrm{AWR}}(\theta)
    =
    \mathbb{E}_{s \sim \cB,\, a_0 \sim \piema(\cdot \mid s)}
    \left[
    \mathrm{sg}\!\left(w_{\mathrm{AWR}}(s,a_0)\right)
    \cdot
    \mathbb{E}_{\tau,\,\xi}
    \left\|
    v_\theta(x_\tau,\tau,s) - (a_0-\xi)
    \right\|^2
    \right],
    \label{eq:awr_loss}
\end{equation}
where $\xi \sim \Normal(0,I)$, $\tau \sim \mathrm{Unif}(0,1)$, and
$x_\tau = \tau a_0 + (1-\tau)\xi$. We also evaluate advantage-weighted \OGPO{} (\AWOGPO{}), which preserves the same group-relative advantage computation but replaces the clipped PPO surrogate with this advantage-weighted CFM update. Empirically, these objectives are brittle on high-precision and long-horizon tasks: they can imitate high-$Q$ samples, but do not reliably suppress bad modes when the critic is imperfect.

\subsubsection{ASPO from Flow Policy Optimization}
\label{sec:fpo_aspo}

We also compare against the asymmetric trust-region objective used in Flow Policy Optimization (FPO) \citep{yi2026flow}. Instead of computing the exact denoising likelihood ratio used by \OGPO{}, FPO constructs a surrogate ratio from the conditional-flow-matching loss:
\begin{equation}
    \ratiofpo
    :=
    \exp\!\left(
    \LCFM(\bar{\theta}; s,a)
    -
    \LCFM(\theta; s,a)
    \right),
    \label{eq:fpo_ratio}
\end{equation}
where $\bar{\theta}$ denotes the EMA/reference policy. ASPO then applies different updates depending on the sign of the advantage. For positive advantages, it uses a PPO-style clipped objective that increases the likelihood of good actions. For negative advantages, it uses an SPO penalty with a dead zone inside the trust region:
\begin{equation}
    \psi_{\mathrm{ASPO}}(\ratiofpo,\advgrpo)
    =
    \begin{cases}
        \min\!\left(
        \ratiofpo \advgrpo,\;
        \mathrm{clip}(\ratiofpo,1-\epsilon,1+\epsilon)\advgrpo
        \right),
        & \advgrpo \geq 0, \\[6pt]
        \ratiofpo \advgrpo
        -
        \dfrac{|\advgrpo|}{4\epsilon}
        \left(
        \max\!\left(0,\; |\ratiofpo-1|-\epsilon\right)
        \right)^2,
        & \advgrpo < 0.
    \end{cases}
    \label{eq:aspo}
\end{equation}
Thus, negative-advantage samples receive no additional SPO penalty while $\ratiofpo \in [1-\epsilon,1+\epsilon]$; the penalty only turns on once the update moves outside the trust-region boundary, and then grows quadratically in the excess violation. Compared to \OGPO{}, FPO avoids explicitly evaluating the full denoising-chain likelihood ratio, but this surrogate also weakens the connection between the extraction objective and the actual stochastic denoising process used during rollout.

Compared to \OGPO{}, FPO has the appealing property that it can be implemented directly through the CFM loss, without explicitly evaluating the full denoising-chain likelihood ratio. However, this surrogate also weakens the connection between the extraction objective and the actual stochastic denoising process used during policy rollout. In our experiments, we find the FPO++'s Asymmetric trust region (ASPO) updates to be more competitive than FPO and hence we call the off-policy version of this line of work as \FPO{}. Although \FPO{} converges more generally than pure AWR, we find that it remains less stable than PPO-style extraction, especially on tasks where critic errors and low-value modes must be suppressed early in online learning.

\subsection{\OGPO{} with Flow vs. Diffusion Instantiation}


While we have presented \OGPO{} in the context of \emph{flow-matching} policies, the algorithm is agnostic to the specific generative parameterization of the GCP and applies directly to diffusion policies as well. Both flow-matching and score-based diffusion policies define an iterative denoising chain $\aiter[K] \to \aiter[K-1] \to \cdots \to \aiter[0]$ from a base noise distribution to the action distribution; the only difference is the parameterization of the per-step transition (a learned velocity field $v_\theta$ for flow policies versus a learned score / $\epsilon$-prediction for diffusion). \OGPO{} depends only on generic properties of the underlying SDE and therefore carry over unchanged to a diffusion-policy GCP, modulo the appropriate noise schedule and score parameterization.

We verify this empirically in \Cref{fig:ogpo_diffusion}, where we instantiate \OGPO{} on top of a diffusion-policy backbone and observe consistent improvement over BC pretraining, mirroring the trends reported for flow-policy backbones throughout the main paper. In practice, however, we default to flow-matching policies for our main experiments: flow policies require substantially fewer denoising steps at inference time (typically $K=4$--$10$ versus $K=50$--$100$ for diffusion) while achieving comparable BC performance, which directly translates to faster environment rollouts and meaningfully reduced wall-clock cost for online RL. We therefore view diffusion-policy \OGPO{} as a drop-in alternative whenever the underlying VLA backbone is itself a diffusion model, and flow-policy \OGPO{} as the preferred default when inference compute is a bottleneck.

\RETURNN{MiQ}




\section{Environment Details}
\label{app:envirs}
\subsection{\fk}
The \fk benchmark \citep{gupta2019relay} tests multi-task sequential manipulation with compositional task structure. The environment features a 9-DoF Franka robot that must manipulate 4 kitchen objects (microwave, kettle, light switch, slide cabinet) to desired goal configurations in a specific sequence. This environment is particularly challenging due to its requirement for long-horizon planning and the need to compose multiple subtasks correctly.

\textbf{State and Action Spaces:} The state space consists of robot joint positions, joint velocities, and object states (\texttt{state\_dim} = 60). Actions are 9-dimensional continuous controls for the robot joints (\texttt{action\_dim} = 9), normalized to $[-1, 1]$.

\textbf{Task Horizon and Other Parameters:} \fk tasks have a medium horizon of approximately 280 timesteps. We use $\gamma = 0.99$ to account for the medium-length temporal dependencies across subtasks. The action chunk size is set to $h=4$ to provide temporal smoothness while maintaining reactivity.

\textbf{Datasets:} We use three offline datasets from D4RL \citep{fu2020d4rl}:
\begin{itemize}
    \item $\fkcomplete$: Complete demonstrations of all 4 subtasks in the correct sequence
    \item $\fkmixed$: Randomized subtask orders where the desired sequence is not completed sequentially
    \item $\fkpartial$: Partial subtrajectories of the desired task
\end{itemize}

\textbf{Reward Structure:} We use a sparse reward structure with a base reward of -7. Each successful subtask completion adds +1, with the final subtask providing +3 upon success. This yields a maximum reward of 0 for completing all subtasks.

\subsection{Robomimic}
The \textsc{Robomimic} benchmark \citep{robomimic2021} provides high-precision manipulation tasks that test fine-grained control and multi-step reasoning. We evaluate on three of the most challenging tasks that represent different aspects of real-world manipulation:

\textbf{Square ($\rmsquare$):} A medium-horizon fine-grained insertion task requiring precise alignment and insertion of a square peg. This task tests contact-rich manipulation with tight tolerances.
\begin{itemize}
    \item \texttt{state\_dim}: 14 (robot end-effector pose, object pose)
    \item \texttt{action\_dim}: 7 (6D end-effector control + gripper)
    \item Horizon: ~400 timesteps
    \item $\gamma = 0.99$
    \item Action chunk size: $h=4$
    \item Dataset: Multi-Human (MH) mixed proficiency
\end{itemize}

\textbf{Tool Hang ($\rmtoolhang$):} A long-horizon, highly-precise multi-step insertion task requiring the robot to grasp a tool and hang it on a rack. This task demands both coarse positioning and fine-grained alignment across multiple phases.
\begin{itemize}
    \item \texttt{state\_dim}: 14
    \item \texttt{action\_dim}: 7
    \item Horizon: 1000 timesteps
    \item $\gamma = 0.999$ (higher due to longer horizon)
    \item Action chunk size: $h=8$ (larger chunks for smoother long-horizon execution)
    \item Dataset: Proficient-Human (PH), BC stopped at 50\% success rate
\end{itemize}

\textbf{Transport ($\rmtransport$):} A bi-manual, multi-step, long-horizon object transfer task where two robot arms must coordinate to transport an object. This tests both individual arm control and bi-manual coordination.
\begin{itemize}
    \item \texttt{\texttt{state\_dim}}: 28 (dual arm configuration)
    \item \texttt{action\_dim}: 14 (7 per arm)
    \item Horizon: 800 timesteps
    \item $\gamma = 0.999$ (higher due to longer horizon)
    \item Action chunk size: $h=8$
    \item Dataset: Multi-Human (MH) mixed proficiency
\end{itemize}

\textbf{Reward Structure:} All Robomimic tasks use sparse rewards: -1 for each non-successful step, with the final successful step returning 0.

\textbf{Note on Hyperparameters:} The different gamma values reflect the relationship between discount factor and task horizon. Longer horizon tasks ($\rmtoolhang$, $\rmtransport$) require larger gamma (0.999) to properly credit distant actions, while medium-horizon tasks ($\rmsquare$) use smaller gamma (0.99). Similarly, longer tasks benefit from larger action chunks ($h=8$) for smoother execution. Importantly, both gamma and chunk size are independent of action dimensionality.

\subsection{Adroit Hand}
The Adroit Hand benchmark tests dexterous manipulation with a 24-DoF anthropomorphic robotic hand performing high-precision, contact-rich tasks. This environment is particularly challenging due to the high-dimensional action space, under-actuated dynamics, and the need for coordinated finger movements.

We evaluate on four standard tasks:
\begin{itemize}
    \item \texttt{AdroitHandDoor-v1}: Door opening requiring articulated finger coordination to grasp and turn a handle
    \item \texttt{AdroitHandHammer-v1}: Hammering a nail with precise force control and wrist articulation
    \item \texttt{AdroitHandPen-v1}: In-hand pen reorientation requiring complex finger gaiting
    \item \texttt{AdroitHandRelocate-v1}: Object relocation requiring coordinated grasping and translation
\end{itemize}

\textbf{State and Action Spaces:} 
\begin{itemize}
    \item \texttt{state\_dim}: 45 (24 joint positions + 24 joint velocities + object state)
    \item \texttt{action\_dim}: 24 (continuous control for each DoF)
    \item Actions normalized to $[-1, 1]$
\end{itemize}

\textbf{Task Horizon and Temporal Parameters:}
\begin{itemize}
    \item Horizon: ~200 timesteps (medium-horizon tasks)
    \item $\gamma = 0.95$
    \item Action chunk size: $h=4$ for stabilized policy execution
\end{itemize}

\textbf{Datasets:} We use expert demonstration datasets provided via the D4RL/Minari interface for pretraining the base policy.

\textbf{Evaluation:} Following prior work, we evaluate performance using the normalized return provided by the environment, scaled to $[0,100]$.

\subsection{LIBERO}
The LIBERO benchmark \citep{liu2023libero} tests vision-based, language-conditioned manipulation for multi-task learning and generalization. Unlike the previous environments, which use state-based observations, LIBERO provides pixel observations and requires following natural-language instructions, thereby testing both visual understanding and instruction-following capabilities.

\begin{figure}[h]
    \vspace{-0.5em}
    \centering
    \includegraphics[width=0.7\linewidth]{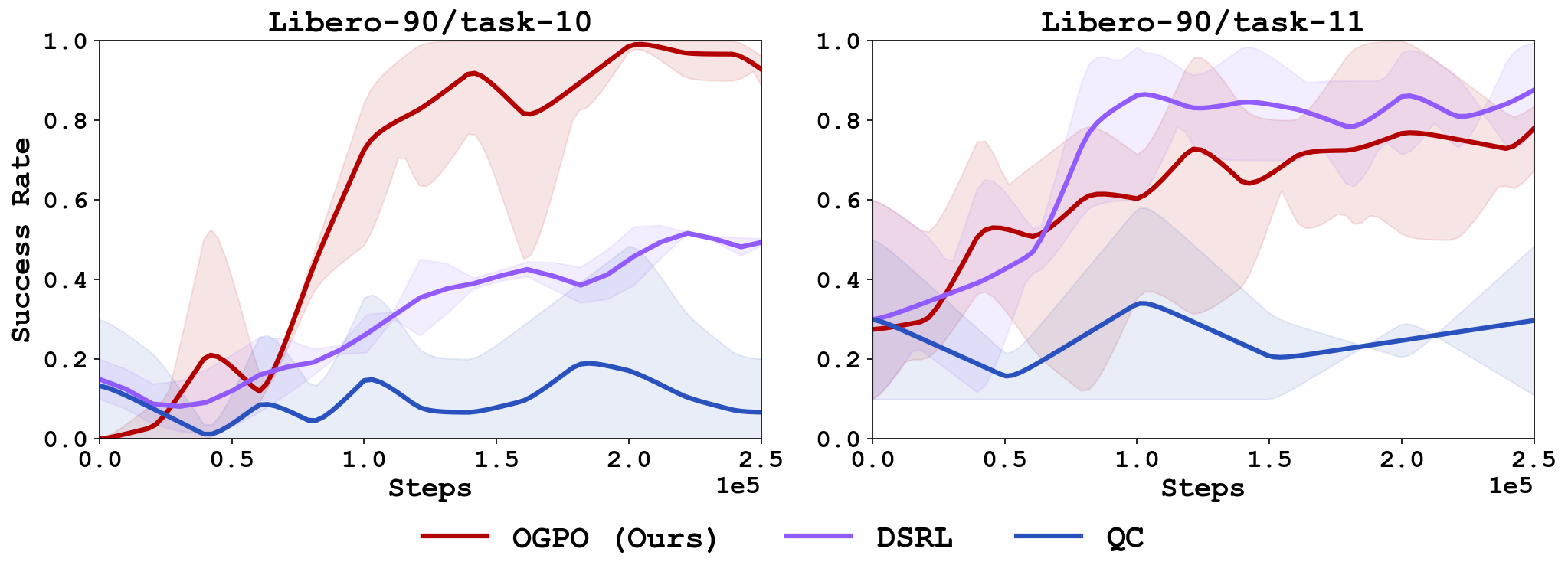}
    \caption{We compare \OGPO{} with \DSRL{} and \QC{} on pixel-based observations and natural language guidance tasks from the LIBERO benchmark}
    \label{fig:libero_comparisons}
\end{figure}

\textbf{Observation and Action Spaces:}
\begin{itemize}
    \item Observations: RGB images (128×128×3 pixels)
    \item \texttt{action\_dim}: 7 (6D end-effector control + gripper)
    \item Actions normalized to $[-1, 1]$
\end{itemize}

\textbf{Task Structure:} LIBERO features procedurally generated tasks with natural language instructions. Tasks require understanding spatial relationships and object attributes from both visual and linguistic modalities.

\textbf{Reward Structure:} All Libero tasks use sparse rewards: -1 for each non-successful step, with the final successful step returning 0.

\textbf{Task Horizon and Temporal Parameters:}
\begin{itemize}
    \item Horizon: ~1000 timesteps (long-horizon tasks)
    \item $\gamma = 0.999$ (for \OGPO{}, \DSRL), 0.99 (for \QC{} since we found this leads to better performance)
    \item Action chunk size: $h=8$
\end{itemize}

\textbf{Training and Evaluation Setup:} The base policy is trained on demonstrations from 10 tasks ($\texttt{task\_id} \in \{0,\ldots,9\}$) in the Libero-90 dataset and evaluated on 2 unseen downstream tasks ($\texttt{task\_id} \in {10, 11}$) to test generalization capabilities. This setup explicitly tests the ability to transfer learned manipulation skills to novel task descriptions and object configurations. Since LIBERO is a language-conditioned benchmark, for both the actor and critic, we follow a widely used design from prior work~\citep{walke2023bridgedata, nakamoto2024steering}: language instructions are first processed by a frozen MUSE encoder~\citep{yang2019multilingualuniversalsentenceencoder} and then passed to an IMPALA encoder~\citep{espeholt2018impalascalabledistributeddeeprl} with FiLM conditioning~\citep{perez2018film}.
\section{Hyper-parameters and Initialization}
\label{app:hyperparameters}
\subsection{Initialization and Warm Starting}
\label{sec:initialization}
\OGPO{} accommodates two primary settings based on data availability, each with corresponding algorithmic choices for initialization.
\colorbold{Setting 1: Offline data available.} When an offline dataset $\rboff$ is available, we pre-train our policy $\piigpbc$ on $\rboff$ using the appropriate BC loss.  The $\texttt{use\_offline}$ flag is toggled $\texttt{True}$, enabling offline data sampling reuse determined by the ratio $\offlineratio$. 

\colorbold{Setting 2: No offline data (online-only).} We finetune a pre-trained IGP with \emph{no} additional demonstration data which has some small but non-trivial base success rate (>10\%). The $\texttt{use\_offline}$ flag is toggled $\texttt{False}$.

In both settings, the online replay buffer $\rb$ is initialized with $\nwarmup$ $\piigp$ rollouts, where $\piigp \gets \piigpbc$. Finally, we initialize an ensemble of Q-functions $Q_{\phi_{1,\dots,M}}$ with random weights and, importantly, find that no offline RL pretraining yields the highest sample efficiency. We defer the details of the offline RL ablations to \cref{alg:initialize}.

\subsection{Hyperparameters}
In this section, we list all the hyper parameters we use for $\OGPO$ across different benchmarks.
Table \ref{tab:env-max-episode-length} shows the maximum episode lengths we use for each environment.

\begin{table}[h]
\centering
\small
\begin{tabular}{l r}
\hline
\textbf{Environment} & \textbf{Max Episode Length} \\
\hline
square & 400 \\
transport & 800 \\
tool\_hang & 1000 \\
kitchen (all) & 600 \\
adroit (all) & 200 \\
\hline
\end{tabular}
\caption{Environment maximum episode lengths}
\label{tab:env-max-episode-length}
\end{table}

We first list the common $\OGPO$ hyper parameters. Unless otherwise stated, these remain constant throughout all our experiments. These are in Table \ref{tab:ogpo-agent-defaults}.

\begin{table}[H]
\centering
\small
\begin{tabular}{l l}
\hline
\textbf{Parameter} & \textbf{Default Value} \\
\hline
lr & $3\mathrm{e}{-4}$ \\
actor\_lr & $3\mathrm{e}{-4}$ \\
critic\_lr & $3\mathrm{e}{-4}$ \\
ppo\_lr & $4.5\mathrm{e}{-5}$ \\
tau & 0.05 \\
actor\_tau & 0.05 \\
discount & 0.99 \\
batch\_size & 256 \\
ppo\_batch\_size & 256 \\
actor\_hidden\_dims & (512, 512, 512, 512) \\
value\_hidden\_dims & (512, 512, 512, 512) \\
num\_qs & 10 \\
q\_agg & mean \\
subsample\_bon & True \\
flow\_steps & 10 \\
grpo\_num\_samples & 32 \\
clip\_epsilon & 0.01 \\
entropy\_coeff & 0.0 \\
bc\_coeff & 1.0 \\
constant\_noise\_std & 0.01 \\
actor\_scheduler & cosine \\
critic\_scheduler & constant \\
actor\_warmup\_steps & 2000 \\
actor\_decay\_steps & 50000 \\
actor\_end\_value & $2\mathrm{e}{-5}$ \\
critic\_warmup\_steps & 500 \\
critic\_decay\_steps & 5000 \\
critic\_end\_value & 0.0 \\
actor\_weight\_decay & 0.0 \\
critic\_weight\_decay & $1\mathrm{e}{-5}$ \\
horizon\_length & 4 \\
policy\_type & flow \\
\hline
\end{tabular}
\caption{OGPO agent default hyperparameters.}
\label{tab:ogpo-agent-defaults}
\end{table}

In Table \ref{tab:ogpo-robomimic}, we list down all $\robomimic$ specific hyper-parameters that are used for our experiments.

\begin{table}[H]
\centering
\small
\setlength{\tabcolsep}{6pt}
\begin{tabular}{l c c c}
\hline
\textbf{Hyperparameter} & $\rmsquare$ & $\rmtoolhang$  & $\rmtransport$ \\
\hline
\multicolumn{4}{l}{\textbf{Training Steps}} \\
offline\_steps & 500{,}000 & 500{,}000 & 1{,}000{,}000 \\
online\_steps & 2{,}000{,}000 & 3{,}000{,}000 & 6{,}000{,}000 \\
start\_training & 20{,}000 & 25{,}000 & 40{,}000 \\
\hline
\multicolumn{4}{l}{\textbf{RL Hyperparameters}} \\
horizon\_length & 4 & 8 & 8 \\
discount & 0.99 & 0.999 & 0.999 \\
tau & 0.05 & 0.05 & 0.05 \\
utd\_warmup & 1 & 1 & 1 \\
utd\_online & 1 & 1 & 1 \\
\hline
\multicolumn{4}{l}{\textbf{Q-Network}} \\
num\_qs & 10 & 10 & 10 \\
q\_agg & mean & mean & mean \\
subsample\_bon & True & True & True \\
best\_of\_n & 8 & 8 & 8 \\
value\_hidden\_dims & (512,512,512,512) & (512,512,512,512) & (512,512,512,512,512) \\
\hline

\multicolumn{4}{l}{\textbf{BC Regularization}} \\
use\_bc\_regularization & True & True & True \\
bc\_coeff & 1.0 & 1.0 & 1.0 \\
pg\_coeff & 1.0 & 1.0 & 1.0 \\
clip\_bc (atmost 50\% success rate) & True & True & False \\
\hline
\end{tabular}
\caption{OGPO hyperparameters for Robomimic environments.}
\label{tab:ogpo-robomimic}
\end{table}

In Table \ref{tab:ogpo-kitchen}, we list all hyper parameters we use for the various $\fk$ environments.

\begin{table}[H]
\centering
\small
\setlength{\tabcolsep}{6pt}
\begin{tabular}{l c c c}
\hline
\textbf{Hyperparameter} & $\fkcomplete$ &  $\fkmixed$ &  $\fkpartial$ \\
\hline
\multicolumn{4}{l}{\textbf{Training Steps}} \\
offline\_steps & 1{,}000{,}000 & 1{,}000{,}000 & 1{,}000{,}000 \\
online\_steps & 3{,}000{,}000 & 3{,}000{,}000 & 3{,}000{,}000 \\
\hline
\multicolumn{4}{l}{\textbf{RL Hyperparameters}} \\
horizon\_length & 4 & 4 & 4 \\
discount & 0.99 & 0.99 & 0.99 \\
tau & 0.05 & 0.05 & 0.05 \\
utd\_warmup & 1 & 1 & 1 \\
utd\_online & 1 & 1 & 1 \\
\hline
\multicolumn{4}{l}{\textbf{Q-Network}} \\
num\_qs & 10 & 10 & 10 \\
q\_agg & mean & mean & mean \\
subsample\_bon & True & True & True \\
best\_of\_n & 8 & 8 & 8 \\
\hline
\multicolumn{4}{l}{\textbf{BC Regularization}} \\
use\_bc\_regularization & True & True & True \\
bc\_coeff & 0.1 & 0.1 & 0.1 \\
clip\_bc & False & False & False \\
\hline
\end{tabular}
\caption{OGPO hyperparameters for $\fk$}
\label{tab:ogpo-kitchen}
\end{table}

In Table \ref{tab:ogpo-adroit}, we list all hyper parameters we use for the various $\AdroitEnv$ environments.

\begin{table}[H]
\centering
\small
\setlength{\tabcolsep}{6pt}
\begin{tabular}{l c c c c}
\hline
\textbf{Hyperparameter} & \texttt{Door-v1} & \texttt{Pen-v1} & \texttt{Hammer-v1} & \texttt{Relocate-v1} \\
\hline
\multicolumn{5}{l}{\textbf{Training Steps}} \\
offline\_steps & 50{,}000 & 50{,}000 & 50{,}000 & 50{,}000 \\
online\_steps & 500{,}000 & 500{,}000 & 500{,}000 & 500{,}000 \\
\hline
\multicolumn{5}{l}{\textbf{RL Hyperparameters}} \\
horizon\_length & 4 & 4 & 4 & 4 \\
discount & 0.95 & 0.95 & 0.95 & 0.95 \\
tau & 0.05 & 0.05 & 0.05 & 0.05 \\
utd\_warmup & 1 & 1 & 1 & 1 \\
utd\_online & 4 & 4 & 4 & 4 \\
\hline
\multicolumn{5}{l}{\textbf{Q-Network}} \\
num\_qs & 10 & 10 & 10 & 10 \\
q\_agg & min & min & min & min \\
subsample\_bon & False & False & False & False \\
best\_of\_n & 8 & 8 & 8 & 8 \\
\hline
\multicolumn{5}{l}{\textbf{BC Regularization}} \\
use\_bc\_regularization & True & True & True & True \\
bc\_coeff & 1.0 & 1.0 & 1.0 & 1.0 \\
clip\_bc & True & True & True & True \\
\hline

\end{tabular}
\caption{OGPO hyperparameters for Adroit.}
\label{tab:ogpo-adroit}
\end{table}

In Table \ref{tab:ogpo-libero}, we list all hyperparameters we use for the Libero environments.

\begin{table}[H]
\centering
\small
\setlength{\tabcolsep}{6pt}
\begin{tabular}{l c}
\hline
\textbf{Hyperparameter} & \texttt{Libero} \\
\hline
\multicolumn{2}{l}{\textbf{Training}} \\
offline\_steps & 50{,}000 \\
online\_steps  & 250{,}000 \\
actor\_tau & 0.001 \\
batch\_size & 64 \\
constant\_noise\_std & 0.01 \\
grpo\_num\_samples & 8 \\
\hline
\multicolumn{2}{l}{\textbf{RL Hyperparameters}} \\
horizon\_length & 8 \\
discount        & 0.999 \\
tau             & 0.05 \\
utd\_online     & 1 \\
\hline
\multicolumn{2}{l}{\textbf{Q-Network}} \\
num\_qs        & 10 \\
q\_agg         & mean \\
encoder         & impala\_small \\
value\_hidden\_dims & (128, 128, 128) \\
\hline
\multicolumn{2}{l}{\textbf{BC Regularization}} \\
use\_bc\_regularization & False \\
offline\_ratio & 0 \\
\hline
\end{tabular}
\caption{OGPO hyperparameters for Libero.}
\label{tab:ogpo-libero}
\end{table}

\end{document}